\documentclass{article}

\usepackage{microtype}
\usepackage{graphicx}
\usepackage{booktabs} 

\usepackage[nohyperref, accepted]{icml2023}

\usepackage{amsmath}
\usepackage{amssymb}
\usepackage{mathtools}
\usepackage{amsthm}

\theoremstyle{plain}
\newtheorem{theorem}{Theorem}[section]

\newtheorem{lemma}[theorem]{Lemma}

\theoremstyle{definition}
\newtheorem{definition}{Definition}

\theoremstyle{remark}

\usepackage{minitoc}
\usepackage[toc,page,header]{appendix}
\usepackage{hyperref}
\usepackage[capitalize,noabbrev]{cleveref}
\usepackage{tabularx}
\usepackage{colortbl}
\usepackage{multicol}
\usepackage{mathtools}
\usepackage{pifont}
\usepackage{cleveref}
\usepackage{amsthm}
\usepackage{tikz}
\usetikzlibrary{calc,math}
\usetikzlibrary{decorations}
\usetikzlibrary{decorations.markings}
\usetikzlibrary{tikzmark}
\usepackage{pgfplots}
\pgfplotsset{compat=1.18}
\usepackage{subcaption}
\usepackage[framemethod=TikZ]{mdframed}
\usepackage{comment}
\usepackage{fontawesome}
\usepackage{pgfplotstable}
\usepgfplotslibrary{groupplots}
\usepackage{wrapfig}
\usepackage{pst-func}
\usepackage[vlined]{algorithm2e}
\usepackage[toc,page,header]{appendix}
\usepackage{tabularx, booktabs}
\usepackage[numbers, sort&compress]{natbib}

\usepackage{enumitem}

\PassOptionsToPackage{unsrtnat, numbers}{natbib}

\hypersetup{%
    colorlinks = true,
    citecolor  = blue,
    linkcolor  = blue,
}

\mdfsetup{%
    middlelinecolor =   none,
    middlelinewidth =   1pt,
    backgroundcolor =   blue!3,
    roundcorner     =   5pt,
}

\newcolumntype{C}{>{\centering\arraybackslash}X}
\newcommand{\point}[1]{\vspace{5pt}\\\ding{117} {\em #1}\enskip}

\NewDocumentCommand{\Mark}{ m O{black} m}{%
    \tikz[remember picture, baseline, anchor=base] 
        \node[inner sep=0pt, outer sep=3pt, text=#2] (#1) {%
            \ensuremath{#3}%
        };    
}

\def\HiLi{\leavevmode\rlap{\hbox to \hsize{\color{green!15}\leaders\hrule height .8\baselineskip depth .5ex\hfill}}}

\newcommand{\cmark}{\ding{51}}
\newcommand{\xmark}{\textcolor{lightgray}{\ding{55}}}

\newcommand{\pa}[1]{\text{Pa}(#1)}
\newcommand{\ch}[1]{\text{Ch}(#1)}

\crefname{definition}{Def.}{Defs.}
\Crefname{definition}{Def.}{Defs.}

\crefname{remark}{Remark}{Remarks}
\Crefname{remark}{Remark}{Remarks}

\crefname{thm}{Theorem}{Theorems}
\Crefname{thm}{Theorem}{Theorems}

\crefname{lemma}{lem.}{lems.}
\Crefname{lemma}{Lem.}{Lems.}

\crefname{algocf}{Alg.}{Algs.}
\Crefname{algocf}{Algorithm}{Algorithms}

\crefname{figure}{Fig.}{Figs.}
\crefname{section}{Section}{Sections}
\crefname{table}{Table}{Tables}
\crefname{appendix}{Appendix}{Appendices}
\Crefname{equation}{Eq.}{Eqs.}
\crefname{equation}{eq.}{eqs.}

\newcommand\independent{\protect\mathpalette{\protect\independenT}{\perp}}
\def\independenT#1#2{\mathrel{\rlap{$#1#2$}\mkern2mu{#1#2}}}

\DeclareMathOperator*{\argmax}{arg\,max}

\usepackage[textsize=tiny]{todonotes}

\icmltitlerunning{Differentiable and Transportable Structure Learning}

\begin{document}
\doparttoc
\faketableofcontents

\twocolumn[
\icmltitle{Differentiable {\em and} Transportable Structure Learning}



\icmlsetsymbol{equal}{*}

\begin{icmlauthorlist}
\icmlauthor{Jeroen Berrevoets}{cam}
\icmlauthor{Nabeel Seedat}{cam}
\icmlauthor{Fergus Imrie}{ucla}
\icmlauthor{Mihaela van der Schaar}{cam,alan}
\end{icmlauthorlist}

\icmlaffiliation{cam}{DAMTP, University of Cambridge, UK}
\icmlaffiliation{ucla}{UCLA, CA, USA}
\icmlaffiliation{alan}{The Alan Turing Institute, UK}

\icmlcorrespondingauthor{Jeroen Berrevoets}{jeroen.berrevoets@maths.cam.ac.uk}

\icmlkeywords{Machine Learning, ICML}

\vskip 0.3in
]



\printAffiliationsAndNotice{}  

\begin{abstract}
Directed acyclic graphs (DAGs) encode a lot of information about a particular distribution in their structure. However, compute required to infer these structures is typically super-exponential in the number of variables, as inference requires a sweep of a combinatorially large space of potential structures. That is, until recent advances made it possible to search this space using a differentiable metric, drastically reducing search time. While this technique--- named NOTEARS ---is widely considered a seminal work in DAG-discovery, it concedes an important property in favour of differentiability: {\it transportability}. To be transportable, the structures discovered on one dataset must apply to another dataset from the same domain. We introduce {\it D-Struct} which recovers transportability in the discovered structures through a novel architecture and loss function while remaining fully differentiable. Because D-Struct remains differentiable, our method can be easily adopted in existing differentiable architectures, as was previously done with NOTEARS. In our experiments, we empirically validate D-Struct with respect to edge accuracy and structural Hamming distance in a variety of settings.
\end{abstract}

\section{Introduction} \label{sec:introduction}
\vspace{-3pt}
Machine learning has proven to be a crucial tool in many disciplines. With disciplines such as causal deep learning \citep{berrevoets2023causal} and applications in medicine \citep{bhardwaj2017study, berrevoets2020organite, van2021artificial, rajkomar2019machine, berrevoets2021learning}, economics \citep{athey2018impact, athey2019machine, mullainathan2017machine}, physics \citep{carleo2019machine, radovic2018machine, sarma2019machine, karniadakis2021physics, breen2020newton, udrescu2020ai}, robotics \citep{peters2007machine, peng2018sim, kehoe2015survey, abbeel2010autonomous}, and even entertainment \citep{kleiman2019boosting,ring2019jumping, wang2022film}, machine learning is transforming the way in which experts interact with their field. These successes are in large part due to increasing accuracy of diagnoses, marketing campaigns, analyses of experiments, and so forth. 

However, machine learning has much more to offer than improved accuracy, as machine learning is slowly recognised as a tool for scientific discovery \citep{davies2021advancing, jumper2021highly, tunyasuvunakool2021highly, RUFF2021167208}. In these successes, machine learning helped uncover previously unknown relationships between variables. In an effort to make these discoveries more robust, we propose D-Struct, a differentiable {\it and} transportable structure learner.

\begin{figure*}[t]
    \centering
    \vspace{-7pt}
    \begin{tikzpicture}
        \node (group) at (0, 0) {\Huge\faGroup};
        
        \node (A) at (3, .6) {\textcolor{FireBrick}{\Huge\faHospitalO}};
        \node (B) at (3, -.6) {\textcolor{blue}{\Huge\faHospitalO}};
        
        \node (dbA) at ($(A) + (3, 0)$) {\textcolor{FireBrick}{\Huge\faDatabase}};
        \node (dbB) at ($(B) + (3, 0)$) {\textcolor{blue}{\Huge\faDatabase}};
        
        \node[inner sep=4mm] (center_graph_A) at ($(dbA) + (3, 0)$) {};
        \node[inner sep=4mm] (center_graph_B) at ($(dbB) + (3, 0)$) {};

        \node[text width=2cm, align=center] at ($(group) + (0, 1.6)$) {\textcolor{black!70}{\it Population of patients.}};
        \node[text width=2cm, align=center] at ($(group) + (3, 0) + (1.5, 0)$) {\textcolor{black!80}{\it \small Different distributions.}};
        \node[text width=2cm, align=center] at ($(A) + (0, 1)$) {\textcolor{black!70}{\it Different hospitals.}};
        \node[text width=2cm, align=center] at ($(dbA) + (0, 1)$) {\textcolor{black!70}{\it Different datasets.}};
        \node[text width=2cm, align=center] at ($(center_graph_A) + (0, 1)$) {\textcolor{black!70}{\it The same DAGs.}};
        
        \draw[rounded corners, fill=blue!5, draw=none] ($(center_graph_A) + (-.7, .5)$) rectangle ($(center_graph_B) + (.8, -.7)$);
        \node[text width=2cm, align=center] at ($(center_graph_A) + (2, -.6)$) {\bf These DAGs are\\ transportable!};
        
        \begin{scope}[rotate=45]
        \node[circle, inner sep=1mm, fill=blue!50, draw=black] (a1) at ($(center_graph_A) + (0, .3)$) {};
        \node[circle, inner sep=1mm, fill=red!50, draw=black] (a2) at ($(center_graph_A) + (-.3, -.3)$) {};
        \node[circle, inner sep=1mm, fill=green!50, draw=black] (a3) at ($(center_graph_A) + (.3, -.3)$) {};
        \end{scope}

        \begin{scope}[rotate=45]
        \node[circle, inner sep=1mm, fill=blue!50, draw=black] (b1) at ($(center_graph_B) + (0, .3)$) {};
        \node[circle, inner sep=1mm, fill=red!50, draw=black] (b2) at ($(center_graph_B) + (-.3, -.3)$) {};
        \node[circle, inner sep=1mm, fill=green!50, draw=black] (b3) at ($(center_graph_B) + (.3, -.3)$) {};
        \end{scope}

        \draw[-latex, thick] (group) to[out=0, in=180] (A);
        \draw[-latex, thick] (group) to[out=0, in=180] (B);
        
        \draw[-latex, thick] (A) -- (dbA);
        \draw[-latex, thick] (B) -- (dbB);
        
        \draw[-latex, thick] (dbA) -- (center_graph_A);
        \draw[-latex, thick] (dbB) -- (center_graph_B);
        
        \draw[->] (a1) -- (a2);
        \draw[->] (a3) -- (a2);
        
        \draw[->] (b1) -- (b2);
        \draw[->] (b3) -- (b2);

    \end{tikzpicture}
    \vspace{-5pt}
    \caption{{\bf Transportability in DAG discovery.} Different patients go to different hospitals (left), yet we wish to infer a {\it general} structure (right) {\it across} hospitals. A structure can only be considered a discovery if it generalises in distributions over the same domain. For example, the way blood pressure interacts with heart disease is the same for all humans and should be reflected in the discovered structure.}
    \label{fig:pullfigure}
    \vspace{-3pt}
    \rule{\textwidth}{.5pt}
    \vspace{-20pt}
\end{figure*}
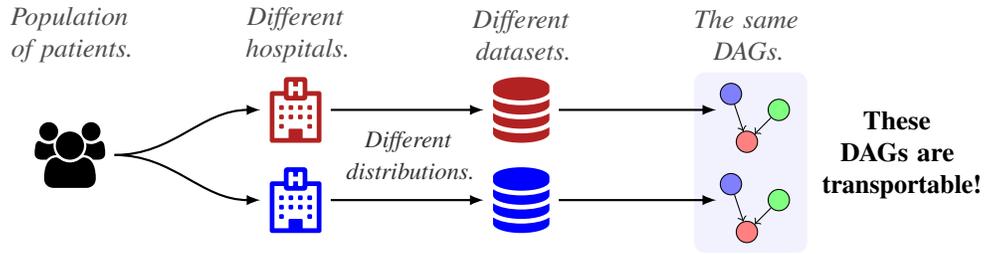

{\bf The structures.} We focus on discovering {\it directed acyclic graphs} (DAGs) in a domain $\mathcal{X}$. A DAG helps us understand how different variables in $\mathcal{X}$ interact. 
Consider a three-variable domain $\mathcal{X}\coloneqq \{X,Y,Z\}$, governed by a joint-distribution, $\mathbb{P}_\mathcal{X}$. A DAG explicitly models variable interactions in $\mathbb{P}_\mathcal{X}$. For example, consider the following DAG: $\mathcal{G} = $ \begin{tikzpicture}[
    roundnode/.style={circle, draw=black, fill=white,  minimum size=5mm, inner sep=0},
    baseline=-1mm
]
      \node[roundnode] (x) at (0,0) {$X$};
      \node[roundnode] (y) at (2,0) {$Y$};
      \node[roundnode] (z) at (1,0) {$Z$};
      
      \draw[->] (x) -- (z);
      \draw[->] (z) -- (y);
\end{tikzpicture}, where $\mathcal{G}$ depicts $\mathbb{P}_\mathcal{X}$ as a DAG. Such a DAG allows useful analysis of the (in)dependence of variables in $\mathbb{P}_\mathcal{X}$ \citep{koller2009probabilistic,wright1934method}. From $\mathcal{G}$, we learn that $X$ does not directly influence $Y$, and that $X \independent Y | Z$ as $X$ does not give any additional information on $Y$ once we know $Z$.

The above forms the basis for conventional DAG-structure learning \citep{chickering1995learning,glymour2019review,geiger1990logic,meek2013strong,eberhardt2017introduction}. In particular, $X \independent Y | Z$ strongly limits the possible DAGs that model $\mathbb{P}_\mathcal{X}$. Given more independence statements, we limit the potential DAGs further. However, independence tests are computationally expensive which is problematic as the number of potential DAGs increases super-exponentially in $|\mathcal{X}|$ \citep{peters2017elements}.

This limitation strongly impacted the adoption of DAG-learning until \citet{zheng2018dags} proposed NOTEARS, which incorporates a differentiable metric to evaluate whether or not a discovered structure is a DAG \citep{zheng2018dags, zheng2020learning}. Using automatic differentiation, NOTEARS learns a DAG-structure in a much more efficient way than methods based on conditional independence tests (CITs).

While NOTEARS makes DAG inference tractable, we recognise an important limitation in the approach: a discovered DAG does not generalise to equally factorisable distributions, i.e. NOTEARS is not {\em transportable} \citep{pearl2018book}. While we explain why this is the case in \cref{sec:prelims:dsf} (and confirm it empirically in \cref{sec:experiments}), we give a brief description of the problem below, helping us to state our contribution.

{\bf Transportability.} Consider \cref{fig:pullfigure}, depicting two hospitals: hospital A \textcolor{FireBrick}{\faHospitalO} and hospital B \textcolor{blue}{\faHospitalO}. 
Each hospital hosts patients described by the same set of features, such as age and gender. However, the hospitals may have different patient distributions, e.g. patients in A are older compared to B. {\it Crucially, their underlying biology remains the same.} Using NOTEARS to learn a DAG from data on hospital A does not guarantee the same DAG is discovered from data in hospital B, despite the two hospitals being governed by the same DAG. 
Being unable to transport findings across distributions is a major shortcoming, as replicating a discovery is considered a hallmark of the scientific method \citep{baker20161, camerer2016evaluating, merton1973sociology, stodden2010scientific}. The ability to carryover information from one distribution to another is referred to as {\it transportability} \citep{pearl2018book}.

{\bf Contributions.} In this paper, we present {\it D-Struct}, the first {\it transportable} differentiable structure learner. Transportability grants D-Struct several advantages over the state-of-the art: \textbf{\em (i) D-Struct is more accurate} which we show in a {\it variety} of settings; \textbf{\em (ii) D-Struct is fast}, in fact, we report time-to-convergence often up to 20 times faster than NOTEARS (\cref{res:computation}); 
\textbf{\em (iii) D-Struct is easily integrated} in existing architectures such as \citep{bhattacharya2021differentiable,kyono2020castle,pamfil2020dynotears,kyono2021miracle,van2021decaf}. 
Finally, despite the motivation being multi-origin data, \textbf{\em (iv) D-Struct also works for a single dataset}: using a novel subsampling routine (\cref{sec:method:single-origin}), we show that D-Struct is also more accurate when applying ideas from transportability to the single dataset setting.

\vspace{-10pt}
\section{Preliminaries and related work} \label{sec:prelims}
\vspace{-5pt}
Our goal is to build a differentiable {\it and} transportable DAG-learner. Without loss of generality, we focus our discussion mostly on NOTEARS \citep{zheng2018dags} (and refinements \citep{zheng2020learning, Lachapelle2020Gradient, yu21a, yu19a,bello2022dagma}) as it is the most adopted differentiable DAG learner. For a more in-depth overview of DAG-learners (CIT-based as well as score-based), we refer to \cref{app:related_work} or relevant literature \citep{koller2009probabilistic,peters2017elements,pearl2009causality}. Here, we discuss transportability, NOTEARS, and why NOTEARS is not transportable.

{\bf Factorisation and independence.} Consider a distribution, $\mathbb{P}_\mathcal{X}$, which we can factorise into,
\begin{equation} \label{eq:decom}
    \prod_i \mathbb{P}_{\mathcal{X}_i | \mathcal{X}_{i+1:d}},
\end{equation}
with $i \in [d]$, where $[d] \coloneqq 1, \dots, d$, and $\mathcal{X}_i$ representing the $i$\textsuperscript{th} element in $\mathcal{X}$. \Cref{eq:decom} may get quite long with increasing $d$, which becomes restrictive when learning a factorisation in $\hat{\mathbb{P}}_\mathcal{X}$ from data. Instead, we can simplify \cref{eq:decom} using independence statements, e.g.  $\mathcal{X}_i \independent \mathcal{X}_k$ invokes the equality: $\mathbb{P}_{\mathcal{X}_i | \mathcal{X}_{j, k}} = \mathbb{P}_{\mathcal{X}_i | \mathcal{X}_{j}}$. Simplifying \cref{eq:decom} results in a smaller {\it Markov boundary} \citep{pearl1988probabilistic} (see \cref{app:defs}).

{\bf Direction.} We are interested in {\it directed} and {\it acyclic} graphical (DAG) structures. Let $\mathcal{G}_\mathcal{X} \coloneqq \{\mathcal{X}, \mathcal{E}\}$ be a DAG, where $\mathcal{E} \subset \mathcal{X} \times \mathcal{X}$ is a set of edges connecting random variables in $\mathcal{X}$, with $(\mathcal{X}_i, \mathcal{X}_j) \in \mathcal{E}$ implying $(\mathcal{X}_j, \mathcal{X}_i) \not\in \mathcal{E}$ \citep{wainwright2008graphical}.

While independence is symmetric, it is still possible to infer non-symmetrical structures using only independence statements and d-separation \citep{pearl1986fusion, pearl2009causality, verma1990causal, geiger1990logic, geiger1990d, geiger1990identifying}. Given a collection of conditional independence statements, e.g. $X \independent Y | Z$, d-separation (see \cref{def:d-separation}) helps identify a directed structure \citep{howard1984principles,smith1989influence}. If a set $\mathcal{X}_d$ d-separates $\mathcal{A}$ and $\mathcal{B}$, then it blocks all their connecting paths, noted as $\text{d-sep}_\mathcal{G}(\mathcal{A};\mathcal{B}|\mathcal{X}_d)$.

With d-separation and the common faithfulness assumption (see \cref{app:related_work}), we have a link between $\mathcal{G}_\mathcal{X}$ and $\mathbb{P}_\mathcal{X}$. Specifically, conditional independence implied by $\mathcal{G}_\mathcal{X}$ corresponds to conditional independence in $\mathbb{P}_\mathcal{X}$ \citep{geiger1993logical}, i.e. if $X \independent_{\mkern-5mu\mathbb{P}\mkern5mu} Y | Z$ then $X \independent_{\mkern-5mu\mathcal{G}\mkern5mu} Y | Z$, where $\independent_{\mkern-5mu \mathcal{S} \mkern5mu}$ denotes independence in $\mathcal{S}$. The reverse is not necessarily true as there can be many {\it Markov equivalent} graphs that correspond with $\mathbb{P}$ in terms of (in)dependence \citep{koller2009probabilistic}.
The set of conditional independence assertions in $\mathbb{P}$ is denoted as $\mathcal{I}(\mathbb{P})$. Similarly, all independence statements implied by a graph $\mathcal{G}$ are denoted as $\mathcal{I}(\mathcal{G}) = \{(\mathcal{X} \independent \mathcal{B} | \mathcal{X}_d) : \text{d-sep}_\mathcal{G}(\mathcal{A}; \mathcal{B} | \mathcal{X}_d)\}$, referred to as the set of {\it global Markov independencies} \citep[Chapter 3]{koller2009probabilistic}.

{\bf Invariance and discovery.} Consider two datasets, $\mathcal{D}_1 = \{X^{(n)} \in \mathcal{X}: n \in [N] \}$ and $\mathcal{D}_2 = \{X^{(m)} \in \mathcal{X}: n \in [M] \}$, spanning the same space $\mathcal{X}$. As a sample $X^{(n)}$ from $\mathcal{D}_1$ depicts the same variables as a sample $X^{(m)}$ from $\mathcal{D}_2$, both datasets should reflect the {\it same} underlying mechanisms. For example, if hospital A collected data on its patients in $\mathcal{X}$ (say $\mathcal{D}_1$) and associated smoking with cancer, then-- {\it if true} --this should also be found in data from hospital B ($\mathcal{D}_2$).

Of course, while the samples in $\mathcal{D}_1$ and $\mathcal{D}_2$ come from the same domain $\mathcal{X}$, they may be sampled from different distributions, $\mathbb{P}_\mathcal{X}^1$ and $\mathbb{P}_\mathcal{X}^2$, respectively. As in \cref{fig:pullfigure}, hospitals A and B may be located in different regions, resulting in different patient characteristics. However, key in a scientific discovery is that it generalises {\it beyond} distributions and carries over the entire domain $\mathcal{X}$. In other words, any structure we may find in $\mathcal{D}_1$ should also be found in $\mathcal{D}_2$, as for almost all distributions $\mathbb{P}_\mathcal{X}^i\in\mathcal{P}$ that factorise over $\mathcal{G}$\footnote{For all distributions except for a measure zero set in the space of conditional probability distribution parameterizations \citep{meek1995strong}.},  $\mathcal{I}(\mathbb{P}^i_\mathcal{X}) = \mathcal{I}(\mathcal{G}) = \mathcal{I}(\mathbb{P}^j_\mathcal{X})$ where $\mathbb{P}^i_\mathcal{X} \neq \mathbb{P}^j_\mathcal{X}$ \citep[Theorem 3.5]{koller2009probabilistic}; if this is not the case, we haven't discovered anything.

{\bf Transportability.} Using the DAG to {\it carryover} conclusions from one dataset to a differently distributed other dataset is the general definition of {\it transportability} \citep{pearl2018book}. In this paper, we refine this by defining it in the context of DAGs specifically. \Cref{def:transportability} defines transportability in our context; when a DAG found in $\mathcal{D}_1$ is also found in $\mathcal{D}_2$, we consider that DAG, and the method proposing it, {\it transportable}.

\begin{definition}[Transportability]\label{def:transportability}
    With multiple datasets $\{\mathcal{D}_k \sim \mathbb{P}^k_\mathcal{X} : k \in [K]\}$ over the same domain $\mathcal{X}$, sampled from potentially different distributions $\mathbb{P}^i_\mathcal{X} \neq \mathbb{P}^j_\mathcal{X}$ if $i \neq j$ for all $i, j \in [K]$, we call a method transportable if it learns a structure that is the same across all datasets: $\{\mathcal{D}_k \to \mathcal{G}_k : k \in [K]\}$ s.t. $\mathcal{G}_1 = \dots = \mathcal{G}_K$.
\end{definition}

In the case of CIT-based methods, we are guaranteed transportability in our setting as transportability is a property directly related to the set of independencies of both distributions and DAGs. {\it But not so for differentiable structure learners.} Our goal is to propose a differentiable structure learner that exhibits this property as well.

{\bf Learning from multi-sourced data.} We stress that we do {\it not} focus on {\it just} learning from multi-sourced data. Contrasting papers on {\it federated structure learning} (FSL) \citep{gao2021federated, ng2022towards} or {\it multitask-learning} (MTL) \citep{chen2021multi}, transportability allows {\it varying} distributions across domains, thereby generalising these settings. Our setting contrasts FSL and MTL as we focus on differently distributed multi-sourced data specifically.

\vspace{-5pt}
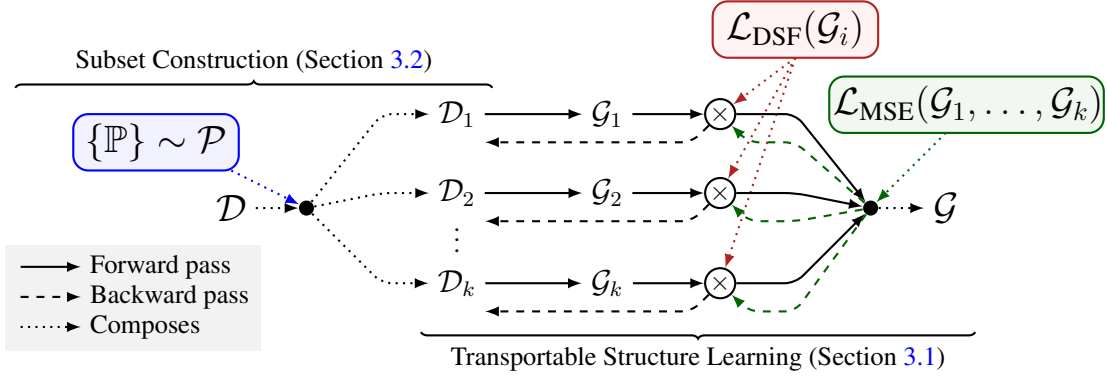
\begin{figure*}[t]
    \vspace{-7pt}
    \centering
    \begin{tikzpicture}[
        f/.style={circle, fill=black,  minimum size=2mm, inner sep=0}
    ]

          \draw[fill=black!5, draw=none] (-3, -.5) rectangle (.4, -1.9);
          \node[anchor=north west] (fp) at (-2, -.5) {Forward pass};
          \draw[thick, -latex] ($(fp.west) - (.8, 0)$) -- (fp.west);
          \node[anchor=north west] (bp) at (-2, -.9) {Backward pass};
          \draw[thick, -latex, dashed] ($(bp.west) - (.8, 0)$) -- (bp.west);
          \node[anchor=north west] (co) at (-2, -1.3) {Composes};
          \draw[thick, -latex, dotted] ($(co.west) - (.8, 0)$) -- (co.west);

          \node (D) at (0, 0) {\Large$\mathcal{D}$};
          
          \node[f] (f_sample) at (1, 0) {};
          \node[f] (f_eval) at (8.5, 0) {};
          
          \coordinate (lower_b_rects) at ($(f_sample) + (0, 2)$);
          
          \node (D1) at (3, 1.25) {\large$\mathcal{D}_1$};
          \node (D2) at (3, .2) {\large$\mathcal{D}_2$};
          \node (dot) at (3, -.4) {\rotatebox{90}{\dots}};
          \node (Dk) at (3, -1) {\large$\mathcal{D}_k$};
          
          \node (G1) at ($(D1) + (2, 0)$) {\large$\mathcal{G}_1$};
          \node (G2) at ($(D2) + (2, 0)$) {\large$\mathcal{G}_2$};
          \node (Gk) at ($(Dk) + (2, 0)$) {\large$\mathcal{G}_k$};
          
          \draw[fill=blue!5, draw=blue, thick, rounded corners=2mm] ($(lower_b_rects) - (2, 1.5) + (-1.1, .75)$) rectangle ($(lower_b_rects) - (2, 1.5) + (1.1, 0)$);
          \draw[-latex, draw=blue, thick, dotted] ($(lower_b_rects) - (1, 1.5)$) -- (f_sample);
          \node at ($(lower_b_rects) - (2, 1.5) + (0, .375)$) {\Large$\{\mathbb{P}\}\sim\mathcal{P}$};

          \draw[-latex, dotted, thick] (D) -- (f_sample);
          \draw[-latex, thick, dotted, rounded corners] (f_sample) -- ($(D1) + (-1, 0)$) -- (D1);
          \draw[-latex, thick, dotted, rounded corners] (f_sample) -- ($(D2) + (-1, 0)$) -- (D2);
          \draw[-latex, thick, dotted, rounded corners] (f_sample) -- ($(Dk) + (-1, 0)$) -- (Dk);
          
          \node[circle, thick, fill=white, draw=black, thick, inner sep=0, minimum size=4mm] (h1) at ($(G1) + (1.5, 0)$) {$\times$};
          \node[circle, thick, fill=white, draw=black, thick, inner sep=0, minimum size=4mm] (h2) at ($(G2) + (1.5, 0)$) {$\times$};
          \node[circle, thick, fill=white, draw=black, thick, inner sep=0, minimum size=4mm] (hk) at ($(Gk) + (1.5, 0)$) {$\times$};

          \draw[fill=red!5, draw=FireBrick, thick, rounded corners=2mm] ($(lower_b_rects) + (h1) - (0, 1.25) + (-1.1, .75)$) rectangle ($(lower_b_rects) + (h1) - (0, 1.25) + (1.1, 0)$);
          \draw[-latex, draw=FireBrick, thick, dotted] ($(lower_b_rects) + (h1) - (0, 1.25)$) -- (h1);
          \draw[-latex, draw=FireBrick, thick, dotted] ($(lower_b_rects) + (h1) - (0, 1.25)$) -- (h2);
          \draw[-latex, draw=FireBrick, thick, dotted] ($(lower_b_rects) + (h1) - (0, 1.25)$) -- (hk);
          \node at ($(lower_b_rects) + (h1) - (0, 1.25) + (0, .375)$) {\Large$\mathcal{L}_\text{DSF}(\mathcal{G}_i)$};

          \draw[fill=DarkGreen!5, draw=DarkGreen, thick, rounded corners=2mm] ($(f_eval) + (1.3, 1) + (-1.85, .75)$) rectangle ($(f_eval) + (1.3, 1) + (1.8, 0)$);
          \draw[-latex, draw=DarkGreen, thick, dotted] ($(f_eval) + (1,1)$) -- (f_eval);
          \node at ($(f_eval) + (1.3,1) + (0, .375)$) {\Large$\mathcal{L}_\text{MSE}(\mathcal{G}_1, \dots, \mathcal{G}_k)$};

          \draw[-latex, thick] (G1) -- (h1);
          \draw[-latex, thick] (G2) -- (h2);
          \draw[-latex, thick] (Gk) -- (hk);
          
          \draw[-latex, thick, rounded corners] (h1) -- ($(h1) + (1, 0)$) -- (f_eval);
          \draw[-latex, thick, rounded corners] (h2) -- ($(h2) + (1, 0)$) -- (f_eval);
          \draw[-latex, thick, rounded corners] (hk) -- ($(hk) + (1, 0)$) -- (f_eval);
          
          \draw[-latex, thick] (D1) -- (G1);
          \draw[-latex, thick] (D2) -- (G2);
          \draw[-latex, thick] (Dk) -- (Gk);

          \draw[-latex, thick, dashed, rounded corners, draw=DarkGreen] (f_eval) -- ($(h1) + (1, 0) - (0, .375)$) -- ($(h1) - (-.375, .375)$) -- (h1);
          \draw[-latex, thick, dashed, rounded corners, draw=black] (h1) -- ($(h1) - (.375, .375)$) -- ($(D1) - (-.375, .375)$);
          
          \draw[-latex, thick, dashed, rounded corners, draw=DarkGreen] (f_eval) -- ($(h2) + (1, 0) - (0, .375)$) -- ($(h2) - (-.375, .375)$) -- (h2);
          \draw[-latex, thick, dashed, rounded corners, draw=black] (h2) -- ($(h2) - (.375, .375)$) -- ($(D2) - (-.375, .375)$);
          
          \draw[-latex, thick, dashed, rounded corners, draw=DarkGreen] (f_eval) -- ($(hk) + (1, 0) - (0, .375)$) -- ($(hk) - (-.375, .375)$) -- (hk);
          \draw[-latex, thick, dashed, rounded corners, draw=black] (hk) -- ($(hk) - (.375, .375)$) -- ($(Dk) - (-.375, .375)$);
          
          \node (G) at ($(f_eval) + (1, 0)$) {\Large$\mathcal{G}$};
          \draw[-latex, dotted, thick] (f_eval) -- (G);

        \draw [
            thick,
            decoration={
                brace,
                mirror,
                raise=.15cm
            },
            decorate
        ] ($(Dk.west) + (-.1, -.5)$) -- ($(G.east) + (.1, -1.5)$) node [pos=0.5,anchor=center,yshift=-.5cm] {Transportable Structure Learning (\cref{sec:method:general})};
        
        \draw [
            thick,
            decoration={
                brace,
                raise=-.15cm
            },
            decorate
        ] ($(D1.west) + (-5.5, .5)$) -- ($(D1.east) + (.1, .5)$) node [pos=0.5,anchor=center,yshift=.2cm] {Subset Construction (\cref{sec:method:single-origin})};

    \end{tikzpicture}
    \caption{{\bf D-Struct architecture.} D-Struct is split into two major parts: subset construction (\cref{sec:method:single-origin}) and the transportable structure learning algorithm (\cref{sec:method:general}). 
    The losses, $\mathcal{L}_\text{DSF}$ and $\mathcal{L}_\text{MSE}$, are combined and backpropagated through the architecture to enforce transportability. Lastly, all DSFs are merged into a final DAG structure $\mathcal{G}$.}
    \rule{\textwidth}{.75pt}
    \label{fig:architecture}
    \vspace{-25pt}
\end{figure*}

\subsection{Differentiable structure learning} \label{sec:prelims:dsf}

CIT-based methods evaluate each Markov equivalent DAG using $\mathcal{I}(\mathcal{G} \in \mathbb{G}_\mathcal{X})$, where $\mathbb{G}_\mathcal{X}$ denotes the space of all possible DAGs in the domain $\mathcal{X}$. The major issue with this is computation. Essentially, there are two aspects that negatively impact computation time: first, the number of to-be-evaluated DAGs in $\mathbb{G}_\mathcal{X}$ increases super-exponentially in $|\mathcal{X}|$ (e.g. 10 variables result in $> 4 \times 10^{18}$ possible DAGs \citep{peters2017elements,robinson1977counting,vowels2021d}); second, simply recovering $\mathcal{I}(\mathbb{P}_\mathcal{X})$ to evaluate each $\mathcal{G}\in\mathbb{G}_\mathcal{X}$ requires many independence tests, each with additional compute. \Cref{app:related_work} includes an overview of the most well-known CIT-based (and score-based) methods.

{\bf Differentiable score functions.} Enter {\it differentiable score functions} (DSFs). With DSFs one traverses $\mathbb{G}_\mathcal{X}$ {\it smartly}, arriving at a DAG much faster \citep{vowels2021d,ng2022convergence,wei2020dags}. Furthermore, a differentiable method is easily included in a variety of differentiable architectures, allowing joint optimisation of both the graphical structure as well as the accompanying structural equations or another downstream use \citep{bhattacharya2021differentiable,kyono2020castle,pamfil2020dynotears,kyono2021miracle,van2021decaf}. 

Most notable is NOTEARS \citep{zheng2018dags}, proposing to optimise:
\begin{equation} \label{eq:notears}
    \min_{A \in \mathcal{A}} F(A) + \lambda_1 \lVert A \rVert_1 + \frac{\rho}{2}|h(A)|^2 + \lambda_2 h(A),
\end{equation}
where $A\in\mathbb{R}^{d\times d}$ denotes an adjacency matrix; $F(A)$ is a likelihood-based loss (like the MSE); $\rho$ and $\lambda_{1,2}$ are the parameters of their proposed augmented Lagrangian; and 
\begin{equation} \label{eq:h}
    h(A) \coloneqq \text{tr}(\exp(A \circ A)) - d,
\end{equation}
is the actual differentiable score function, where $\text{tr}(\cdot)$ is the matrix trace operator and $\circ$ is the element-wise (Hadamard) product. Importantly, $h(A) = 0$ indicates $A$ is a DAG. Considering that \cref{eq:h} is differentiable, we can take its derivative with respect to $A$ and minimise \cref{eq:notears,eq:h}.

Naturally, gradient-based learning may guide optimisation in different directions with different random initialisations of $A$, potentially arriving at different local minima. This is certainly the case in recent improvements of NOTEARS as they almost exclusively focus on non-linear structural equations which result in non-convex losses \citep{zheng2020learning,yu19a,yu21a}.

{\bf Transportability of DSFs.} Current DSFs are not transportable due to \cref{eq:notears} having conflicting solutions--- contrasting the single solution (set) that transforms $\mathcal{I}(\mathbb{P})$ to $\mathcal{G}$. Essentially, the approximate nature of (stochastic) gradient-based learning can result in conflicting estimate structures \citep{chen2021multi}, shown empirically in \cref{sec:experiments,app:notearsdags}.

\vspace{-5pt}
\section{D-Struct: {\em D}ifferentiable and transportable {\em Struct}ure learning} \label{sec:method}
\vspace{-5pt}
Structure learners transform finite data into structure:
\begin{equation*}
    \mathcal{D} \to \mathcal{G},
\end{equation*}
as does D-Struct. We introduce D-Struct in \cref{sec:method:general} and immediately extend to a single-dataset setting in \cref{sec:method:single-origin}. In \cref{sec:method:notears} we provide implementation details using NOTEARS as a comparison. Each of our claims is backed by empirical evidence in \cref{sec:experiments,app:additional-experiments}.

\vspace{-8pt}
\subsection{D-Struct: Transportable structure learning} \label{sec:method:general}
\vspace{-5pt}

To enforce transportability, D-Struct employs an ensemble architecture of multiple initialisations of a chosen DSF and their architecture. Each loss is then combined with a regularisation function based on the D-Struct architecture. \Cref{fig:architecture} depicts this architecture, highlighting how our regularisation scheme is backpropagated throughout the entire network.

Given datasets ${\mathcal{D}_1, \dots, \mathcal{D}_K}$, we can use any DSF (e.g. \citep{zheng2018dags, zheng2020learning,Lachapelle2020Gradient,yu19a,yu21a,bello2022dagma}) to learn a DAG. Specifically, we let $K$ {\it distinct} DSFs learn a DAG from one of the $K$ datasets, agnostic from each other. We consider these learning objectives to be $K$ parallel objectives, as illustrated in \cref{fig:architecture} in the rightmost part. Crucially, D-Struct does not restrict which type of DSF we can use. In a linear setting, one can use vanilla NOTEARS \citep{zheng2018dags}, whereas in a non-linear setting, one can use the non-parametric version \citep{zheng2020learning}. Naturally, any restriction posed by the chosen DSF will be inherited by D-Struct. We use NOTEARS-MLP \citep{zheng2020learning} in \cref{sec:experiments,sec:method:notears}, while \Cref{app:additional-experiments} includes pairings with other DSFs.

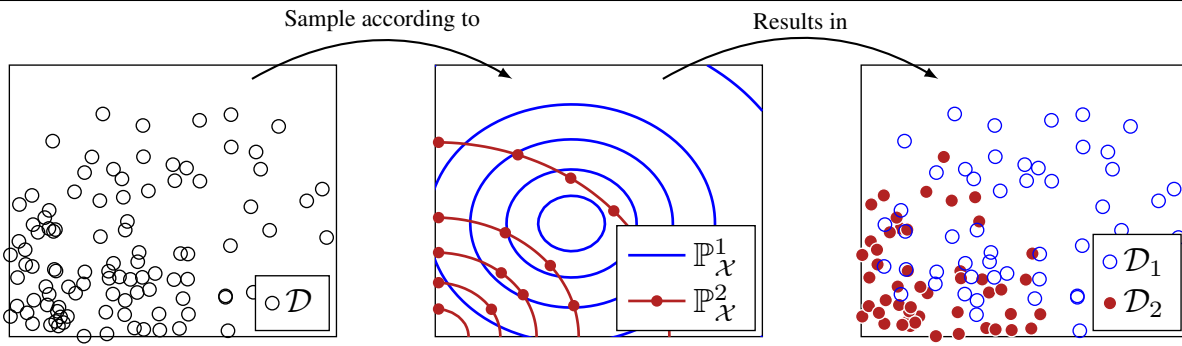
\begin{figure*}[t]
\vspace{15pt}
    \centering
    \begin{subfigure}[t]{0.32\textwidth}
    \centering
    \begin{tikzpicture}[remember picture]

        \node[inner sep=0, anchor=north] at (0, 0) {
            \includegraphics[width=.85\textwidth]{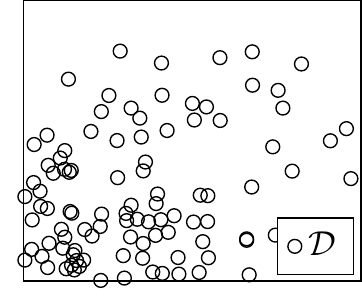}
        };

        \coordinate (A) at (1.2, -.2);
    \end{tikzpicture}
    \caption{We are presented with a dataset $\mathcal{D}$ over the domain $\mathcal{X}$.}
    \label{fig:subsets:dataset}
    
    \end{subfigure}%
    ~
    \begin{subfigure}[t]{.32\textwidth}
    \centering
    \begin{tikzpicture}[
        remember picture,
    ]

  

        
        
        \node[inner sep=0, anchor=north] at (0, 0) {
            \includegraphics[width=.85\textwidth]{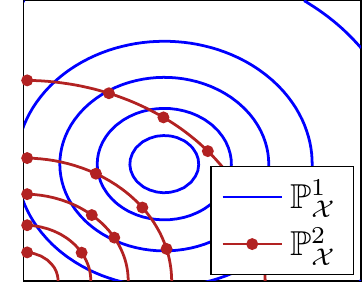}
        };

        \coordinate (B) at (-1, -.2);
        \coordinate (C) at (1, -.2);
    \end{tikzpicture}
    \caption{With two distributions $\mathbb{P}_\mathcal{X}^1$ and $\mathbb{P}_\mathcal{X}^2$, we can sample from $\mathcal{D}$.}
    \label{fig:subsets:dists}
    
    \end{subfigure}%
    ~
    \begin{subfigure}[t]{.32\textwidth}
    \centering
    \begin{tikzpicture}[remember picture]
        \node[inner sep=0, anchor=north] at (0, 0) {
            \includegraphics[width=.85\textwidth]{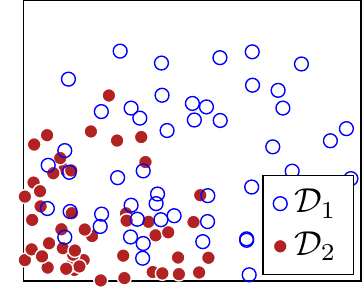}
        };
        \coordinate (D) at (-1, -.2);
    \end{tikzpicture}
    \caption{Sampling according to two distributions results in two subsets $\mathcal{D}_1 \cup \mathcal{D}_2 = \mathcal{D}$.}
    \label{fig:subsets:sampled}
    
    \end{subfigure}%
    \begin{tikzpicture}[overlay, remember picture]
          \path[-latex, thick] (A) edge [bend left] node[above, midway] {\small Sample according to} (B);
          \path[-latex, thick] (C) edge [bend left] node[above, midway] {\small Results in} (D);
    \end{tikzpicture}
    \vspace{-8pt}
    \caption{{\bf Differently distributed {\em single-origin} data.} (a) We illustrate a single-origin dataset $\mathcal{D}$, sampled from one distribution. (b) We illustrate two distributions over the domain of $\mathcal{D}$, which are used to resample two subsets from $\mathcal{D}$, thereby creating a new multi-origin datasource (c).}
    \vspace{-5pt}
    \rule{\textwidth}{.75pt}
    \label{fig:subsets}
    \vspace{-20pt}
\end{figure*}

At this point, we identify a first loss term: $\mathcal{L}_\text{DSF}(\mathcal{G}_k)$, which depends on the chosen DSF (illustrated in red in \cref{fig:architecture}). In the case of NOTEARS, $\mathcal{L}_\text{DSF}(\mathcal{G}_k)$ corresponds with \cref{eq:notears,eq:h}. Whenever data is passed through the architecture-- without mixing distinct datasets --we evaluate the discovered structure as $\mathcal{L}_\text{DSF}(\mathcal{G}_k | \mathbf{X} \sim \mathcal{D}_k)$, where $\mathbf{X} \subseteq \mathcal{D}$. If the chosen DSF requires hyperparameters (such as $\lambda_{1, 2}$ and $\rho$ in \cref{eq:A:loss}), we have to also include these in D-Struct's set of required hyperparameters. While it is possible to set different hyperparameter values for each of the DSFs separately (which is potentially helpful when there is a lot of variety in the $K$ distinct datasets), we fix these across DSFs in light of simplicity. A discussion on D-Struct's hyperparameters can be found in \cref{app:experiment-details}.

Given $\{\mathcal{L}_\text{DSF}(\mathcal{G}_k) : k \in [K]\}$ we enforce transportability across each $\mathcal{D}_k$ by comparing the structures $\mathcal{G}_1, \dots, \mathcal{G}_k$. We do this by calculating the difference of the adjacency matrices $A_k \in \mathbb{R}^{d \times d}$. Specifically, for each gradient calculation (before we perform a backward pass), we take the (element-wise) mean adjacency matrix, $\Bar{A}_{1:K} = \frac{1}{K}\sum_k A_k$, detach it from the gradient and backpropagate the MSE for each parallel DSF. In particular, we include the following regularisation term in D-Struct's loss:
\begin{equation}\label{eq:A:loss}
    \mathcal{L}_\text{MSE}(A_k) \coloneqq \lVert A_k - \Bar{A}_{1:K} \rVert_2^2.
\end{equation}
Minimising \cref{eq:A:loss} results in transportable structures (see  \cref{rem:transportability}). Note that \cref{eq:A:loss} (green in \cref{fig:architecture}) remains differentiable, which was our goal for D-Struct. We add $\mathcal{L}_\text{MSE}(\mathcal{G}_k)$ to the DSF loss,
\begin{equation} \label{eq:loss}
    \mathcal{L}(\mathcal{G}_k | \mathcal{D}_k) \coloneqq \mathcal{L}_\text{DSF}(\mathcal{G} | \mathcal{D}_k) + \alpha \mathcal{L}_\text{MSE}(A(\mathcal{G}_k)),
\end{equation}
where $A(\mathcal{G})$ indicates the adjacency matrix of $\mathcal{G}$, and $\alpha$ is a scalar hyperparameter (refer to \cref{app:experiment-details} for hyperparameter settings, details, and further insights). Note that the second term in \cref{eq:loss} does not depend on $\mathcal{D}_k$. Having $\mathcal{L}_\text{MSE}$ be agnostic to the data makes sense as transportability is not a property of the data. Indeed, recall from \cref{sec:prelims} that transportability is a property of the structure learner instead. 

Including the term given by \cref{eq:A:loss} in \cref{eq:loss} enforces transportability as the architecture encourages the DSFs to converge to the same adjacency matrix, as per \cref{rem:transportability}.

\begin{theorem}[\textbf{Minimising \cref{eq:A:loss} yields transportability.}] \label{rem:transportability}
\end{theorem}

\vspace{-20pt}
\begin{proof}
\Cref{eq:A:loss} is equal to $0$--- for {\bf every} adjacency matrix $A_k$ ---when $A_1 = \dots = A_K$. Even a slight difference in one of the $A_k$ will result in a non-zero \eqref{eq:A:loss} as $\Bar{A}_{1:K}$ will be affected, resulting in $|A_k - \Bar{A}_{1:K}|>0$. Having every $A_1 = \dots = A_K$ and thus equal structures in $\mathcal{G}_k$--- where each $A_k$ is learned from a distinct $\mathcal{D}_k$ ---corresponds with transportable structures as we have defined in \cref{def:transportability}
\end{proof}

\subsection{D-Struct: Subset construction} \label{sec:method:single-origin}

\begin{figure}
    \centering
    \includegraphics[width=.77\linewidth]{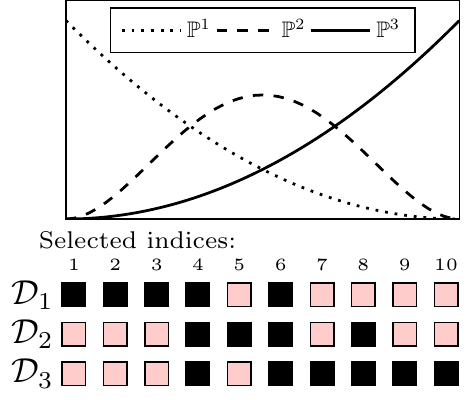}
    \vspace{-8pt}
    \caption{{\bf $K$ distributions.} We have illustrated the subset sampling with beta-distributions above, for $K=3$. For each density and index, we evaluate its PDF, normalize it and perform a Bernoulli experiment. The selected indices are shown below the PDFs (black indicates a selected index).}
    \label{fig:k-distributions}
    \vspace{-5pt}
    \rule{\linewidth}{.5pt}
    \vspace{-30pt}
\end{figure}

In \cref{sec:method:general}, we assumed data is provided in multiple distinct datasets, i.e. they stem from a multi-origin datasource. However, here we explain how even in the single-origin case D-Struct is applicable, irrespective of which DSF we end up choosing. 
Naturally, if one already has distinct data, $\mathcal{D}_k \sim \mathbb{P}^k$, one can use D-Struct as proposed in \cref{sec:method:general}.

Different distributions may guide each (distinct) optimisation target in a different direction. Combining their results will encourage the total model to be more robust and generalisable.
However, while a multi-origin datasource may be governed by multiple distributions, a single-origin one is not. Our task is clear: from a single-origin datasource, we have to {\it mimic} a multi-origin datasource in such a way that we know each subset has a different distribution, yet maintains the  properties of the original single-origin-distribution. Doing so allows us to enforce transportability through \cref{eq:A:loss}. 

The lefthand side of \Cref{fig:architecture} shows that we need to {\it construct} a multi-origin setup, prior to using D-Struct as we have done in \cref{sec:method:general}. We preface the multi-origin case with a step that divides $\mathcal{D}$ into subsets $\{\mathcal{D}_1, \dots, \mathcal{D}_k\}$, according to different distributions $\mathcal{P} \coloneqq \{\mathbb{P}_\mathcal{X}^1,\dots, \mathbb{P}_\mathcal{X}^k \}$. In \cref{fig:subsets}, we illustrate how we sample from $\mathcal{D}$ using $\mathbb{P}^k \in \mathcal{P}$. In principle, each element $X^{(n)} \in \mathcal{D}$ has a $\mathbb{P}^k(X^{(n)})$ probability to be sampled from $\mathcal{D}$, for each $\mathbb{P}^k \in \mathcal{P}$. As such, each distribution leads to a subset $\mathbb{P}^k \times \mathcal{D} \to \mathcal{D}_k$ where $\bigcup_k \mathcal{D}_k = \mathcal{D}$, and $\mathcal{D}_k$ need not be disjoint but is not equal to $\mathcal{D}$.

We perform this preprocessing step in three parts: {\bf Step 1}, we correlate the index of each element in $\mathcal{D}$ with their values in $\mathcal{X}$. {\bf Step 2}, we define $K$ distributions over $[N]$ and then in {\bf Step 3} we use these distributions to sample indices. The sampled indices compose the subset. While we have included a detailed description of our implementation in \cref{app:subsample}, we give a brief step-wise explanation below.
\point{{\bf Step 1} Correlating indices and values.} Reindexing $\mathcal{D}$ according to some ordering in $\mathcal{X}$ ensures a dependency between $\mathcal{X}$ and $i\in[N]$, where $i < j$ indicates $X^{(i)} < X^{(j)}$, i.e. the {\it order} of $X$'s in the data structure representing $\mathcal{D}$ is correlated with the {\it values} of the $X$'s. 
\point{{\bf Step 2} Distributions over $[N]$.} Step 1 allows us to create subsets based on one-dimensional distributions $\{\mathbb{P}_{[N]}^k:k\in[K]\}$, rather than more complicated distributions over $\mathcal{X}$. An added bonus to these one-dimensional distributions is that they easily scale to more dimensions in $\mathcal{X}$. Of course, the number of distributions, and consequentially their shape, should change as a function of $K$. Specifically, with higher $K$, we have to ensure that the probability mass of each distribution is concentrated in different areas of $[N]$. As such, we chose to model these as beta-distributions with, 
\begin{multline*}
    \alpha, \beta \in \big\{(i, K), (K, K), (K, j): \\ i \in \text{interp}(1, K-1), j \in \text{interp}(K-1, 1)\big\},
\end{multline*}
where $\text{interp}(a, b)$ is a linear interpolation between $a$ and $b$, used to sample $\lfloor \frac{K}{2} \rfloor$ $i$'s and $j$'s. When $K$ is even we omit $(K, K)$ so that the number of distributions always equals $K$.
\point{{\bf Step 3} Selecting indices.} Our final task is to create $K$ subsets, which due to Step 1 is simplified to choosing indices. These indices are selected based on the distributions defined in Step 2. First, we evaluate each density's PDF for every index (after normalisation: $\frac{i}{N}$) and normalise the output to be between $0$ and $1$. Once we have $K$ values for each index, we perform Bernoulli experiments to determine whether the index is selected as part of subset $k\in[K]$. This is illustrated in \cref{fig:k-distributions} for $K=3$ using beta distributions.

In our experiments (\cref{sec:experiments,app:additional-experiments}), we show that D-Struct greatly improves the performance of non-transportable DSFs. Furthermore, we empirically validate our subsampling routing compared to random sampling.

\vspace{-5pt}
\subsection{Example implementation using \texttt{NOTEARS-MLP}} \label{sec:method:notears}
\vspace{-5pt}
D-Struct works with any DSF, though it is instructive to illustrate this with an example. For this, we chose \texttt{NOTEARS-MLP} \citep{zheng2020learning} which is a non-parametric (cfr. the structural equations) extension of the classic NOTEARS paper \citep{zheng2018dags}. The main challenge to incorporating D-Struct into \texttt{NOTEARS-MLP} is to integrate it into its dual ascent strategy, which solves the (non-convex) constrained optimisation problem in \cref{eq:notears} \citep{Lachapelle2020Gradient} with an augmented Lagrangian method \citep[Chapter 5]{bertsekas1999}. 

The constraint in the optimisation problem stems from, for example, knowing that the diagonal of $A$ can only contain zeros \citep{zheng2018dags, zheng2020learning,Lachapelle2020Gradient}. NOTEARS (and its extensions) solve this problem by using the L-BFGS-B optimizer \citep{byrd1995limited}, which can handle parameter bounds out-of-the-box, making it a suitable choice to optimise the augmented Lagrangian\footnote{This also allows including prior knowledge on $\mathcal{I}(\mathbb{P})$. We discuss this in more detail in \cref{app:priors}.}. This is made explicit in \cref{alg:dstruct-notears,alg:dstruct-notears:training-step}. 
\vspace{-2pt}


\begin{algorithm}
\caption{Outer-loop of dual ascent procedure for D-Struct(\texttt{NOTEARS-MLP})}\label{alg:dstruct-notears}
\SetKwInput{In}{Input}
\SetKwInOut{Setup}{Setup}
\SetKwInput{Init}{Init.}

\HiLi\Init{$\theta_k$ for each $k \in [K]$}

\In{$h_\text{tol}$, $\rho_\text{max}$}
\Setup{$h \leftarrow\infty$, $\rho_{1,\dots,K} \leftarrow 1$, $\rho \leftarrow 1$}

\For{maximum amount of epochs}{
    
    \HiLi\For{$k \in [K]$}{
        \For{$\text{batch} \sim \mathcal{D}_k$}{
            \HiLi\texttt{training\_step(}$\theta_k$, {\it batch}\texttt{)}\tikzmark{algo1}\;
            \HiLi$h \leftarrow \max_k h(A(\theta_k))$\;
            \HiLi$\rho \leftarrow \min_k \rho_k$\;
        }
    }
}
\end{algorithm}

\begin{algorithm}

\caption{\texttt{training\_step} for D-Struct(\texttt{NOTEARS-MLP}) cfr. \cref{alg:dstruct-notears}}\label{alg:dstruct-notears:training-step}
\SetKwInput{In}{Input}
\SetKwInOut{Setup}{Setup}
\SetKwInput{Init}{Init.}

\In{$\theta_k$, batch \tikzmark{algo2}}

\While{$\rho < \rho_\text{max}$}{
    \HiLi$l_\text{m}\leftarrow\mathcal{L}_\text{MSE}(\theta_1,\dots,\theta_K)$\;
    $l_\text{d} \leftarrow \mathcal{L}_\text{DSF}(\text{\it batch})$\;
    \HiLi$\theta \leftarrow $\texttt{L-BFGS-B.update(}$l_\text{m}, l_\text{d}$\texttt{)}\;
    \HiLi$h' \leftarrow h(A(\theta_k))$\;
    \uIf{$h' > 0.25h$}{
        $\rho_k \leftarrow 10\rho_k$\;
    }
    \Else{
        \texttt{break}\;
    }
}

\end{algorithm}


\vspace{-7pt}
\Cref{alg:dstruct-notears,alg:dstruct-notears:training-step} \colorbox{green!15}{highlight} algorithmic differences between D-Struct and \texttt{NOTEARS-MLP}. Most obvious is the creation of multiple parameters $\theta_k$ for each $k\in[K]$, where each $\theta_k$ indicates the set of parameters for one initialisation of \texttt{NOTEARS-MLP}, following the architecture depicted in \cref{fig:architecture}. The set $\{\theta_k:k\in[K]\}$ then denotes the parameters for D-Struct. As such, the number of parameters for D-Struct scales linearly in $K$, compared to the used DSFs.

\begin{table*}[t]
    \vspace{-5pt}
    \centering
    \caption{{\bf Results on Erdos-Renyì (ER) graphs.} {\it First block:} We sample ten different ER random graphs, and accompanying non-linear structural equations as in \citet{zheng2020learning}. From each system, we then sample a varying number of samples and evaluate \texttt{NOTEARS-MLP} {\it with} D-Struct (indicated as {``\cmark''}) and {\it without} D-Struct (indicated as {``\xmark''}). {\it Second block:} For each row we sample ten new graphs with varying connectedness ($s$ is the expected number of edges). {\it Third block:} Each row varies the variables-count ($d$) and samples ten new random graphs. In all cases, we report the average performance in terms of SHD, FPR, TPR, and FDR, with std in scriptsize. Unless otherwise indicated, $n=1000, d=5, K=3, s=2d$.}
    \vspace{-4pt}
    \label{tab:results}
    \begin{tabularx}{\textwidth}{r  *{4}{|CC}}
    \toprule
        {\it metric} & 
        
        \multicolumn{2}{c}{\bf SHD ($\downarrow$)}  &
        \multicolumn{2}{c}{\bf FPR ($\downarrow$)}  &
        \multicolumn{2}{c}{\bf TPR ($\uparrow$)}    &
        \multicolumn{2}{c}{\bf FDR ($\downarrow$)}  \\
        
        \midrule
        
        {\it D-Struct}& 
        \cmark & \xmark & 
        \cmark & \xmark & 
        \cmark & \xmark & 
        \cmark & \xmark \\
        
        \toprule
        $n$& \multicolumn{8}{c}{\it varying sample size}\\
        \midrule
        $200$&
        {\footnotesize \bf 3.60}{\scriptsize $\pm$0.27}  &   {\footnotesize 4.20}{\scriptsize $\pm$0.44}&
        {\footnotesize \bf 2.00}{\scriptsize $\pm$0.67}  &   {\footnotesize 4.20}{\scriptsize $\pm$0.44}&
        {\footnotesize \bf 0.67}{\scriptsize $\pm$0.05}  &   {\footnotesize  0.64}{\scriptsize $\pm$0.05}&
        {\footnotesize \bf 0.25}{\scriptsize $\pm$0.06}  &   {\footnotesize 0.42}{\scriptsize $\pm$0.04}\\

        $500$&
        {\footnotesize \bf 3.20}{\scriptsize $\pm$0.80}  &   {\footnotesize 3.94}{\scriptsize $\pm$0.33}&
        {\footnotesize \bf 1.20}{\scriptsize $\pm$0.44}  &   {\footnotesize 3.94}{\scriptsize $\pm$0.33}&
        {\footnotesize \bf 0.66}{\scriptsize $\pm$0.24}  &   {\footnotesize 0.56}{\scriptsize $\pm$0.04}&
        {\footnotesize \bf 0.18}{\scriptsize $\pm$0.05}  &   {\footnotesize 0.44}{\scriptsize $\pm$0.04}\\

        $1000$&
        {\footnotesize \bf 2.75}{\scriptsize $\pm$0.47}  &   {\footnotesize 3.67}{\scriptsize $\pm$0.82}&
        {\footnotesize \bf 1.00}{\scriptsize $\pm$0.37}  &   {\footnotesize 2.67}{\scriptsize $\pm$0.63}&
        {\footnotesize \bf 0.75}{\scriptsize $\pm$0.08}  &   {\footnotesize  0.63}{\scriptsize $\pm$0.13}&
        {\footnotesize \bf 0.18}{\scriptsize $\pm$0.03}  &   {\footnotesize 0.39}{\scriptsize $\pm$0.11}\\

        $2000$&
        {\footnotesize \bf 2.66}{\scriptsize $\pm$0.80}  &   {\footnotesize 3.54}{\scriptsize $\pm$0.16}&
        {\footnotesize \bf 1.88}{\scriptsize $\pm$0.67}  &   {\footnotesize 2.09}{\scriptsize $\pm$0.31}&
        {\footnotesize \bf 0.81}{\scriptsize $\pm$0.11}  &   {\footnotesize 0.75}{\scriptsize $\pm$0.03}&
        {\footnotesize \bf 0.27}{\scriptsize $\pm$0.07}  &   {\footnotesize 0.33}{\scriptsize $\pm$0.00}\\

        \midrule
        $s$& \multicolumn{8}{c}{\it varying graph connectedness}\\
        \midrule
        $0.5d$&
        {\footnotesize \bf 3.75}{\scriptsize $\pm$1.6}  &   {\footnotesize 7.33}{\scriptsize $\pm$0.13}&
        {\footnotesize \bf 0.50}{\scriptsize $\pm$0.25}  &   {\footnotesize 1.05}{\scriptsize $\pm$0.02}&
        {\footnotesize 0.83}{\scriptsize $\pm$0.19}  &   {\footnotesize \bf 0.88}{\scriptsize $\pm$0.04}&
        {\footnotesize \bf 0.42}{\scriptsize $\pm$0.16}  &   {\footnotesize 0.73}{\scriptsize $\pm$0.01}\\

        $1d$&
        {\footnotesize \bf 3.50}{\scriptsize $\pm$0.86}  &   {\footnotesize 7.67}{\scriptsize $\pm$0.45}&
        {\footnotesize \bf 0.55}{\scriptsize $\pm$0.22}  &   {\footnotesize 1.53}{\scriptsize $\pm$0.09}&
        {\footnotesize \bf 0.75}{\scriptsize $\pm$0.09}  &   {\footnotesize 0.46}{\scriptsize $\pm$0.09}&
        {\footnotesize \bf 0.40}{\scriptsize $\pm$0.09}  &   {\footnotesize  0.77}{\scriptsize $\pm$0.07}\\

        $1.5d$&
        {\footnotesize \bf 3.00}{\scriptsize $\pm$1.15}  &   {\footnotesize 5.67}{\scriptsize $\pm$1.75}&
        {\footnotesize \bf 1.00}{\scriptsize $\pm$0.19}  &   {\footnotesize 1.55}{\scriptsize $\pm$0.08}&
        {\footnotesize \bf 0.89}{\scriptsize $\pm$0.07}  &   {\footnotesize 0.62}{\scriptsize $\pm$0.06}&
        {\footnotesize \bf 0.32}{\scriptsize $\pm$0.05}  &   {\footnotesize 0.53}{\scriptsize $\pm$0.04}\\

        $2d$&
        {\footnotesize \bf 2.28}{\scriptsize $\pm$0.80}  &   {\footnotesize 3.67}{\scriptsize $\pm$0.82}&
        {\footnotesize \bf 1.00}{\scriptsize $\pm$0.32}  &   {\footnotesize 2.67}{\scriptsize $\pm$0.63}&
        {\footnotesize 0.67}{\scriptsize $\pm$0.17}  &   {\footnotesize \bf 0.70}{\scriptsize $\pm$0.09}&
        {\footnotesize \bf 0.11}{\scriptsize $\pm$0.03}  &   {\footnotesize 0.32}{\scriptsize $\pm$0.08}\\

        \midrule
        $d$& \multicolumn{8}{c}{\it varying dimension count}\\
        \midrule
        $5$&
        {\footnotesize \bf 2.28}{\scriptsize $\pm$0.80}  &   {\footnotesize 3.67}{\scriptsize $\pm$0.82}&
        {\footnotesize \bf 1.00}{\scriptsize $\pm$0.32}  &   {\footnotesize 2.67}{\scriptsize $\pm$0.63}&
        {\footnotesize 0.67}{\scriptsize $\pm$0.17}  &   {\footnotesize \bf 0.70}{\scriptsize $\pm$0.09}&
        {\footnotesize \bf 0.11}{\scriptsize $\pm$0.03}  &   {\footnotesize 0.32}{\scriptsize $\pm$0.08}\\

        $7$&
        {\footnotesize \bf 8.67}{\scriptsize $\pm$0.56}  &    {\footnotesize  12.9}{\scriptsize $\pm$0.15}&
        {\footnotesize \bf 0.72}{\scriptsize $\pm$0.05}  &   {\footnotesize 1.07}{\scriptsize $\pm$0.01}&
        {\footnotesize \bf 0.96}{\scriptsize $\pm$0.02}  &   {\footnotesize 0.83}{\scriptsize $\pm$0.01}&
        {\footnotesize \bf 0.49}{\scriptsize $\pm$0.01}  &   {\footnotesize  0.63}{\scriptsize $\pm$0.01}\\

        $10$&
        {\footnotesize \bf 19.71}{\scriptsize $\pm$0.72}  &   {\footnotesize 30.8}{\scriptsize $\pm$0.98}&
        {\footnotesize \bf 0.42}{\scriptsize $\pm$0.13}  &   {\footnotesize 1.18}{\scriptsize $\pm$0.04}&
        {\footnotesize  0.70}{\scriptsize $\pm$0.16}  &   {\footnotesize \bf 0.71}{\scriptsize $\pm$0.06}&
        {\footnotesize \bf 0.34}{\scriptsize $\pm$0.08}  &   {\footnotesize 0.70}{\scriptsize $\pm$0.02}\\

    \bottomrule
    \end{tabularx}
    \vspace{-15pt}
\end{table*}

From \cref{alg:dstruct-notears,alg:dstruct-notears:training-step}, we learn that information across the different \texttt{NOTEARS-MLP}s is shared in \texttt{training\_step} (corresponding to \cref{alg:dstruct-notears:training-step}). Typically, a training step is solely focused on one structure learner leaving the learner unaware of the other DSFs, as is also implied in \cref{alg:dstruct-notears} which iterates over each learner separately. Sharing information {\it across} each learner--- through $\mathcal{L}_\text{MSE}(\theta_1, \dots, \theta_k)$ computed in the first line in \cref{alg:dstruct-notears:training-step}'s while loop ---enforces transportability.

D-Struct hardly increases implementation complexity. In fact, besides architectural alterations (as explained in \cref{sec:method:general,fig:architecture}), the optimisation strategy is mostly adopted from the underlying DSF. This is an important advantage. \citet{zheng2018dags} already state the importance of an easy to implement model; we only add 10 lines to their approximate 60 lines. Furthermore, we also noticed improvements in efficiency as D-Struct drastically reduces computation time compared to NOTEARS despite the ensemble architecture (see \cref{app:additional-experiments:computation}). 

\begin{figure*}[t]
    \centering
    \begin{subfigure}[t]{0.95\textwidth}
    \centering
    \resizebox{\textwidth}{!}{\begin{tikzpicture}
        \pgfplotsset{footnotesize,samples=10}
        
        \begin{groupplot}[group style = {group size = 4 by 1, horizontal sep = 20pt}, width = 6.0cm, height = 4.0cm]
            
            \nextgroupplot[
                legend style = { 
                    column sep = 5pt,
                    legend columns = 4, 
                    legend to name = grouplegend, 
                    anchor=north, 
                    scale=1.4
                },
                xlabel={$n$},
                x label style={at={(axis description cs:0.5,-0.13)},anchor=north},
                ylabel={SHD},
                ylabel near ticks,
                ymin=2, ymax=5,
                xtick={200,500,1000,2000},
                ymajorgrids=true,
                grid style=dashed,
            ]
            \addplot[color=FireBrick,mark=diamond,very thick]
            coordinates {(200,3.6) (500, 3.2) (1000, 2.75) (2000,2.66)};
            \addlegendentry[Black]{D-Struct [ER]}
                
            \addplot[color=blue,mark=o,very thick]
            coordinates {(200, 4.20) (500, 3.94) (1000, 3.67) (2000,3.54)};
            \addlegendentry[Black]{NOTEARS [ER]}

            \nextgroupplot[ 
                legend style = { 
                    column sep = 5pt, 
                    legend columns = -1, 
                    legend to name = grouplegend, anchor=west
                    },
                xlabel={$sd$},
                x label style={at={(axis description cs:0.5,-0.11)},anchor=north},
                ymin=1, ymax=10,
                xtick={0.5,1,1.5,2},
                ymajorgrids=true,
                grid style=dashed,
            ]
            \addplot[color=FireBrick,mark=diamond,very thick]
            coordinates {(0.5,3.75) (1, 3.5) (1.5,3) (2, 2.28)};
            \addlegendentry[Black]{D-Struct [ER]}
            
            \addplot[color=blue,mark=o,very thick]
            coordinates {(0.5, 7.33) (1, 7.66) (1.5, 5.67) (2, 3.66)};
            \addlegendentry[Black]{NOTEARS [ER]}
            
            

            \nextgroupplot[ 
                legend style = { 
                    column sep = 5pt, 
                    legend columns = -1, 
                    legend to name = grouplegend, anchor=west
                    },
                xlabel={$d$},
                x label style={at={(axis description cs:0.5,-0.11)},anchor=north},
                ymin=0, ymax=35,
                xtick={5,7,10},
                ymajorgrids=true,
                grid style=dashed,
            ]
            \addplot[color=FireBrick,mark=diamond,very thick]
            coordinates {(5,2.28) (7,8.67)  (10, 19.71)};
            \addlegendentry[Black]{D-Struct}
            
            \addplot[color=blue,mark=o,very thick]
            coordinates {(5, 3.67) (7,12.88) (10, 30.8)};
            \addlegendentry[Black]{NOTEARS}

            \nextgroupplot[
                legend style = { column sep = 10pt, legend columns = 4, legend to name = grouplegend,},
                xlabel={$K$},
                x label style={at={(axis description cs:0.5,-0.11)},anchor=north},
                 ymin=1, ymax=6,
                xtick={2,3,5},
                ymajorgrids=true,
                grid style=dashed,
            ]
            \addplot[color=FireBrick,mark=diamond,very thick]
            coordinates {(2,1.25)  (3, 2.29) (5,4)};
            \addlegendentry[Black]{D-Struct [ER]}
                    
            \addplot[color=blue,mark=o,very thick]
            coordinates {(2, 4.333)  (3, 3.667) (5,5.667)};
            \addlegendentry[Black]{NOTEARS [ER]}

        \end{groupplot}

        \draw [
            very thick,
            decoration={
                brace,
                raise=-.15cm,
            },
            decorate
        ] ($(group c1r1.north west) + (-.3, .5)$) -- ($(group c1r1.north east) + (.2, .5)$) node [pos=0.5,anchor=center,yshift=.3cm] {\Large \bf varying $n$};

        \draw [
            very thick,
            decoration={
                brace,
                raise=-.15cm,
            },
            decorate
        ] ($(group c2r1.north west) + (-.3, .5)$) -- ($(group c2r1.north east) + (.2, .5)$) node [pos=0.5,anchor=center,yshift=.3cm] {\Large \bf varying $s$};
        
        \draw [
            very thick,
            decoration={
                brace,
                raise=-.15cm,
            },
            decorate
        ] ($(group c3r1.north west) + (-.3, .5)$) -- ($(group c3r1.north east) + (.2, .5)$) node [pos=0.5,anchor=center,yshift=.3cm] {\Large \bf varying $d$};

        \draw [
            very thick,
            decoration={
                brace,
                raise=-.15cm,
            },
            decorate
        ] ($(group c4r1.north west) + (-.3, .5)$) -- ($(group c4r1.north east) + (.2, .5)$) node [pos=0.5,anchor=center,yshift=.3cm] {\Large \bf varying $K$};

        \draw [
            very thick,
            decoration={
                brace,
                raise=-.15cm,
            },
            decorate
        ] ($(group c4r1.north east) + (.5, .2)$) -- ($(group c4r1.south east) + (.5, -.5)$) node [pos=0.5,anchor=center,xshift=.4cm,rotate=-90] {\Large \bf Erdos-Renyì};

    \end{tikzpicture}}
    \end{subfigure}

    \begin{subfigure}[t]{0.95\textwidth}
    \centering
    \resizebox{\textwidth}{!}{\begin{tikzpicture}
        \pgfplotsset{footnotesize,samples=10}
        
        \begin{groupplot}[group style = {group size = 4 by 1, horizontal sep = 20pt}, width = 6.0cm, height = 4.0cm]
            
            \nextgroupplot[
                legend style = { 
                    column sep = 5pt,
                    legend columns = 4, 
                    legend to name = grouplegend, 
                    anchor=north, 
                    scale=1.4
                },
                xlabel={$n$},
                x label style={at={(axis description cs:0.5,-0.13)},anchor=north},
                ylabel={SHD},
                ylabel near ticks,
                ymin=0.5, ymax=8,
                xtick={200,500,1000,2000},
                ymajorgrids=true,
                grid style=dashed,
            ]

            \addplot[color=FireBrick,mark=diamond,very thick]
            coordinates {(200, 2.8) (500, 2.2) (1000, 2.75) (2000,1)};
            \addlegendentry[Black]{D-Struct [SF]}
                
            \addplot[color=blue,mark=o,very thick]
            coordinates {(200, 6.2) (500, 7.2) (1000, 5.3) (2000,3)};
            \addlegendentry[Black]{NOTEARS [SF]}
            
            \nextgroupplot[ 
                legend style = { 
                    column sep = 5pt, 
                    legend columns = -1, 
                    legend to name = grouplegend, anchor=west
                    },
                xlabel={$sd$},
                x label style={at={(axis description cs:0.5,-0.11)},anchor=north},
                ymin=1, ymax=10,
                xtick={0.5,1,1.5,2},
                ymajorgrids=true,
                grid style=dashed,
            ]

            \addplot[color=FireBrick,mark=diamond,very thick]
            coordinates {(0.5, 3.33) (1, 3.33) (1.5,3.25)(2,2.75)};
            \addlegendentry[Black]{D-Struct [SF]}
            
            \addplot[color=blue,mark=o,very thick]
            coordinates {(0.5, 8) (1, 8) (1.5,7.6)(2, 5)};
            \addlegendentry[Black]{NOTEARS [SF]}

            \nextgroupplot[ 
                legend style = { 
                    column sep = 5pt, 
                    legend columns = -1, 
                    legend to name = grouplegend, anchor=west
                    },
                xlabel={$d$},
                x label style={at={(axis description cs:0.5,-0.11)},anchor=north},
                ymin=0, ymax=35,
                xtick={5,7,10},
                ymajorgrids=true,
                grid style=dashed,
            ]
            \addplot[color=FireBrick,mark=diamond,very thick]
            coordinates {(5,2.75) (7,8.22) (10, 19.71)};
            \addlegendentry[Black]{D-Struct}
            
            \addplot[color=blue,mark=o,very thick]
            coordinates {(5, 3.813) (7,15.56) (10, 30.8)};
            \addlegendentry[Black]{NOTEARS}

            \nextgroupplot[
                legend style = { column sep = 10pt, legend columns = 4, legend to name = grouplegend,},
                xlabel={$K$},
                x label style={at={(axis description cs:0.5,-0.11)},anchor=north},
                 ymin=0, ymax=10,
                xtick={2,3,5},
                ymajorgrids=true,
                grid style=dashed,
            ]

            \addplot[color=FireBrick,mark=diamond,very thick]
            coordinates {(2,2.4)(3, 2) (5, 0.75)};
            \addlegendentry[Black]{D-Struct(\texttt{NOTEARS-MLP})}

            \addplot[color=blue,mark=o,very thick]
            coordinates {(2,6.5) (3, 5.33) (5, 7.66)};
            \addlegendentry[Black]{\texttt{NOTEARS-MLP}}
                    
        \end{groupplot}
        \node at ($(group c2r1)!0.5!(group c3r1) + (0cm, -2.3cm)$) {\ref{grouplegend}};
        
        \draw [
            very thick,
            decoration={
                brace,
                raise=-.15cm,
            },
            decorate
        ] ($(group c4r1.north east) + (.5, .3)$) -- ($(group c4r1.south east) + (.5, -.5)$) node [pos=0.5,anchor=center,xshift=.4cm,rotate=-90] {\Large \bf Scale-Free};
    \end{tikzpicture}}
    \end{subfigure}

    \vspace{-3pt}
    \caption{{\bf Structure recovery.} We report the SHD ($\downarrow$) compared to the true graph. We report performance as a function of four different parameters (changing the properties of the task). D-Struct outperforms \texttt{NOTEARS-MLP} in all these settings. Additional results are reported in \cref{app:additional-experiments}. Unless otherwise indicated, $n=1000, d=5, K=3, s=2d$.}
    \label{fig:sensitivity}
    \vspace{-3pt}
    \rule{\linewidth}{.75pt}
    \vspace{-15pt}
\end{figure*}
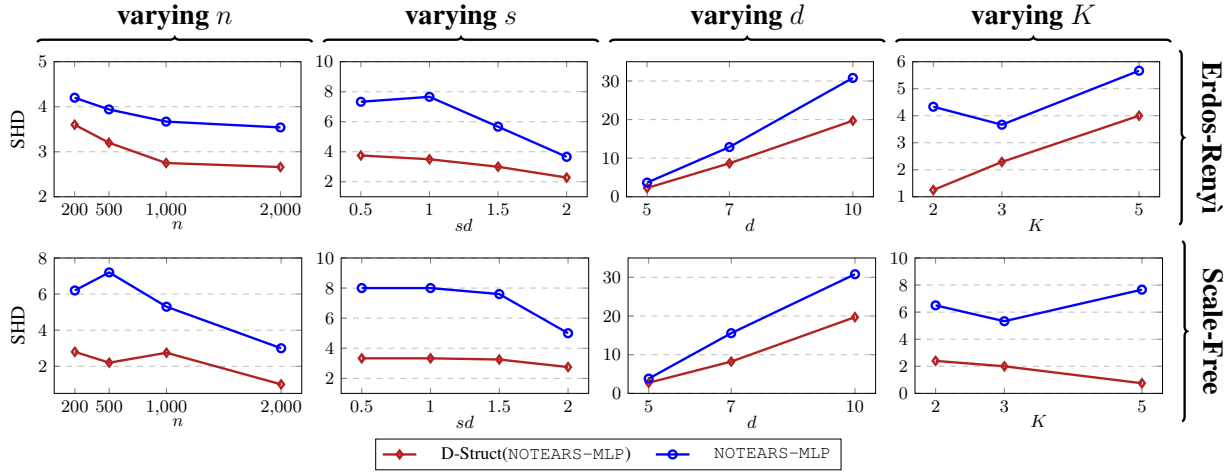

\vspace{-10pt}
\section{Experiments} \label{sec:experiments}
\vspace{-5pt}

Recall from \cref{sec:method} that D-Struct's objective is to transform a dataset into a DAG, whilst remaining differentiable. With D-Struct, our aim is to increase performance of any DSF by enforcing transportability on the learner's outcome structure. As such, the most pressing questions are: {\it (1) Are the discovered structures transportable?}, {\it (2) Does D-Struct improve existing learners?}, and {\it (3) Do we really need our subsampling routine?} We answer these questions one-by-one below with empirical validation.

However, before we answer these questions, we would also like to point to \cref{app:additional-experiments} which answers (many) more questions, such as: {\it Does D-Struct pay for accuracy with computation?} (\cref{app:additional-experiments:computation}) {\it What about different threshold values?} (\cref{app:additional-experiments:threshold}) {\it Does D-Struct also work with other DSFs?} (\cref{app:additional_results:dsfs}) {\it What if we don't use the subsampling routine?} (\cref{app:additional-experiments:subsampling}) {\it Does it also work with two datasets?} (\cref{app:additional_results:multiset}), and so forth. Furthermore, we only present a snapshot of the experimental results in the main text. For almost all experiments, we have included a ``completed'' set in the relevant appendices.

\begin{figure}
\vspace{-10pt}
    \begin{subfigure}{.35\linewidth}
        \centering
        
        \begin{tikzpicture}[
            vert/.style = {circle, inner sep=0, minimum size =5mm, fill=none, draw=black}
        ]

            \coordinate (true) at (0, 0);
    
            \node[vert] (x0) at ($(true) + .7*(0, 1)$) {\small $A$};
            \node[vert] (x1) at ($(true) + .7*(0.951, 0.309)$) {\small $B$};
            \node[vert] (x2) at ($(true) + .7*(0.587, -0.809)$) {\small $C$};
            \node[vert] (x3) at ($(true) + .7*(-0.587, -0.809)$) {\small $D$};
            \node[vert] (x4) at ($(true) + .7*(-0.95, 0.309)$) {\small $E$};


            \draw[-latex] (x0) -- (x4);

            \draw[-latex] (x1) -- (x0);
            \draw[-latex] (x1) -- (x2);
            
            \draw[-latex] (x2) -- (x0);
            \draw[-latex] (x2) -- (x4);
            
            \draw[-latex] (x3) -- (x0);
            \draw[-latex] (x3) -- (x1);
            \draw[-latex] (x3) -- (x2);
            \draw[-latex] (x3) -- (x4);

            \node at ($(true) + (0, -1.1)$) {True graph};

        \end{tikzpicture}
        
    \end{subfigure}%
    ~
    \begin{subfigure}{.6\linewidth}

    \begin{tikzpicture}[
            vert/.style = {circle, inner sep=0, minimum size =3mm, fill=none, draw=black},
            missing_edge/.style = {dash pattern={on 1pt off .8pt}}
        ]

            \coordinate (notears_m) at (0, 0);
    
            \node[vert] (x0) at ($(notears_m) + .5*(0, 1)$) {\tiny $A$};
            \node[vert] (x1) at ($(notears_m) + .5*(0.951, 0.309)$) {\tiny $B$};
            \node[vert] (x2) at ($(notears_m) + .5*(0.587, -0.809)$) {\tiny $C$};
            \node[vert] (x3) at ($(notears_m) + .5*(-0.587, -0.809)$) {\tiny $D$};
            \node[vert] (x4) at ($(notears_m) + .5*(-0.95, 0.309)$) {\tiny $E$};


            \draw[-latex] (x1) -- (x0);
            \draw[-latex] (x1) -- (x2);
            \draw[-latex, color=FireBrick] (x1) -- (x4);
            
            \draw[-latex] (x2) -- (x0);

            \draw[-latex] (x3) -- (x0);
            \draw[-latex] (x3) -- (x2);
            \draw[-latex] (x3) -- (x4);

            
            \draw[-latex, missing_edge, color=FireBrick] (x0) -- (x4);
            \draw[-latex, missing_edge, color=FireBrick] (x2) -- (x4);
            \draw[-latex, missing_edge, color=FireBrick] (x3) -- (x1);

            \node at ($(notears_m) + (0, -.8)$) {Mean NT};

            \coordinate (notears_1) at ($(notears_m) + (2, 0)$);
    
            \node[vert] (x0) at ($(notears_1) + .5*(0, 1)$) {\tiny $A$};
            \node[vert] (x1) at ($(notears_1) + .5*(0.951, 0.309)$) {\tiny $B$};
            \node[vert] (x2) at ($(notears_1) + .5*(0.587, -0.809)$) {\tiny $C$};
            \node[vert] (x3) at ($(notears_1) + .5*(-0.587, -0.809)$) {\tiny $D$};
            \node[vert] (x4) at ($(notears_1) + .5*(-0.95, 0.309)$) {\tiny $E$};


             \draw[-latex] (x1) -- (x0);
             \draw[-latex] (x1) -- (x2);
             \draw[-latex] (x1) -- (x3);
             \draw[-latex] (x1) -- (x4);

             \draw[-latex] (x2) -- (x0);

             \draw[-latex] (x3) -- (x0);
             \draw[-latex] (x3) -- (x2);
             \draw[-latex] (x3) -- (x4);

             \draw[-latex] (x4) -- (x0);

             \draw[-latex, color=blue, missing_edge] (x2) -- (x4);

            \node at ($(notears_1) + (0, -.8)$) {NT 1};

            \coordinate (notears_2) at ($(notears_1) + (1.5, 0)$);
    
            \node[vert] (x0) at ($(notears_2) + .5*(0, 1)$) {\tiny $A$};
            \node[vert] (x1) at ($(notears_2) + .5*(0.951, 0.309)$) {\tiny $B$};
            \node[vert] (x2) at ($(notears_2) + .5*(0.587, -0.809)$) {\tiny $C$};
            \node[vert] (x3) at ($(notears_2) + .5*(-0.587, -0.809)$) {\tiny $D$};
            \node[vert] (x4) at ($(notears_2) + .5*(-0.95, 0.309)$) {\tiny $E$};


            \draw[-latex, color=blue] (x0) -- (x4);

            \draw[-latex] (x1) -- (x0);
            \draw[-latex] (x1) -- (x2);
            \draw[-latex] (x1) -- (x4);

            \draw[-latex] (x2) -- (x0);
            \draw[-latex] (x2) -- (x4);

            \draw[-latex] (x3) -- (x0);
            \draw[-latex, color=blue] (x3) -- (x1);
            \draw[-latex] (x3) -- (x2);
            \draw[-latex] (x3) -- (x4);

            \node at ($(notears_2) + (0, -.8)$) {NT 2};


            \coordinate (dstruct_m) at ($(notears_m) + (0, -1.8)$);
    
            \node[vert] (x0) at ($(dstruct_m) + .5*(0, 1)$) {\tiny $A$};
            \node[vert] (x1) at ($(dstruct_m) + .5*(0.951, 0.309)$) {\tiny $B$};
            \node[vert] (x2) at ($(dstruct_m) + .5*(0.587, -0.809)$) {\tiny $C$};
            \node[vert] (x3) at ($(dstruct_m) + .5*(-0.587, -0.809)$) {\tiny $D$};
            \node[vert] (x4) at ($(dstruct_m) + .5*(-0.95, 0.309)$) {\tiny $E$};


            \draw[-latex] (x1) -- (x0);
            \draw[-latex] (x1) -- (x2);
            \draw[-latex, color=FireBrick] (x1) -- (x4);

            \draw[-latex] (x2) -- (x0);
            \draw[-latex] (x2) -- (x4);

            \draw[-latex] (x3) -- (x0);
            \draw[-latex] (x3) -- (x1);
            \draw[-latex] (x3) -- (x2);
            \draw[-latex] (x3) -- (x4);

            \draw[-latex, color=FireBrick] (x4) -- (x0);

            \node at ($(dstruct_m) + (0, -.8)$) {Mean DS};

            \coordinate (dstruct_1) at ($(dstruct_m) + (2, 0)$);
    
            \node[vert] (x0) at ($(dstruct_1) + .5*(0, 1)$) {\tiny $A$};
            \node[vert] (x1) at ($(dstruct_1) + .5*(0.951, 0.309)$) {\tiny $B$};
            \node[vert] (x2) at ($(dstruct_1) + .5*(0.587, -0.809)$) {\tiny $C$};
            \node[vert] (x3) at ($(dstruct_1) + .5*(-0.587, -0.809)$) {\tiny $D$};
            \node[vert] (x4) at ($(dstruct_1) + .5*(-0.95, 0.309)$) {\tiny $E$};

            
            \draw[-latex] (x1) -- (x0);
            \draw[-latex] (x1) -- (x2);
            \draw[-latex] (x1) -- (x4);

            \draw[-latex] (x2) -- (x0);
            \draw[-latex] (x2) -- (x4);

            \draw[-latex] (x3) -- (x0);
            \draw[-latex] (x3) -- (x1);
            \draw[-latex] (x3) -- (x2);
            \draw[-latex] (x3) -- (x4);

            \draw[-latex] (x4) -- (x0);

            \node at ($(dstruct_1) + (0, -.8)$) {DS 1};

            \coordinate (dstruct_2) at ($(dstruct_1) + (1.5, 0)$);
    
            \node[vert] (x0) at ($(dstruct_2) + .5*(0, 1)$) {\tiny $A$};
            \node[vert] (x1) at ($(dstruct_2) + .5*(0.951, 0.309)$) {\tiny $B$};
            \node[vert] (x2) at ($(dstruct_2) + .5*(0.587, -0.809)$) {\tiny $C$};
            \node[vert] (x3) at ($(dstruct_2) + .5*(-0.587, -0.809)$) {\tiny $D$};
            \node[vert] (x4) at ($(dstruct_2) + .5*(-0.95, 0.309)$) {\tiny $E$};

            
            \draw[-latex] (x1) -- (x0);
            \draw[-latex] (x1) -- (x2);
            \draw[-latex] (x1) -- (x4);

            \draw[-latex] (x2) -- (x0);
            \draw[-latex] (x2) -- (x4);

            \draw[-latex] (x3) -- (x0);
            \draw[-latex] (x3) -- (x1);
            \draw[-latex] (x3) -- (x2);
            \draw[-latex] (x3) -- (x4);

            \draw[-latex] (x4) -- (x0);

            \node at ($(dstruct_2) + (0, -.8)$) {DS 2};

        \end{tikzpicture}
        
    \end{subfigure}

    \vspace{-3pt}
    \caption{{\bf Evaluating transportability.} We ran two NOTEARS-MLP (NT) and one D-Struct (DS) with $K=2$ on the experimental setting in {\bf (1)}. \textcolor{FireBrick}{\bf Red} indicates violations w.r.t. to the true DAG, and \textcolor{blue}{\bf blue} indicates violations across subgraphs. Dashed edges were missing w.r.t. the comparison DAG. We observe a smaller SHD in both the subgraphs (SHD=0) and the mean graph (SHD=2) by DS, while the DAGs discovered by DS are perfectly transportable, unlike NT which are not (NT-mean SHD=4, NT-subs SHD=3).}
    \label{fig:res:transportability}
    \rule{\linewidth}{.5pt}
    \vspace{-30pt}
\end{figure}
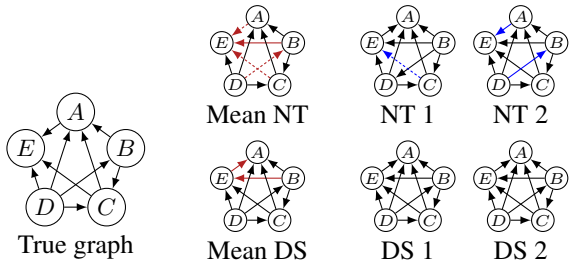

\textbf{(1) Transportability.} Before testing accuracy, we first empirically confirm that NOTEARS is not transportable while D-Struct is. We compare NOTEARS with D-Struct using 1000 samples drawn from an Erdos-Renyì (ER) random graph, and split the samples into two equal-sized subsets. 
We evaluate the structural Hamming distance (SHD) between the graphs learned by NOTEARS on each dataset, and the same for the internal graphs learned by D-Struct. The DAGs learnt by D-Struct are perfectly transportable (SHD$=0$) in 8/10 runs (mean SHD $0.46\pm0.27$), with only minor discrepancies in the other cases. Conversely, NOTEARS has a mean SHD of $1.14\pm0.20$, only displaying transportability in 2 cases. Similar results for other DSFs are reported in \cref{app:additional-experiments}. 
In \cref{app:notearsdags}, we extend this experiment to more DAGs, though our conclusion remains the same. For illustration, we have also included some of the DAGs reported in our \cref{app:notearsdags} in \cref{fig:res:transportability}.

From \cref{fig:res:transportability}, it is clear that neglecting transportability leads to conflicting results and increased SHD. Furthermore, the subgraphs in D-Struct are perfectly transportable (SHD=0). More instances of this experiment can be found in \cref{app:notearsdags}, or by running our code (cfr. \cref{app:experiment-details}). 

{\bf (2) Accuracy.} The most straightforward way to see if D-Struct is better is by repeating the experiments in \citet{zheng2020learning}. We report only a subset of our outcomes in the main text, mainly on D-Struct's improvement over \texttt{NOTEARS-MLP}. However, more metrics and experiments on different DSFs can be found in \cref{app:additional-experiments}. In \cref{tab:results} we report the false positive rate (FPR), true positive rate (TPR), false discovery rate (FDR), and structural Hamming distance (SHD) of the estimated DAGs using data sampled from different ER random graphs with varying sample size ($n$), expected number of edges ($s$), and dimension count ($d$). In all cases, we find that D-Struct significantly improves \texttt{NOTEARS-MLP} (other DSFs in \cref{app:additional-experiments}). A similar in \cref{fig:sensitivity} which reports the SHD for more parameters and data from Erdos-Reny\`i as well as Scale-Free graphs \citep{zheng2018dags}.

\begin{table}
    
    \centering
    \caption{{\bf Usefulness of our subsampling routine.} We sample ten different ER graphs like in \cref{tab:results}. From each system, we sample $n=2000$ samples and evaluate \texttt{NOTEARS-MLP} {\it with} ({``\cmark''}) our subsampling routine from \cref{sec:method:single-origin} and {\it without} ({``\xmark''}) the subsampling routine, using random splits instead. Each row repeats our experiment with different $K$. We report the average (and std) performance in terms of the SHD.}
    \vspace{-5pt}
    \label{tab:results:subsample}
    \begin{tabularx}{.4\textwidth}{r  *{1}{|CC}}
    \toprule
        {\it metric} & 
        
        \multicolumn{2}{c}{\bf SHD ($\downarrow$)}  \\
        
        \midrule
        
        {\it Subsample}& 
        \cmark & \xmark  \\
        
        \toprule

     $K$& \multicolumn{2}{c}{\it varying amount of splits}\\
        \midrule
        $2$&
        {\footnotesize \bf 2.80}{\scriptsize $\pm$0.53}  &   {\footnotesize 3.40}{\scriptsize $\pm$0.58}\\

        $3$&
        {\footnotesize \bf 3.00}{\scriptsize $\pm$0.37}  &   {\footnotesize 4.00}{\scriptsize $\pm$0.59}\\

        $5$&
        {\footnotesize \bf 2.80}{\scriptsize $\pm$0.57}  &   {\footnotesize 4.40}{\scriptsize $\pm$1.29}\\

    \bottomrule
    \end{tabularx}
    \vspace{-20pt}
\end{table}


{\bf (3) Subset construction.}
A final property we wish to validate is the need for sampling $K$ different subsets using our subsampling routine from \cref{sec:method:single-origin}. This is an important validation as it shows that D-Struct does not {\it only} gain in performance due to its ensemble architecture. For this, we compare D-Struct's performance {\it with} and {\it without} our subsampling routine. Using D-Struct without our subsampling routine amounts to providing $K$ random splits, rather than carefully sampling $K$ distinct $\mathcal{D}_k \sim \mathbb{P}^k$. \Cref{tab:results:subsample} shows that our subsampling routine {\it does} improve D-Struct's performance as expected, validating our goal to explicitly optimise for transportable structure learners.

We believe that these experiments confirm that D-Struct can help us create useful structure learners. 
In \cref{app:experiment-details}, we include a link to our code repository, encouraging readers to reproduce our results, as well as provide hyperparameter ablations and settings.

\vspace{-8pt}
\section{Discussion}
\vspace{-5pt}
D-Struct advances differentiable structure learning by introducing transportability, a property guaranteed by CIT-based methods. 
We show empirically that enforcing this property substantially improves the performance of a range of DSFs.
We believe D-Struct can have a positive impact on architectures and problems relying on differentiable structure learners, as well as on general scientific data analysis.

{\bf Relating DSFs to causality.} As pointed out by \citet{kaiser2022unsuitability} and \citet{reisach2021beware}, DSFs are often wrongly used to recover a {\it causal} DAG. While DAGs are indeed the model of choice to model causality, there is currently no guarantee that a DAG discovered using any DSF can be identified (and thus used) as such. With this, we wish to state explicitly that a DSF's output is {\it not} to be interpreted as a causal model (see \cref{app:causality} for more discussion).

{\bf Future work.} The inability to recover causal structure is a consequence of there existing many more useful properties stemming from a CIT-based approach (multiple books concern this very topic, e.g. \citet{koller2009probabilistic,pearl1988probabilistic,jordan1999learning,lauritzen1996graphical}). Bridging the gap between these methods is a clear path forward, hopefully increasing differentiable structure learners' potential even further. Specifically, using structure learners to uncover a causal structure from observational data requires stricter assumptions. As such, one particularly interesting avenue of future work is to allow DSFs (not only D-Struct) to adhere to some of these assumptions and use them to guarantee causal discovery, taking DSFs to the next level.

Finally, D-Struct is only the {\it first step} of scientific discovery. Like other DSFs, D-Struct {\it suggests} a link between variables, the scientist should still confirm this link in the lab.

\section*{Acknowledgements}
We thank our funding agencies:
Jeroen Berrevoets is funded by the W.D. Armstrong Trust. Nabeel Seedat is funded by The Cystic Fibrosis Trust. Fergus Imrie is funded by an NSF grant (1722516).

We would also like to thank our reviewers and labmates at the vanderschaar-lab (\url{https://vanderschaar-lab.com}) for their helpful suggestions.

\bibliography{example_paper}
\bibliographystyle{unsrtnat}

\newpage
\onecolumn

\appendix

\addcontentsline{toc}{section}{Appendix}
\part{Appendix: D-Struct}
\parttoc

\section{Additional experiments} \label{app:additional-experiments}

Please find our online code repository at:
\begin{mdframed}
\centering
\url{https://github.com/jeroenbe/d-struct}\\
or\\
\url{https://github.com/vanderschaarlab}
\end{mdframed}
Our code is based on code provided by \citet{zheng2020learning}, and we annotated our code where we used their implementation.

\subsection{Settings and details} \label{app:experiment-details}

In the interest of space, we left out a few details in our main text. Here we discuss hyperparameters (those in addition to the hyperparameters required for the selected DSFs), the evaluation metrics, and how we combine the different parallel DAGs.

{\bf Hyperparameters.} D-Struct inherits hyperparameters from the chosen underlying DSFs. These hyperparameters act in the same way as they would in their original incarnation. For a discussion on these hyperparameters, we refer to the relevant literature on these methods specifically.

However, D-Struct also adds two additional parameters: $K$ and $\alpha$. The impact of $K$ is already discussed in the main text, recapitulated as: $K$ implicitly determines the sizes of the subsets used to train the parallel DSFs, as such, {\it for high $K$ we should have high $n$}. With both increasing, we report better performance (particularly in Scale-Free DAGs).

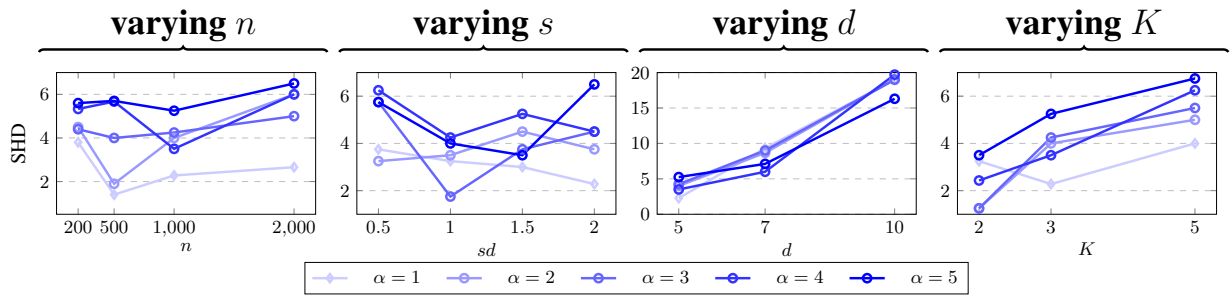
\begin{figure}[h]
    \centering
    \begin{subfigure}[t]{0.95\textwidth}
    \centering
    \resizebox{\textwidth}{!}{\begin{tikzpicture}
        \pgfplotsset{footnotesize,samples=10}
        
        \begin{groupplot}[group style = {group size = 4 by 1, horizontal sep = 20pt}, width = 6.0cm, height = 4.0cm]
            
            \nextgroupplot[
                legend style = { 
                    column sep = 5pt,
                    legend columns = 4, 
                    legend to name = grouplegend, 
                    anchor=north, 
                    scale=1.4
                },
                xlabel={$n$},
                x label style={at={(axis description cs:0.5,-0.15)},anchor=north},
                ylabel={SHD},
                ylabel near ticks,
                ymin=0.5, ymax=7,
                xtick={200,500,1000,2000},
                ymajorgrids=true,
                grid style=dashed,
            ]
            \addplot[color=blue!20,mark=diamond,very thick]
            coordinates {(200,3.8) (500,1.4 ) (1000, 2.28) (2000,2.66)};
            \addlegendentry[Black]{$\alpha=1$}
                
            \addplot[color=blue!40,mark=o,very thick]
            coordinates {(200, 4.5) (500, 1.9) (1000, 4) (2000,6)};
            \addlegendentry[Black]{$\alpha=2$}
            
            \addplot[color=blue!60,mark=o,very thick]
            coordinates {(200, 4.4) (500, 4) (1000, 4.25) (2000,5)};
            \addlegendentry[Black]{$\alpha=3$}
            
            \addplot[color=blue!80,mark=o,very thick]
            coordinates {(200, 5.33) (500, 5.67) (1000, 3.5) (2000,6)};
            \addlegendentry[Black]{$\alpha=4$}
            
            \addplot[color=blue,mark=o,very thick]
            coordinates {(200, 5.6) (500, 5.7) (1000, 5.25) (2000,6.5)};
            \addlegendentry[Black]{$\alpha=5$}
                
                
            
            \nextgroupplot[ 
                legend style = { 
                    column sep = 5pt, 
                    legend columns = -1, 
                    legend to name = grouplegend, anchor=west
                    },
                xlabel={$sd$},
                x label style={at={(axis description cs:0.5,-0.15)},anchor=north},
                ymin=1, ymax=7,
                xtick={0.5,1,1.5,2},
                ymajorgrids=true,
                grid style=dashed,
            ]

             \addplot[color=blue!20,mark=diamond,very thick]
            coordinates {(0.5,3.75) (1,3.25) (1.5, 3) (2,2.28)};
            \addlegendentry[Black]{$\alpha=1$}
                
            \addplot[color=blue!40,mark=o,very thick]
             coordinates {(0.5,3.25) (1,3.5 ) (1.5, 4.5) (2,3.75)};
            \addlegendentry[Black]{$\alpha=2$}
            
            \addplot[color=blue!60,mark=o,very thick]
             coordinates {(0.5,5.75) (1,1.75 ) (1.5, 3.75) (2,4.5)};
            \addlegendentry[Black]{$\alpha=3$}
            
            \addplot[color=blue!80,mark=o,very thick]
             coordinates {(0.5,6.25) (1,4.25 ) (1.5, 5.25) (2,4.5)};
            \addlegendentry[Black]{$\alpha=4$}
            
            \addplot[color=blue,mark=o,very thick]
             coordinates {(0.5,5.75) (1,4) (1.5, 3.5) (2,6.5)};
            \addlegendentry[Black]{$\alpha=5$}
            
            

            \nextgroupplot[ 
                legend style = { 
                    column sep = 5pt, 
                    legend columns = -1, 
                    legend to name = grouplegend, anchor=west
                    },
                xlabel={$d$},
                x label style={at={(axis description cs:0.5,-0.15)},anchor=north},
                ymin=0, ymax=20,
                xtick={5,7,10},
                ymajorgrids=true,
                grid style=dashed,
            ]
            \addplot[color=blue!20,mark=diamond,very thick]
            coordinates {(5,2.28) (7,9.33) (10, 19)};
            \addlegendentry[Black]{$\alpha=1$}
                
            \addplot[color=blue!40,mark=o,very thick]
            coordinates {(5,4) (7,8.67) (10, 19)};
            \addlegendentry[Black]{$\alpha=2$}
            
            \addplot[color=blue!60,mark=o,very thick]
            coordinates {(5,4.25) (7,9) (10, 19)};
            \addlegendentry[Black]{$\alpha=3$}
            
            \addplot[color=blue!80,mark=o,very thick]
            coordinates {(5,3.5) (7,6) (10, 19.71)};
            \addlegendentry[Black]{$\alpha=4$}
            
            \addplot[color=blue,mark=o,very thick]
            coordinates {(5,5.25) (7,7.11) (10, 16.3)};
            \addlegendentry[Black]{$\alpha=5$}
            
            \nextgroupplot[
                legend style = { column sep = 10pt, legend columns = 5, legend to name = grouplegend,},
                xlabel={$K$},
                x label style={at={(axis description cs:0.5,-0.15)},anchor=north},
                 ymin=1, ymax=7,
                xtick={2,3,5},
                ymajorgrids=true,
                grid style=dashed,
            ]
             \addplot[color=blue!20,mark=diamond,very thick]
            coordinates {(2,3.25) (3,2.28 ) (5, 4)};
            \addlegendentry[Black]{$\alpha=1$}
                
            \addplot[color=blue!40,mark=o,very thick]
            coordinates {(2,1.25) (3,4 ) (5, 5)};
            \addlegendentry[Black]{$\alpha=2$}
            
            \addplot[color=blue!60,mark=o,very thick]
            coordinates {(2,1.25) (3,4.25 ) (5, 5.5)};
            \addlegendentry[Black]{$\alpha=3$}
            
            \addplot[color=blue!80,mark=o,very thick]
            coordinates {(2,2.43) (3,3.5 ) (5, 6.25)};
            \addlegendentry[Black]{$\alpha=4$}
            
            \addplot[color=blue,mark=o,very thick]
            coordinates {(2,3.5) (3,5.25 ) (5, 6.75)};
            \addlegendentry[Black]{$\alpha=5$}

        \end{groupplot}
        
        \node at ($(group c2r1)!0.5!(group c3r1) + (0cm, -2.3cm)$) {\ref{grouplegend}};

        \draw [
            very thick,
            decoration={
                brace,
                raise=-.15cm,
            },
            decorate
        ] ($(group c1r1.north west) + (-.3, .5)$) -- ($(group c1r1.north east) + (.2, .5)$) node [pos=0.5,anchor=center,yshift=.3cm] {\LARGE \bf varying $n$};

        \draw [
            very thick,
            decoration={
                brace,
                raise=-.15cm,
            },
            decorate
        ] ($(group c2r1.north west) + (-.3, .5)$) -- ($(group c2r1.north east) + (.2, .5)$) node [pos=0.5,anchor=center,yshift=.3cm] {\LARGE \bf varying $s$};
        
        \draw [
            very thick,
            decoration={
                brace,
                raise=-.15cm,
            },
            decorate
        ] ($(group c3r1.north west) + (-.3, .5)$) -- ($(group c3r1.north east) + (.2, .5)$) node [pos=0.5,anchor=center,yshift=.3cm] {\LARGE \bf varying $d$};

        \draw [
            very thick,
            decoration={
                brace,
                raise=-.15cm,
            },
            decorate
        ] ($(group c4r1.north west) + (-.3, .5)$) -- ($(group c4r1.north east) + (.2, .5)$) node [pos=0.5,anchor=center,yshift=.3cm] {\LARGE \bf varying $K$};

    \end{tikzpicture}}
    \end{subfigure}

    \caption{{\bf Results showing the effect of $\alpha$.} Depending on the nature of the problem the degree of regularization imposed by $\alpha$ can vary. This then changes the amount we enforce the similarity between the different D-Struct adjacency. matrices.}
    \label{fig:sensitivity:alpha}
    \rule{\textwidth}{.5pt}
\end{figure}

The impact of $\alpha$ is a bit more subtle, and also a function of $K$. First, consider \cref{fig:sensitivity:alpha}, displaying the impact on each evaluation metric as a function of different $\alpha$. What we find is that setting $\alpha$ is mostly dependent on $K$ as lower $\alpha$ tend to work better with higher $K$, and vice versa for lower $K$. This makes sense as we sum each $\mathcal{L}_\textbf{MSE}$, resulting in a higher value with more $K$. If $\alpha$ is large in a setting with large $K$, the regularisation effect would simply be too large. We set our hyperparameters to those which yielded best performance (deduced from \cref{fig:sensitivity:alpha} for $\alpha$, and $K=3$ when not varied over as this yielded the most stable results overall).

{\bf Evaluation metrics.} The learned graphs from NOTEARS and D-Struct are assessed using four graph metrics namely: (1) Structural Hamming distance (SHD), (2) False discovery rate (FDR), (3) False positive rate (FPR) and (4) True positive rate (TPR). These values are standard when evaluating structure learning methods. We provide some insight into these evaluation metrics below.
\begin{description}[align=left, labelwidth=5.5cm, leftmargin=1.5cm]
\item[Structural Hamming distance (SHD)] SHD is the total number of edge additions, deletions, and reversals needed to convert the estimated DAG into the true DAG. That means that the worst case SHD is $d^2 - d$, as we bound the diagonal to be $0$ at all times. As such, the reported SHD with varying $d$ is expected to be higher, not due to hardness of the problem, but as a property of the SHD (see for example \cref{fig:sensitivity}).
\item[False discovery rate (FDR)] Whenever an edge is suggested in the estimated DAG, which is incorrect, we add to the falsely discovered edges. As such, the FDR is defined as the number of reversed edges and edges that should not exist, divided by the number of edges in total. Of course, the exception is when no edges are suggested at all (which implies dividing by $0$), which naturally has an FDR of zero.
\item[False positive rate (FPR)] We sum the edges that should have been reversed and those that should not exist, and divide by the total number of {\it non-edges} in the ground truth DAG. A non-edge is an edge that does not exist. With a more connected ground truth DAG, we expect this number to be lower automatically (as the numerator of the FPR would be higher). This is the reason why we let $s$ be a function of $d$, as increasing the number of expected edges with $d$ would somewhat counter this effect. Note that, in \cref{tab:results} we see the FPR increasing proportionate to the factor multiplied with $d$, which is as we would expect.
\item[True positive rate (TPR)] This signifies the number of correctly estimated edges, over the number of edges in the true graph. Note that, reversed edges are counted as wrong edges. 
\end{description}

{\bf Combining graphs.} Inference is done by combining the $K$ internal graphs. In our implementation of D-Struct, we combine graphs by averaging the adjacency matrices and apply a threshold to convert the average graph into a binary matrix. The latter is a similar strategy to most DSFs' strategies to convert a continuous matrix into a binary one. This is a relatively simple method with promising results, in line with what is currently done in the literature. 

However, given that D-Struct has multiple graphs, we can actually come up with different strategies (a potential topic for future research). Naturally, this would be more relevant with high $K$, which in turn requires a larger sample size, as per our discussion above. Specifically, we enter the domain of ensemble learning. Like D-Struct, ensemble methods need to combine, potentially conflicting, outcomes and provide the user with only one outcome. 

One avenue is to not vote on a per-element basis, but on a per-graph basis. Imagine, two graphs in $K$ that are exactly the same aspire more confidence in their accuracy. We could even relax similarity to an SHD across graphs, where we weigh each graph's ``vote'' proportionally to their combined SHD. We believe this to be a promising area of future research.

{\bf Experimental procedure.} Here we explain how our experimental setup works, which steps we need to perform before starting an experiment, and which information each model is provided.

There are two main parts to an experimental setup: (i) we need a structure, (ii) we need a set of structural equations accompanying the structure of step (i).

{\it (i) The structure.} In our setup, a structure can only be a DAG. To reduce bias as much as possible, we do not determine structures up front, but sample random structures for each experimental run. Of course, the same random structure is presented for each benchmark. Sampling random structures happens in two ways: either we sample a random Erd\"os-Renyi graph, which requires a dimension count ($d$), and an expected number of edges ($ds$); or we use a scale-free graph which is generated using the process described in \citet{barabasi1999emergence} as was also done in \citet{zheng2020learning}, which needs a parameter $\beta=1$ (the exponent for the preferential attachment process). The expected number of edges in our setup depends on $d$ such that $s$ resembles the ratio of edges versus non-edges in the random graph. 

{\it (ii) The equations.} With a sampled structure from (i), we can now sample some structural equations. In our paper, we use an index model to sample these. In short, an index model is randomly parameterised as: $f_j(X_{\text{pa}(j)}) = \sum_{m=1}^3 h_m(\sum_{k\in\text{pa}(j)}\theta_{jmk}X_k)$, where $h_1 = \tanh$, $h_2=\cos$, $h_3=\sin$, and each $\theta_{jmk}$ is drawn uniformly from range $[-2, -0.5] \cup [0.5, 2]$. Exactly as was reported in \citet{zheng2020learning}.

\subsection{Completed results}
Recall from \cref{sec:experiments} that we only reported a subset of the results. In \cref{tab:extra_results:nt-mlp-SF,tab:app:results:ER-MLP} we report the remainder for \texttt{NOTEARS-MLP} and the D-Struct implementation on scale-free graphs. 

\begin{table}[t]
    
    \centering
    \caption{{\bf Results on Erdos-Renyì (ER) graphs.} {\it First block:} We sample five different ER random graphs, and accompanying non-linear structural equations using an index-model. From each system, we then sample a varying number of samples and evaluate \texttt{NOTEARS-MLP} {\it with} D-Struct (indicated as {``\cmark''}) and {\it without} D-Struct (indicated as {``\xmark''}). {\it Second block:} For each row we sample a new ER graph with a varying degree of connectedness ($s$ indicates the expected number of edges). In both cases, we report the average performance in terms of SHD, FPR, TPR, and FDR, with std in scriptsize.}
    \label{tab:app:results:ER-MLP}
    \begin{tabularx}{\textwidth}{r  *{4}{|CC}}
    \toprule
        {\it metric} & 
        
        \multicolumn{2}{c}{\bf SHD ($\downarrow$)}  &
        \multicolumn{2}{c}{\bf FPR ($\downarrow$)}  &
        \multicolumn{2}{c}{\bf TPR ($\uparrow$)}    &
        \multicolumn{2}{c}{\bf FDR ($\downarrow$)}  \\
        
        \midrule
        
        {\it D-Struct}& 
        \cmark & \xmark & 
        \cmark & \xmark & 
        \cmark & \xmark & 
        \cmark & \xmark \\
        
        \toprule
        $n$& \multicolumn{8}{c}{\it varying sample size}\\
        \midrule
        $200$&
        {\footnotesize \bf 3.60}{\scriptsize $\pm$0.27}  &   {\footnotesize 4.20}{\scriptsize $\pm$0.44}&
        {\footnotesize \bf 2.00}{\scriptsize $\pm$0.67}  &   {\footnotesize 4.20}{\scriptsize $\pm$0.44}&
        {\footnotesize \bf 0.67}{\scriptsize $\pm$0.05}  &   {\footnotesize  0.64}{\scriptsize $\pm$0.05}&
        {\footnotesize \bf 0.25}{\scriptsize $\pm$0.06}  &   {\footnotesize 0.42}{\scriptsize $\pm$0.04}\\

        $500$&
        {\footnotesize \bf 3.20}{\scriptsize $\pm$0.80}  &   {\footnotesize 3.94}{\scriptsize $\pm$0.33}&
        {\footnotesize \bf 1.20}{\scriptsize $\pm$0.44}  &   {\footnotesize 3.94}{\scriptsize $\pm$0.33}&
        {\footnotesize \bf 0.66}{\scriptsize $\pm$0.24}  &   {\footnotesize 0.56}{\scriptsize $\pm$0.04}&
        {\footnotesize \bf 0.18}{\scriptsize $\pm$0.05}  &   {\footnotesize 0.44}{\scriptsize $\pm$0.04}\\

        $1000$&
        {\footnotesize \bf 2.75}{\scriptsize $\pm$0.47}  &   {\footnotesize 3.67}{\scriptsize $\pm$0.82}&
        {\footnotesize \bf 1.00}{\scriptsize $\pm$0.37}  &   {\footnotesize 2.67}{\scriptsize $\pm$0.63}&
        {\footnotesize \bf 0.75}{\scriptsize $\pm$0.08}  &   {\footnotesize  0.63}{\scriptsize $\pm$0.13}&
        {\footnotesize \bf 0.18}{\scriptsize $\pm$0.03}  &   {\footnotesize 0.39}{\scriptsize $\pm$0.11}\\

        \midrule
        $s$& \multicolumn{8}{c}{\it varying graph connectedness}\\
        \midrule
        $0.5d$&
        {\footnotesize \bf 3.75}{\scriptsize $\pm$1.60}  &   {\footnotesize 7.33}{\scriptsize $\pm$0.13}&
        {\footnotesize \bf 0.50}{\scriptsize $\pm$0.25}  &   {\footnotesize 1.05}{\scriptsize $\pm$0.02}&
        {\footnotesize 0.83}{\scriptsize $\pm$0.19}  &   {\footnotesize \bf 0.88}{\scriptsize $\pm$0.04}&
        {\footnotesize \bf 0.42}{\scriptsize $\pm$0.16}  &   {\footnotesize 0.73}{\scriptsize $\pm$0.01}\\

        $1d$&
        {\footnotesize \bf 3.50}{\scriptsize $\pm$0.86}  &   {\footnotesize 7.67}{\scriptsize $\pm$0.45}&
        {\footnotesize \bf 0.55}{\scriptsize $\pm$0.22}  &   {\footnotesize 1.53}{\scriptsize $\pm$0.09}&
        {\footnotesize \bf 0.75}{\scriptsize $\pm$0.09}  &   {\footnotesize 0.46}{\scriptsize $\pm$0.09}&
        {\footnotesize \bf 0.40}{\scriptsize $\pm$0.09}  &   {\footnotesize  0.77}{\scriptsize $\pm$0.07}\\

        $1.5d$&
        {\footnotesize \bf 3.00}{\scriptsize $\pm$1.15}  &   {\footnotesize 5.67}{\scriptsize $\pm$1.75}&
        {\footnotesize \bf 1.00}{\scriptsize $\pm$0.19}  &   {\footnotesize 1.55}{\scriptsize $\pm$0.08}&
        {\footnotesize \bf 0.89}{\scriptsize $\pm$0.07}  &   {\footnotesize 0.62}{\scriptsize $\pm$0.06}&
        {\footnotesize \bf 0.32}{\scriptsize $\pm$0.05}  &   {\footnotesize 0.53}{\scriptsize $\pm$0.04}\\

        $2d$&
        {\footnotesize \bf 2.28}{\scriptsize $\pm$0.80}  &   {\footnotesize 3.67}{\scriptsize $\pm$0.82}&
        {\footnotesize \bf 1.00}{\scriptsize $\pm$0.32}  &   {\footnotesize 2.67}{\scriptsize $\pm$0.63}&
        {\footnotesize 0.67}{\scriptsize $\pm$0.17}  &   {\footnotesize \bf 0.70}{\scriptsize $\pm$0.09}&
        {\footnotesize \bf 0.11}{\scriptsize $\pm$0.03}  &   {\footnotesize 0.32}{\scriptsize $\pm$0.08}\\

        \midrule
        $d$& \multicolumn{8}{c}{\it varying dimension count}\\
        \midrule
        $5$&
        {\footnotesize \bf 2.28}{\scriptsize $\pm$0.80}  &   {\footnotesize 3.67}{\scriptsize $\pm$0.82}&
        {\footnotesize \bf 1.00}{\scriptsize $\pm$0.32}  &   {\footnotesize 2.67}{\scriptsize $\pm$0.63}&
        {\footnotesize 0.67}{\scriptsize $\pm$0.17}  &   {\footnotesize \bf 0.70}{\scriptsize $\pm$0.09}&
        {\footnotesize \bf 0.11}{\scriptsize $\pm$0.03}  &   {\footnotesize 0.32}{\scriptsize $\pm$0.08}\\

        $7$&
        {\footnotesize \bf 8.67}{\scriptsize $\pm$0.56}  &    {\footnotesize  12.88}{\scriptsize $\pm$0.15}&
        {\footnotesize \bf 0.72}{\scriptsize $\pm$0.05}  &   {\footnotesize 1.07}{\scriptsize $\pm$0.01}&
        {\footnotesize \bf 0.96}{\scriptsize $\pm$0.02}  &   {\footnotesize 0.83}{\scriptsize $\pm$0.01}&
        {\footnotesize \bf 0.49}{\scriptsize $\pm$0.01}  &   {\footnotesize  0.63}{\scriptsize $\pm$0.01}\\

        $10$&
        {\footnotesize \bf 19.71}{\scriptsize $\pm$0.72}  &   {\footnotesize 30.82}{\scriptsize $\pm$0.98}&
        {\footnotesize \bf 0.42}{\scriptsize $\pm$0.13}  &   {\footnotesize 1.18}{\scriptsize $\pm$0.04}&
        {\footnotesize  0.70}{\scriptsize $\pm$0.16}  &   {\footnotesize \bf 0.71}{\scriptsize $\pm$0.06}&
        {\footnotesize \bf 0.34}{\scriptsize $\pm$0.08}  &   {\footnotesize 0.70}{\scriptsize $\pm$0.02}\\

    \bottomrule
    \end{tabularx}
    \vspace{-10pt}
\end{table}

\begin{table}[t]
    \centering
    \caption{{\bf Results on Erdos-Renyì (ER) graphs.} {\it First block:} We sample five different ER random graphs, and accompanying non-linear structural equations using an index-model. From each system, we then sample a varying number of samples and evaluate \texttt{NOTEARS-SOB} {\it with} D-Struct-SOB (indicated as {``\cmark''}) and {\it without} D-Struct (indicated as {``\xmark''}). {\it Second block:} For each row we sample a new ER graph with a varying degree of connectedness ($s$ indicates the expected number of edges). In both cases, we report the average performance in terms of SHD, FPR, TPR, and FDR, with std in scriptsize.}
    \label{tab:app:results:ER-Sob}
    \begin{tabularx}{\textwidth}{r  *{4}{|CC}}
    \toprule
        {\it metric} & 
        
        \multicolumn{2}{c}{\bf SHD ($\downarrow$)}  &
        \multicolumn{2}{c}{\bf FPR ($\downarrow$)}  &
        \multicolumn{2}{c}{\bf TPR ($\uparrow$)}    &
        \multicolumn{2}{c}{\bf FDR ($\downarrow$)}  \\
        
        \midrule
        
        {\it D-Struct-SOB}& 
        \cmark & \xmark & 
        \cmark & \xmark & 
        \cmark & \xmark & 
        \cmark & \xmark \\
        
        \toprule
        $n$& \multicolumn{8}{c}{\it varying sample size}\\
        \midrule
        $200$&
        {\footnotesize 5.20}{\scriptsize $\pm$1.11}  &   {\footnotesize \bf 3.80}{\scriptsize $\pm$0.43}&
        {\footnotesize \bf 3.60}{\scriptsize $\pm$0.81}  &   {\footnotesize 3.80}{\scriptsize $\pm$0.43}&
        {\footnotesize 0.41}{\scriptsize $\pm$0.13}  &   {\footnotesize \bf  0.68}{\scriptsize $\pm$0.05}&
        {\footnotesize 0.45}{\scriptsize $\pm$0.09}  &   {\footnotesize \bf  0.39}{\scriptsize $\pm$0.04}\\

        $500$&
        {\footnotesize  4.40 }{\scriptsize $\pm$1.36}  &   {\footnotesize  \bf  4.20}{\scriptsize $\pm$0.39}&
        {\footnotesize \bf 1.60}{\scriptsize $\pm$0.67}  &   {\footnotesize 4.20}{\scriptsize $\pm$0.39}&
        {\footnotesize 0.58}{\scriptsize $\pm$0.14}  &   {\footnotesize \bf  0.64}{\scriptsize $\pm$0.04}&
        {\footnotesize \bf 0.26}{\scriptsize $\pm$0.14}  &   {\footnotesize 0.42}{\scriptsize $\pm$0.04}\\

        $1000$&
        {\footnotesize \bf 3.00}{\scriptsize $\pm$0.36}  &   {\footnotesize 4.00}{\scriptsize $\pm$0.52}&
        {\footnotesize \bf 2.33}{\scriptsize $\pm$0.21}  &   {\footnotesize 4.00}{\scriptsize $\pm$0.52}&
        {\footnotesize \bf  0.74}{\scriptsize $\pm$0.05}  &   {\footnotesize  0.67}{\scriptsize $\pm$0.06}&
        {\footnotesize \bf 0.26}{\scriptsize $\pm$0.02}  &   {\footnotesize 0.40}{\scriptsize $\pm$0.05}\\

        $2000$&
        {\footnotesize 3.50}{\scriptsize $\pm$0.37}  &   {\footnotesize \bf  2.50}{\scriptsize $\pm$0.20}&
        {\footnotesize \bf 1.86}{\scriptsize $\pm$0.26}  &   {\footnotesize 2.50}{\scriptsize $\pm$0.20}&
        {\footnotesize  0.66}{\scriptsize $\pm$0.05}  &   {\footnotesize  \bf  0.83}{\scriptsize $\pm$0.02}&
        {\footnotesize \bf 0.24}{\scriptsize $\pm$0.03}  &   {\footnotesize 0.25}{\scriptsize $\pm$0.02}\\

        \midrule
        $s$& \multicolumn{8}{c}{\it varying graph connectedness}\\
        \midrule
        $0.5d$&
        {\footnotesize \bf 31.50}{\scriptsize $\pm$7.50}  &   {\footnotesize 42.00}{\scriptsize $\pm$0.25}&
        {\footnotesize \bf 0.74}{\scriptsize $\pm$0.19}  &   {\footnotesize 1.00}{\scriptsize $\pm$0.005}&
        {\footnotesize 0.83}{\scriptsize $\pm$0.17}  &   {\footnotesize \bf 0.88}{\scriptsize $\pm$0.05}&
        {\footnotesize \bf 0.82}{\scriptsize $\pm$0.0043}  &   {\footnotesize 0.94}{\scriptsize $\pm$0.003}\\
       
        $1d$&
        {\footnotesize \bf 24.00}{\scriptsize $\pm$1.35}  &   {\footnotesize 42.00}{\scriptsize $\pm$0.49}&
        {\footnotesize \bf 0.59}{\scriptsize $\pm$0.04}  &   {\footnotesize 1.05}{\scriptsize $\pm$0.01}&
        {\footnotesize \bf 0.95}{\scriptsize $\pm$0.05}  &   {\footnotesize 0.60}{\scriptsize $\pm$0.10}&
        {\footnotesize \bf 0.83}{\scriptsize $\pm$0.009}  &   {\footnotesize  0.93}{\scriptsize $\pm$0.01}\\
         
        $1.5d$&
        {\footnotesize \bf 30.67}{\scriptsize $\pm$3.52}  &   {\footnotesize 40.38}{\scriptsize $\pm$0.33}&
        {\footnotesize \bf 0.81}{\scriptsize $\pm$0.09}  &   {\footnotesize 1.06}{\scriptsize $\pm$0.008}&
        {\footnotesize \bf 0.90}{\scriptsize $\pm$0.05}  &   {\footnotesize 0.66}{\scriptsize $\pm$0.05}&
        {\footnotesize \bf 0.83}{\scriptsize $\pm$0.02}  &   {\footnotesize 0.89}{\scriptsize $\pm$0.007}\\

        $2d$&
        {\footnotesize \bf 30.50}{\scriptsize $\pm$0.50}  &   {\footnotesize 38.00}{\scriptsize $\pm$0.64}&
        {\footnotesize \bf 0.87}{\scriptsize $\pm$0.01}  &   {\footnotesize 1.09}{\scriptsize $\pm$0.02}&
        {\footnotesize \bf  1.00}{\scriptsize $\pm$0.00}  &   {\footnotesize 0.68}{\scriptsize $\pm$0.05}&
        {\footnotesize \bf 0.75}{\scriptsize $\pm$0.00}  &   {\footnotesize 0.84}{\scriptsize $\pm$0.01}\\

        \midrule
        $d$& \multicolumn{8}{c}{\it varying dimension count}\\
        \midrule
        $5$&
        {\footnotesize \bf 3.00}{\scriptsize $\pm$0.36}  &   {\footnotesize 4.00}{\scriptsize $\pm$0.52}&
        {\footnotesize \bf 2.33}{\scriptsize $\pm$0.21}  &   {\footnotesize 4.00}{\scriptsize $\pm$0.52}&
        {\footnotesize  \bf  0.74}{\scriptsize $\pm$0.05}  &   {\footnotesize  0.67}{\scriptsize $\pm$0.06}&
        {\footnotesize \bf 0.26}{\scriptsize $\pm$0.02}  &   {\footnotesize 0.40}{\scriptsize $\pm$0.05}\\

        $7$&
        {\footnotesize \bf 7.19}{\scriptsize $\pm$0.48}  &    {\footnotesize  14.95}{\scriptsize $\pm$0.37}&
        {\footnotesize \bf 0.53}{\scriptsize $\pm$0.04}  &   {\footnotesize 1.24}{\scriptsize $\pm$0.03}&
        {\footnotesize \bf 0.86}{\scriptsize $\pm$0.02}  &   {\footnotesize 0.65}{\scriptsize $\pm$0.04}&
        {\footnotesize \bf 0.44}{\scriptsize $\pm$0.02}  &   {\footnotesize  0.72}{\scriptsize $\pm$0.02}\\
    
        $10$&
        {\footnotesize \bf 29.67}{\scriptsize $\pm$2.33}  &   {\footnotesize 38.33}{\scriptsize $\pm$0.17}&
        {\footnotesize \bf 0.81}{\scriptsize $\pm$0.06}  &   {\footnotesize 1.06}{\scriptsize $\pm$0.004}&
        {\footnotesize  \bf  0.96}{\scriptsize $\pm$0.04}  &   {\footnotesize  0.70}{\scriptsize $\pm$0.02}&
        {\footnotesize \bf 0.77}{\scriptsize $\pm$0.02}  &   {\footnotesize 0.86}{\scriptsize $\pm$0.005}\\

           \midrule
        $K$& \multicolumn{8}{c}{\it varying subset count}\\
        \midrule
        $2$&
        {\footnotesize \bf 4.50}{\scriptsize $\pm$0.866}  &   {\footnotesize 5.67}{\scriptsize $\pm$0.46}&
        {\footnotesize \bf 4.00}{\scriptsize $\pm$0.71}  &   {\footnotesize 5.67}{\scriptsize $\pm$0.46}&
        {\footnotesize \bf 0.56}{\scriptsize $\pm$0.11}  &   {\footnotesize  0.48}{\scriptsize $\pm$0.05}&
        {\footnotesize \bf 0.46}{\scriptsize $\pm$0.09}  &   {\footnotesize 0.56}{\scriptsize $\pm$0.05}\\

        $3$&
        {\footnotesize \bf 3.00}{\scriptsize $\pm$0.36}  &   {\footnotesize 4.00}{\scriptsize $\pm$0.52}&
        {\footnotesize \bf 2.33}{\scriptsize $\pm$0.21}  &   {\footnotesize 4.00}{\scriptsize $\pm$0.52}&
        {\footnotesize \bf  0.74}{\scriptsize $\pm$0.05}  &   {\footnotesize  0.67}{\scriptsize $\pm$0.06}&
        {\footnotesize \bf 0.26}{\scriptsize $\pm$0.02}  &   {\footnotesize 0.40}{\scriptsize $\pm$0.05}\\

        $5$&
        {\footnotesize \bf 2.50}{\scriptsize $\pm$0.29}  &   {\footnotesize 4.17}{\scriptsize $\pm$0.63}&
        {\footnotesize \bf 2.50}{\scriptsize $\pm$0.29}  &   {\footnotesize 4.17}{\scriptsize $\pm$0.63}&
        {\footnotesize  \bf 0.77}{\scriptsize $\pm$0.06}  &   {\footnotesize  0.65}{\scriptsize $\pm$0.07}&
        {\footnotesize \bf 0.27}{\scriptsize $\pm$0.04}  &   {\footnotesize 0.42}{\scriptsize $\pm$0.06}\\

    \bottomrule
    \end{tabularx}
    \vspace{-10pt}
\end{table}

\subsection{Other DSFs} \label{app:additional_results:dsfs}
We repeat the results above for \texttt{NOTEARS-SOB} which is a Sobolev-based implementation of NOTEARS, in \cref{tab:extra_results:nt-sob-SF,tab:app:results:ER-Sob}. The main difference here with \texttt{NOTEARS-MLP} is the nonparametric estimation of the structural equations in $\hat{\mathcal{G}}$. Note that, future implementations of DSFs broadly alter the way in which the structural equations are estimated, and much less on how the proposed structure is evaluated to be a DAG (as they are mostly based on \cref{eq:h}). Overall, we find that \texttt{NOTEARS-SOB} behaves the same as \texttt{NOTEARS-MLP}: D-Struct vastly improves performance. We further test D-Struct for high dimensional settings using DAGMA \citep{bello2022dagma} in \cref{tab:res:highdim}, a recent DSF where the score function relies on a log-determinant and can be optimised using Adam which results in large performance increases.

\begin{table}[]
    \centering
    \caption{{\bf D-Struct in high dimensions.} We use a D-Struct variant of DAGMA \citep{bello2022dagma} which can be optimised using the Adam optimiser resulting in large performance increases. This allows us to scale D-Struct to high dimensions.}
    \label{tab:res:highdim}
    \begin{tabularx}{.7\linewidth}{l | *{6}{X}}
        \toprule
        {\it method} & $d$ & {\bf SHD($\downarrow$)} & {\bf FPR($\downarrow$)} & {\bf TPR($\uparrow$)} & {\bf FDR($\downarrow$)}\\
        \midrule
        D-Struct (DAGMA)& 	100&	{\bf 3}&	{\bf 0.0}&	{\bf 0.0}&	{\bf 0.94}\\
        DAGMA&	100&	11&	0.001&	0.085&	0.86\\
        \midrule
        D-Struct (DAGMA)&	50&		{\bf 1}&	{\bf 0.0}&	{\bf 0.0}&	{\bf 0.98}\\
        DAGMA&	50&	15&	0.003&	0.093&	0.78\\
        \bottomrule
    \end{tabularx}
\end{table}

Note that code to reproduce the above results is provided in the online code repository linked to above.

\begin{table}[t]
    
    \centering
    \caption{{\bf Results on Scale-Free (SF) graphs.} {\it First block:} We sample five different SF random graphs, and accompanying non-linear structural equations using an index-model. From each system, we then sample a varying number of samples and evaluate \texttt{NOTEARS-SOB} {\it with} D-Struct (indicated as {``\cmark''}) and {\it without} D-Struct (indicated as {``\xmark''}). {\it Second block:} For each row we sample a new SF graph with a varying degree of connectedness ($s$ indicates the expected number of edges).  {\it Third block:} For each row we vary the feature dimension count ($d$). {\it Fourth block:} For each row we vary the number of subsets for D-Struct ($s$).In all cases, we report the average performance in terms of SHD, FPR, TPR, and FDR, with std in scriptsize.}
    \label{tab:extra_results:nt-sob-SF}
    \begin{tabularx}{\textwidth}{r  *{4}{|CC}}
    \toprule
        {\it metric} & 
        
        \multicolumn{2}{c}{\bf SHD ($\downarrow$)}  &
        \multicolumn{2}{c}{\bf FPR ($\downarrow$)}  &
        \multicolumn{2}{c}{\bf TPR ($\uparrow$)}    &
        \multicolumn{2}{c}{\bf FDR ($\downarrow$)}  \\
        
        \midrule
        
        {\it D-Struct}& 
        \cmark & \xmark & 
        \cmark & \xmark & 
        \cmark & \xmark & 
        \cmark & \xmark \\
        
        \toprule
        $n$& \multicolumn{8}{c}{\it varying sample size}\\
        \midrule
        $200$&
        {\footnotesize  6.00}{\scriptsize $\pm$0.69}  &   {\footnotesize 3.8}{\scriptsize $\pm$0.25}&
        {\footnotesize  1.8}{\scriptsize $\pm$0.20}  &   {\footnotesize 1.27}{\scriptsize $\pm$0.08}&
        {\footnotesize 0.49}{\scriptsize $\pm$0.12}  &   {\footnotesize  0.83}{\scriptsize $\pm$0.03}&
        {\footnotesize  0.63}{\scriptsize $\pm$0.08}  &   {\footnotesize 0.39}{\scriptsize $\pm$0.02}\\

        $500$&
        {\footnotesize \bf 3.40}{\scriptsize $\pm$0.88}  &   {\footnotesize 4.60}{\scriptsize $\pm$0.25}&
        {\footnotesize \bf 0.67}{\scriptsize $\pm$0.14}  &   {\footnotesize 1.53}{\scriptsize $\pm$0.08}&
        {\footnotesize  0.57}{\scriptsize $\pm$0.09}  &   {\footnotesize 0.69}{\scriptsize $\pm$0.03}&
        {\footnotesize \bf 0.41}{\scriptsize $\pm$0.12}  &   {\footnotesize 0.48}{\scriptsize $\pm$0.03}\\

        $1000$&
        {\footnotesize \bf 2.75}{\scriptsize $\pm$0.86}  &   {\footnotesize 4.33}{\scriptsize $\pm$0.50}&
        {\footnotesize \bf 0.58}{\scriptsize $\pm$0.22}  &   {\footnotesize 1.44}{\scriptsize $\pm$0.17}&
        {\footnotesize  0.61}{\scriptsize $\pm$0.15}  &   {\footnotesize  0.76}{\scriptsize $\pm$0.05}&
        {\footnotesize \bf 0.36}{\scriptsize $\pm$0.15}  &   {\footnotesize 0.44}{\scriptsize $\pm$0.05}\\

        \midrule
        $s$& \multicolumn{8}{c}{\it varying graph connectedness}\\
        \midrule
        $0.5d$&
        {\footnotesize \bf 14.11}{\scriptsize $\pm$5.40}  &   {\footnotesize 39.53}{\scriptsize $\pm$0.37}&
        {\footnotesize \bf 0.31}{\scriptsize $\pm$0.13}  &   {\footnotesize 0.89}{\scriptsize $\pm$0.01}&
        {\footnotesize \bf 0.22}{\scriptsize $\pm$0.14}  &   {\footnotesize  0.20}{\scriptsize $\pm$0.11}&
        {\footnotesize \bf 0.42}{\scriptsize $\pm$0.17}  &   {\footnotesize 0.93}{\scriptsize $\pm$0.01}\\

        $1d$&
        {\footnotesize \bf 8.11}{\scriptsize $\pm$3.96}  &   {\footnotesize 39.46}{\scriptsize $\pm$0.38}&
        {\footnotesize \bf 0.13}{\scriptsize $\pm$0.09}  &   {\footnotesize 0.89}{\scriptsize $\pm$0.01}&
        {\footnotesize \bf 0.30}{\scriptsize $\pm$0.18}  &   {\footnotesize 0.16}{\scriptsize $\pm$0.09}&
        {\footnotesize \bf 0.22}{\scriptsize $\pm$0.15}  &   {\footnotesize  0.99}{\scriptsize $\pm$0.01}\\

        $1.5d$&
        {\footnotesize \bf 15.20}{\scriptsize $\pm$3.44}  &   {\footnotesize 38.31}{\scriptsize $\pm$0.41}&
        {\footnotesize \bf 0.32}{\scriptsize $\pm$0.14}  &   {\footnotesize 1.05}{\scriptsize $\pm$0.01}&
        {\footnotesize \bf 0.58}{\scriptsize $\pm$0.23}  &   {\footnotesize 0.52}{\scriptsize $\pm$0.07}&
        {\footnotesize \bf 0.40}{\scriptsize $\pm$0.17}  &   {\footnotesize 0.98}{\scriptsize $\pm$0.01}\\

        $2d$&
        {\footnotesize \bf 15.20}{\scriptsize $\pm$3.44}  &   {\footnotesize 38.25}{\scriptsize $\pm$0.44}&
        {\footnotesize \bf 0.32}{\scriptsize $\pm$0.14}  &   {\footnotesize 1.04}{\scriptsize $\pm$0.01}&
        {\footnotesize \bf 0.58}{\scriptsize $\pm$0.23}  &   {\footnotesize  0.50}{\scriptsize $\pm$0.07}&
        {\footnotesize \bf 0.40}{\scriptsize $\pm$0.17}  &   {\footnotesize 0.89}{\scriptsize $\pm$0.01}\\

        \midrule
        $d$& \multicolumn{8}{c}{\it varying dimension count}\\
        \midrule
        $5$&
        {\footnotesize \bf 2.75}{\scriptsize $\pm$0.86}  &   {\footnotesize 4.33}{\scriptsize $\pm$0.50}&
        {\footnotesize \bf 0.58}{\scriptsize $\pm$0.22}  &   {\footnotesize 1.44}{\scriptsize $\pm$0.17}&
        {\footnotesize 0.61}{\scriptsize $\pm$0.15}  &   {\footnotesize  0.76}{\scriptsize $\pm$0.05}&
        {\footnotesize \bf 0.36}{\scriptsize $\pm$0.15}  &   {\footnotesize 0.44}{\scriptsize $\pm$0.05}\\

        $7$&
        {\footnotesize \bf 8.25}{\scriptsize $\pm$3.09}  &    {\footnotesize  15.00}{\scriptsize $\pm$0.22}&
        {\footnotesize \bf 0.55}{\scriptsize $\pm$0.21}  &   {\footnotesize 1.00}{\scriptsize $\pm$0.01}&
        {\footnotesize \bf 0.96}{\scriptsize $\pm$0.08}  &   {\footnotesize 0.78}{\scriptsize $\pm$0.02}&
        {\footnotesize \bf 0.49}{\scriptsize $\pm$0.16}  &   {\footnotesize  0.76}{\scriptsize $\pm$0.01}\\

        $10$&
        {\footnotesize \bf 16.80}{\scriptsize $\pm$4.21}  &   {\footnotesize 35.75}{\scriptsize $\pm$0.33}&
        {\footnotesize \bf 0.36}{\scriptsize $\pm$0.17}  &   {\footnotesize 0.99}{\scriptsize $\pm$0.01}&
        {\footnotesize  0.58}{\scriptsize $\pm$0.24}  &   {\footnotesize 0.67}{\scriptsize $\pm$0.03}&
        {\footnotesize \bf 0.42}{\scriptsize $\pm$0.17}  &   {\footnotesize 0.85}{\scriptsize $\pm$0.01}\\
        
        \midrule
        $K$& \multicolumn{8}{c}{\it varying subset count}\\
        \midrule
        $2$&
        {\footnotesize \bf 3.00}{\scriptsize $\pm$0.42}  &   {\footnotesize 6.00}{\scriptsize $\pm$0.30}&
        {\footnotesize \bf 0.53}{\scriptsize $\pm$0.21}  &   {\footnotesize 2.00}{\scriptsize $\pm$0.10}&
        {\footnotesize \bf 0.66}{\scriptsize $\pm$0.04}  &   {\footnotesize  0.57}{\scriptsize $\pm$0.04}&
        {\footnotesize \bf 0.21}{\scriptsize $\pm$0.07}  &   {\footnotesize 0.60}{\scriptsize $\pm$0.03}\\

        $3$&
        {\footnotesize \bf 2.75}{\scriptsize $\pm$0.86}  &   {\footnotesize 4.33}{\scriptsize $\pm$0.5}&
        {\footnotesize \bf 0.58}{\scriptsize $\pm$0.22}  &   {\footnotesize 1.44}{\scriptsize $\pm$0.17}&
        {\footnotesize  0.61}{\scriptsize $\pm$0.15}  &   {\footnotesize  0.76}{\scriptsize $\pm$0.05}&
        {\footnotesize \bf 0.36}{\scriptsize $\pm$0.15}  &   {\footnotesize 0.44}{\scriptsize $\pm$0.05}\\

        $5$&
        {\footnotesize \bf 2.80}{\scriptsize $\pm$0.57}  &   {\footnotesize 5.25}{\scriptsize $\pm$0.21}&
        {\footnotesize \bf 0.73}{\scriptsize $\pm$0.15}  &   {\footnotesize 1.75}{\scriptsize $\pm$0.07}&
        {\footnotesize  \bf 0.74}{\scriptsize $\pm$0.09}  &   {\footnotesize  0.68}{\scriptsize $\pm$0.03}&
        {\footnotesize \bf 0.31}{\scriptsize $\pm$0.07}  &   {\footnotesize 0.53}{\scriptsize $\pm$0.02}\\

    \bottomrule
    \end{tabularx}
    \vspace{-10pt}
\end{table}

\begin{table}[t]
    
    \centering
    \caption{{\bf Results on Scale-Free (SF) graphs.} {\it First block:} We sample five different SF random graphs, and accompanying non-linear structural equations using an index-model. From each system, we then sample a varying number of samples and evaluate \texttt{NOTEARS-MLP} {\it with} D-Struct (indicated as {``\cmark''}) and {\it without} D-Struct (indicated as {``\xmark''}). {\it Second block:} For each row we sample a new SF graph with a varying degree of connectedness ($s$ indicates the expected number of edges).  {\it Third block:} For each row we vary the feature dimension count ($d$). {\it Fourth block:} For each row we vary the number of subsets for D-Struct ($s$).In all cases, we report the average performance in terms of SHD, FPR, TPR, and FDR, with std in scriptsize.}
    \label{tab:extra_results:nt-mlp-SF}
    \begin{tabularx}{\textwidth}{r  *{4}{|CC}}
    \toprule
        {\it metric} & 
        
        \multicolumn{2}{c}{\bf SHD ($\downarrow$)}  &
        \multicolumn{2}{c}{\bf FPR ($\downarrow$)}  &
        \multicolumn{2}{c}{\bf TPR ($\uparrow$)}    &
        \multicolumn{2}{c}{\bf FDR ($\downarrow$)}  \\
        
        \midrule
        
        {\it D-Struct}& 
        \cmark & \xmark & 
        \cmark & \xmark & 
        \cmark & \xmark & 
        \cmark & \xmark \\
        
        \toprule
        $n$& \multicolumn{8}{c}{\it varying sample size}\\
        \midrule
        $200$&
        {\footnotesize \bf 2.80}{\scriptsize $\pm$0.86}  &   {\footnotesize 6.20}{\scriptsize $\pm$0.57}&
        {\footnotesize \bf 0.73}{\scriptsize $\pm$0.28}  &   {\footnotesize 2.07}{\scriptsize $\pm$0.19}&
        {\footnotesize 
        \bf 0.80}{\scriptsize $\pm$0.11}  &   {\footnotesize  0.54}{\scriptsize $\pm$0.08}&
        {\footnotesize \bf 0.26}{\scriptsize $\pm$0.11}  &   {\footnotesize 0.62}{\scriptsize $\pm$0.06}\\

        $500$&
        {\footnotesize \bf 2.20}{\scriptsize $\pm$0.80}  &   {\footnotesize 7.20}{\scriptsize $\pm$0.66}&
        {\footnotesize \bf 0.27}{\scriptsize $\pm$0.12}  &   {\footnotesize 2.20}{\scriptsize $\pm$0.18}&
        {\footnotesize \bf 0.77}{\scriptsize $\pm$0.13}  &   {\footnotesize 0.37}{\scriptsize $\pm$0.09}&
        {\footnotesize \bf 0.14}{\scriptsize $\pm$0.06}  &   {\footnotesize 0.72}{\scriptsize $\pm$0.06}\\

        $1000$&
        {\footnotesize \bf 3.25}{\scriptsize $\pm$1.49}  &   {\footnotesize 5.33}{\scriptsize $\pm$0.61}&
        {\footnotesize \bf 0.75}{\scriptsize $\pm$0.43}  &   {\footnotesize 1.78}{\scriptsize $\pm$0.20}&
        {\footnotesize \bf 0.68}{\scriptsize $\pm$0.15}  &   {\footnotesize  0.66}{\scriptsize $\pm$0.08}&
        {\footnotesize \bf 0.29}{\scriptsize $\pm$0.18}  &   {\footnotesize 0.53}{\scriptsize $\pm$0.06}\\

        \midrule
        $s$& \multicolumn{8}{c}{\it varying graph connectedness}\\
        \midrule
        $0.5d$&
        {\footnotesize \bf 3.33}{\scriptsize $\pm$0.88}  &   {\footnotesize 8.00}{\scriptsize $\pm$0.37}&
        {\footnotesize \bf 0.50}{\scriptsize $\pm$0.19}  &   {\footnotesize 1.17}{\scriptsize $\pm$0.06}&
        {\footnotesize \bf 0.92}{\scriptsize $\pm$0.08}  &   {\footnotesize  0.38}{\scriptsize $\pm$0.05}&
        {\footnotesize \bf 0.41}{\scriptsize $\pm$0.08}  &   {\footnotesize 0.82}{\scriptsize $\pm$0.03}\\

        $1d$&
        {\footnotesize \bf 3.33}{\scriptsize $\pm$0.89}  &   {\footnotesize 8.00}{\scriptsize $\pm$1.00}&
        {\footnotesize \bf 0.50}{\scriptsize $\pm$0.19}  &   {\footnotesize 1.17}{\scriptsize $\pm$0.17}&
        {\footnotesize \bf 0.92}{\scriptsize $\pm$0.08}  &   {\footnotesize 0.38}{\scriptsize $\pm$0.13}&
        {\footnotesize \bf 0.41}{\scriptsize $\pm$0.08}  &   {\footnotesize  0.82}{\scriptsize $\pm$0.07}\\

        $1.5d$&
        {\footnotesize \bf 3.25}{\scriptsize $\pm$0.41}  &   {\footnotesize 7.67}{\scriptsize $\pm$0.31}&
        {\footnotesize \bf 0.50}{\scriptsize $\pm$0.07}  &   {\footnotesize 1.17}{\scriptsize $\pm$0.04}&
        {\footnotesize \bf 0.94}{\scriptsize $\pm$0.04}  &   {\footnotesize 0.42}{\scriptsize $\pm$0.06}&
        {\footnotesize \bf 0.43}{\scriptsize $\pm$0.03}  &   {\footnotesize 0.80}{\scriptsize $\pm$0.03}\\

        $2d$&
        {\footnotesize \bf 2.75}{\scriptsize $\pm$1.03}  &   {\footnotesize 5.00}{\scriptsize $\pm$1.00}&
        {\footnotesize \bf 0.33}{\scriptsize $\pm$0.23}  &   {\footnotesize 1.22}{\scriptsize $\pm$0.22}&
        {\footnotesize \bf 0.64}{\scriptsize $\pm$0.15}  &   {\footnotesize  0.50}{\scriptsize $\pm$0.07}&
        {\footnotesize \bf 0.14}{\scriptsize $\pm$0.09}  &   {\footnotesize 0.48}{\scriptsize $\pm$0.12}\\

        \midrule
        $d$& \multicolumn{8}{c}{\it varying dimension count}\\
        \midrule
        $5$&
        {\footnotesize \bf 3.25}{\scriptsize $\pm$1.49}  &   {\footnotesize 5.33}{\scriptsize $\pm$0.61}&
        {\footnotesize \bf 0.75}{\scriptsize $\pm$0.43}  &   {\footnotesize 1.78}{\scriptsize $\pm$0.20}&
        {\footnotesize \bf 0.68}{\scriptsize $\pm$0.15}  &   {\footnotesize  0.66}{\scriptsize $\pm$0.08}&
        {\footnotesize \bf 0.29}{\scriptsize $\pm$0.18}  &   {\footnotesize 0.53}{\scriptsize $\pm$0.06}\\

        $7$&
        {\footnotesize \bf 8.22}{\scriptsize $\pm$1.31}  &    {\footnotesize  15.67}{\scriptsize $\pm$0.14}&
        {\footnotesize \bf 0.54}{\scriptsize $\pm$0.09}  &   {\footnotesize 1.04}{\scriptsize $\pm$0.01}&
        {\footnotesize \bf 0.98}{\scriptsize $\pm$0.02}  &   {\footnotesize 0.83}{\scriptsize $\pm$0.03}&
        {\footnotesize \bf 0.54}{\scriptsize $\pm$0.04}  &   {\footnotesize  0.76}{\scriptsize $\pm$0.01}\\

        $10$&
        {\footnotesize \bf 16.80}{\scriptsize $\pm$4.21}  &   {\footnotesize 35.75}{\scriptsize $\pm$0.33}&
        {\footnotesize \bf 0.36}{\scriptsize $\pm$0.17}  &   {\footnotesize 0.99}{\scriptsize $\pm$0.01}&
        {\footnotesize  0.58}{\scriptsize $\pm$0.24}  &   {\footnotesize 0.67}{\scriptsize $\pm$0.03}&
        {\footnotesize \bf 0.42}{\scriptsize $\pm$0.17}  &   {\footnotesize 0.85}{\scriptsize $\pm$0.01}\\
        
        \midrule
        $K$& \multicolumn{8}{c}{\it varying subset count}\\
        \midrule
        $2$&
        {\footnotesize \bf 2.40}{\scriptsize $\pm$0.24}  &   {\footnotesize 6.50}{\scriptsize $\pm$0.46}&
        {\footnotesize \bf 0.53}{\scriptsize $\pm$0.08}  &   {\footnotesize 2.16}{\scriptsize $\pm$0.15}&
        {\footnotesize \bf 0.83}{\scriptsize $\pm$0.05}  &   {\footnotesize  0.50}{\scriptsize $\pm$0.06}&
        {\footnotesize \bf 0.21}{\scriptsize $\pm$0.03}  &   {\footnotesize 0.65}{\scriptsize $\pm$0.06}\\

        $3$&
        {\footnotesize \bf 2.00}{\scriptsize $\pm$1.04}  &   {\footnotesize 5.33}{\scriptsize $\pm$0.6}&
        {\footnotesize \bf 0.33}{\scriptsize $\pm$0.47}  &   {\footnotesize 1.78}{\scriptsize $\pm$0.20}&
        {\footnotesize \bf 0.68}{\scriptsize $\pm$0.14}  &   {\footnotesize  0.66}{\scriptsize $\pm$0.09}&
        {\footnotesize \bf 0.14}{\scriptsize $\pm$0.09}  &   {\footnotesize 0.53}{\scriptsize $\pm$0.06}\\

        $5$&
        {\footnotesize \bf 0.75}{\scriptsize $\pm$0.48}  &   {\footnotesize 5.25}{\scriptsize $\pm$0.21}&
        {\footnotesize \bf 0.25}{\scriptsize $\pm$0.16}  &   {\footnotesize 2.55}{\scriptsize $\pm$0.11}&
        {\footnotesize \bf  1.00}{\scriptsize $\pm$0.00}  &   {\footnotesize  0.33}{\scriptsize $\pm$0.05}&
        {\footnotesize \bf 0.09}{\scriptsize $\pm$0.05}  &   {\footnotesize 0.76}{\scriptsize $\pm$0.03}\\

    \bottomrule
    \end{tabularx}
    \vspace{-10pt}
\end{table}

\subsection{Computational efficiency} \label{app:additional-experiments:computation}

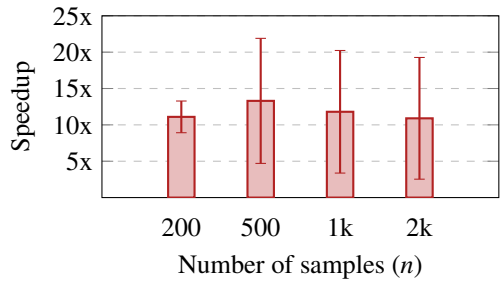
\begin{figure}
    \centering
    
    \begin{tikzpicture}
    \begin{axis}[
            title={},
            no markers,
            ybar,
            width=.4\textwidth,
            height=4cm,
            ymin=0, ymax=25,
            xmin=0, xmax=5,
            ytick={5,10,15,20,25,30},
            yticklabels={5x,10x,15x,20x,25x,30x},
            ylabel={Speedup},
            ylabel near ticks,
            xlabel={Number of samples (\textit{n})},
            xtick={0, 1, 2, 3, 4, 5},
            xtick style={draw=none},
            xticklabels={,200,500,1k,2k,},
            ymajorgrids=true,
            grid style=dashed,
        ]
        \addplot+[
            draw=FireBrick, 
            fill=FireBrick!30, 
            thick,
            error bars/.cd,
            error bar style={
                thick,
                FireBrick
            },
            y dir=both,
            y explicit,
        ] coordinates {
        
             (1,11.1) +- (0,2.18)
            (2,13.3) +- (0,8.6)
            (3,11.8) +- (0,8.43)
            (4,10.9)+- (0,8.37)
        };
        
        
    \end{axis}
    \end{tikzpicture}
    \caption{{\bf Speedup of D-Struct over NOTEARS.} Difference in computation time between \texttt{NOTEARS-MLP} and D-Struct, over $n$. On average, D-Struct is 10x quicker.}
    
    \label{res:computation}
\end{figure}

In \cref{res:computation}, we learn that despite its parallel ensemble architecture, D-Struct is actually {\it much} faster than NOTEARS. Note that D-Struct is built {\it on top} of NOTEARS, meaning this computational gain is not due to differences in implementation. Instead, we believe computation gains are largely due to D-Struct's learning scheme. Rather than using the {\it entire} dataset at once to learn one (computationally intensive) DSF, D-Struct splits the data and learns multiple DSFs from several {\it smaller} datasets. We believe this is an important result: the whole reason for having differentiable structure learners is due to their efficiency gains over CIT-based methods.

\subsection{Subsampling datasets} \label{app:additional-experiments:subsampling}

We refer to \cref{tab:results:subsample:full} for the full results presented originally in \cref{tab:results:subsample}. While FPR may be a little higher, using D-Struct still outperforms not using D-Struct in terms of the FPR-- already shown in \cref{tab:results}. Furthermore, as the subsampling routine forces D-Struct to learn on different distributions, it is possible that this increase in FPR is a result of initially more conflicting DAG structures. When combined, these structures include more edges which in turn result in more potential for a false positive edge discovery. In fact, we observe a lower necessary threshold when using our subsampling routine, necessary to transform the real-values matrix into a binary adjacency matrix.

\subsection{Binary adjacency matrices}\label{app:additional-experiments:threshold}

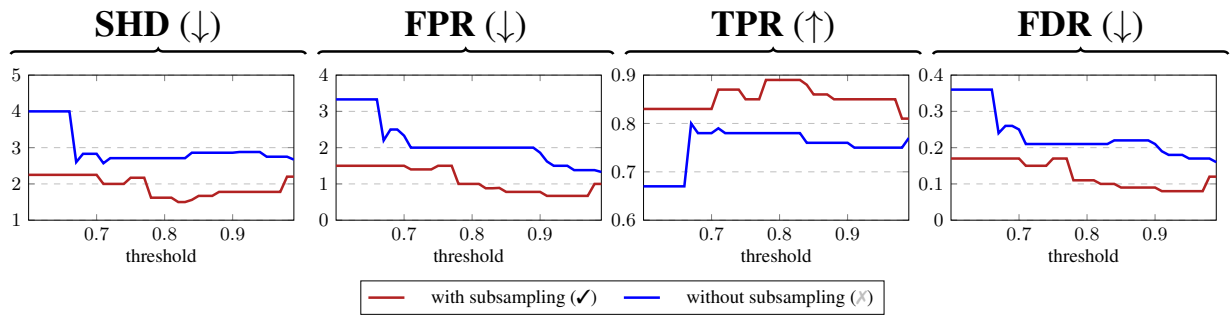
\begin{figure}[h]
    \centering
    \begin{subfigure}[t]{0.95\textwidth}
    \centering
    \resizebox{\textwidth}{!}{\begin{tikzpicture}
        \pgfplotsset{footnotesize,samples=10}
        
        \begin{groupplot}[group style = {group size = 4 by 1, horizontal sep = 20pt}, width = 6.0cm, height = 4.0cm]
            
            \nextgroupplot[
                legend style = { column sep = 10pt, legend columns = 5, legend to name = grouplegend,},
                xlabel={threshold},
                x label style={at={(axis description cs:0.5,-0.15)},anchor=north},
                ymin=1, ymax=5,
                ymajorgrids=true,
                xtick={.7,.8,.9},
                grid style=dashed,
                xmin=.6, xmax=.99
            ]
            \addplot[color=FireBrick,very thick]
            coordinates {(0.6, 2.25)
                        (0.61, 2.25)
                        (0.62, 2.25)
                        (0.63, 2.25)
                        (0.64, 2.25)
                        (0.65, 2.25)
                        (0.66, 2.25)
                        (0.67, 2.25)
                        (0.68, 2.25)
                        (0.69, 2.25)
                        (0.7, 2.25)
                        (0.71, 2.0)
                        (0.72, 2.0)
                        (0.73, 2.0)
                        (0.74, 2.0)
                        (0.75, 2.17)
                        (0.76, 2.17)
                        (0.77, 2.17)
                        (0.78, 1.62)
                        (0.79, 1.62)
                        (0.8, 1.62)
                        (0.81, 1.62)
                        (0.82, 1.5)
                        (0.83, 1.5)
                        (0.84, 1.56)
                        (0.85, 1.67)
                        (0.86, 1.67)
                        (0.87, 1.67)
                        (0.88, 1.78)
                        (0.89, 1.78)
                        (0.9, 1.78)
                        (0.91, 1.78)
                        (0.92, 1.78)
                        (0.93, 1.78)
                        (0.94, 1.78)
                        (0.95, 1.78)
                        (0.96, 1.78)
                        (0.97, 1.78)
                        (0.98, 2.2)
                        (0.99, 2.2)};
            \addlegendentry[Black]{with OUR subsampling (\cmark)}
                
            \addplot[color=blue,very thick]
            coordinates {(0.6, 4.0)
                        (0.61, 4.0)
                        (0.62, 4.0)
                        (0.63, 4.0)
                        (0.64, 4.0)
                        (0.65, 4.0)
                        (0.66, 4.0)
                        (0.67, 2.6)
                        (0.68, 2.83)
                        (0.69, 2.83)
                        (0.7, 2.83)
                        (0.71, 2.57)
                        (0.72, 2.71)
                        (0.73, 2.71)
                        (0.74, 2.71)
                        (0.75, 2.71)
                        (0.76, 2.71)
                        (0.77, 2.71)
                        (0.78, 2.71)
                        (0.79, 2.71)
                        (0.8, 2.71)
                        (0.81, 2.71)
                        (0.82, 2.71)
                        (0.83, 2.71)
                        (0.84, 2.86)
                        (0.85, 2.86)
                        (0.86, 2.86)
                        (0.87, 2.86)
                        (0.88, 2.86)
                        (0.89, 2.86)
                        (0.9, 2.86)
                        (0.91, 2.88)
                        (0.92, 2.88)
                        (0.93, 2.88)
                        (0.94, 2.88)
                        (0.95, 2.75)
                        (0.96, 2.75)
                        (0.97, 2.75)
                        (0.98, 2.75)
                        (0.99, 2.67)};
            \addlegendentry[Black]{without OUR subsampling (\xmark)}
            
            \nextgroupplot[ 
                legend style = { column sep = 10pt, legend columns = 5, legend to name = grouplegend,},
                xlabel={threshold},
                x label style={at={(axis description cs:0.5,-0.15)},anchor=north},
                ymin=0, ymax=4,
                ymajorgrids=true,
                grid style=dashed,
                xtick={.7,.8,.9},
                xmin=0.60, xmax=0.99
            ]

            \addplot[color=FireBrick,very thick]
            coordinates {(0.6, 1.5)
                        (0.61, 1.5)
                        (0.62, 1.5)
                        (0.63, 1.5)
                        (0.64, 1.5)
                        (0.65, 1.5)
                        (0.66, 1.5)
                        (0.67, 1.5)
                        (0.68, 1.5)
                        (0.69, 1.5)
                        (0.7, 1.5)
                        (0.71, 1.4)
                        (0.72, 1.4)
                        (0.73, 1.4)
                        (0.74, 1.4)
                        (0.75, 1.5)
                        (0.76, 1.5)
                        (0.77, 1.5)
                        (0.78, 1.0)
                        (0.79, 1.0)
                        (0.8, 1.0)
                        (0.81, 1.0)
                        (0.82, 0.88)
                        (0.83, 0.88)
                        (0.84, 0.89)
                        (0.85, 0.78)
                        (0.86, 0.78)
                        (0.87, 0.78)
                        (0.88, 0.78)
                        (0.89, 0.78)
                        (0.9, 0.78)
                        (0.91, 0.67)
                        (0.92, 0.67)
                        (0.93, 0.67)
                        (0.94, 0.67)
                        (0.95, 0.67)
                        (0.96, 0.67)
                        (0.97, 0.67)
                        (0.98, 1.0)
                        (0.99, 1.0)};
            \addlegendentry[Black]{with subsampling (\cmark)}
                
            \addplot[color=blue,very thick]
            coordinates {(0.6, 3.33)
                    (0.61, 3.33)
                    (0.62, 3.33)
                    (0.63, 3.33)
                    (0.64, 3.33)
                    (0.65, 3.33)
                    (0.66, 3.33)
                    (0.67, 2.2)
                    (0.68, 2.5)
                    (0.69, 2.5)
                    (0.7, 2.33)
                    (0.71, 2.0)
                    (0.72, 2.0)
                    (0.73, 2.0)
                    (0.74, 2.0)
                    (0.75, 2.0)
                    (0.76, 2.0)
                    (0.77, 2.0)
                    (0.78, 2.0)
                    (0.79, 2.0)
                    (0.8, 2.0)
                    (0.81, 2.0)
                    (0.82, 2.0)
                    (0.83, 2.0)
                    (0.84, 2.0)
                    (0.85, 2.0)
                    (0.86, 2.0)
                    (0.87, 2.0)
                    (0.88, 2.0)
                    (0.89, 2.0)
                    (0.9, 1.86)
                    (0.91, 1.62)
                    (0.92, 1.5)
                    (0.93, 1.5)
                    (0.94, 1.5)
                    (0.95, 1.38)
                    (0.96, 1.38)
                    (0.97, 1.38)
                    (0.98, 1.38)
                    (0.99, 1.33)};
            \addlegendentry[Black]{without subsampling (\xmark)}

            \nextgroupplot[ 
                legend style = { column sep = 10pt, legend columns = 5, legend to name = grouplegend,},
                xlabel={threshold},
                x label style={at={(axis description cs:0.5,-0.15)},anchor=north},
                ymin=.6, ymax=.9,
                ymajorgrids=true,
                grid style=dashed,
                xtick={.7,.8,.9},
                xmin=0.60, xmax=0.99
            ]
            \addplot[color=FireBrick,very thick]
            coordinates {(0.6, 0.83)
                    (0.61, 0.83)
                    (0.62, 0.83)
                    (0.63, 0.83)
                    (0.64, 0.83)
                    (0.65, 0.83)
                    (0.66, 0.83)
                    (0.67, 0.83)
                    (0.68, 0.83)
                    (0.69, 0.83)
                    (0.7, 0.83)
                    (0.71, 0.87)
                    (0.72, 0.87)
                    (0.73, 0.87)
                    (0.74, 0.87)
                    (0.75, 0.85)
                    (0.76, 0.85)
                    (0.77, 0.85)
                    (0.78, 0.89)
                    (0.79, 0.89)
                    (0.8, 0.89)
                    (0.81, 0.89)
                    (0.82, 0.89)
                    (0.83, 0.89)
                    (0.84, 0.88)
                    (0.85, 0.86)
                    (0.86, 0.86)
                    (0.87, 0.86)
                    (0.88, 0.85)
                    (0.89, 0.85)
                    (0.9, 0.85)
                    (0.91, 0.85)
                    (0.92, 0.85)
                    (0.93, 0.85)
                    (0.94, 0.85)
                    (0.95, 0.85)
                    (0.96, 0.85)
                    (0.97, 0.85)
                    (0.98, 0.81)
                    (0.99, 0.81)};
            \addlegendentry[Black]{with subsampling (\cmark)}
                
            \addplot[color=blue,very thick]
            coordinates {(0.6, 0.67)
                        (0.61, 0.67)
                        (0.62, 0.67)
                        (0.63, 0.67)
                        (0.64, 0.67)
                        (0.65, 0.67)
                        (0.66, 0.67)
                        (0.67, 0.8)
                        (0.68, 0.78)
                        (0.69, 0.78)
                        (0.7, 0.78)
                        (0.71, 0.79)
                        (0.72, 0.78)
                        (0.73, 0.78)
                        (0.74, 0.78)
                        (0.75, 0.78)
                        (0.76, 0.78)
                        (0.77, 0.78)
                        (0.78, 0.78)
                        (0.79, 0.78)
                        (0.8, 0.78)
                        (0.81, 0.78)
                        (0.82, 0.78)
                        (0.83, 0.78)
                        (0.84, 0.76)
                        (0.85, 0.76)
                        (0.86, 0.76)
                        (0.87, 0.76)
                        (0.88, 0.76)
                        (0.89, 0.76)
                        (0.9, 0.76)
                        (0.91, 0.75)
                        (0.92, 0.75)
                        (0.93, 0.75)
                        (0.94, 0.75)
                        (0.95, 0.75)
                        (0.96, 0.75)
                        (0.97, 0.75)
                        (0.98, 0.75)
                        (0.99, 0.77)};
            \addlegendentry[Black]{without subsampling (\xmark)}
            
            \nextgroupplot[
                legend style = { column sep = 10pt, legend columns = 5, legend to name = grouplegend,},
                xlabel={threshold},
                x label style={at={(axis description cs:0.5,-0.15)},anchor=north},
                ymin=.0, ymax=.4,
                ymajorgrids=true,
                grid style=dashed,
                xtick={.7,.8,.9},
                xmin=0.60, xmax=0.99
            ]
            \addplot[color=FireBrick,very thick]
            coordinates {(0.6, 0.17)
                        (0.61, 0.17)
                        (0.62, 0.17)
                        (0.63, 0.17)
                        (0.64, 0.17)
                        (0.65, 0.17)
                        (0.66, 0.17)
                        (0.67, 0.17)
                        (0.68, 0.17)
                        (0.69, 0.17)
                        (0.7, 0.17)
                        (0.71, 0.15)
                        (0.72, 0.15)
                        (0.73, 0.15)
                        (0.74, 0.15)
                        (0.75, 0.17)
                        (0.76, 0.17)
                        (0.77, 0.17)
                        (0.78, 0.11)
                        (0.79, 0.11)
                        (0.8, 0.11)
                        (0.81, 0.11)
                        (0.82, 0.1)
                        (0.83, 0.1)
                        (0.84, 0.1)
                        (0.85, 0.09)
                        (0.86, 0.09)
                        (0.87, 0.09)
                        (0.88, 0.09)
                        (0.89, 0.09)
                        (0.9, 0.09)
                        (0.91, 0.08)
                        (0.92, 0.08)
                        (0.93, 0.08)
                        (0.94, 0.08)
                        (0.95, 0.08)
                        (0.96, 0.08)
                        (0.97, 0.08)
                        (0.98, 0.12)
                        (0.99, 0.12)};
            \addlegendentry[Black]{with subsampling (\cmark)}
                
            \addplot[color=blue,very thick]
            coordinates {(0.6, 0.36)
                        (0.61, 0.36)
                        (0.62, 0.36)
                        (0.63, 0.36)
                        (0.64, 0.36)
                        (0.65, 0.36)
                        (0.66, 0.36)
                        (0.67, 0.24)
                        (0.68, 0.26)
                        (0.69, 0.26)
                        (0.7, 0.25)
                        (0.71, 0.21)
                        (0.72, 0.21)
                        (0.73, 0.21)
                        (0.74, 0.21)
                        (0.75, 0.21)
                        (0.76, 0.21)
                        (0.77, 0.21)
                        (0.78, 0.21)
                        (0.79, 0.21)
                        (0.8, 0.21)
                        (0.81, 0.21)
                        (0.82, 0.21)
                        (0.83, 0.21)
                        (0.84, 0.22)
                        (0.85, 0.22)
                        (0.86, 0.22)
                        (0.87, 0.22)
                        (0.88, 0.22)
                        (0.89, 0.22)
                        (0.9, 0.21)
                        (0.91, 0.19)
                        (0.92, 0.18)
                        (0.93, 0.18)
                        (0.94, 0.18)
                        (0.95, 0.17)
                        (0.96, 0.17)
                        (0.97, 0.17)
                        (0.98, 0.17)
                        (0.99, 0.16)};
            \addlegendentry[Black]{without subsampling (\xmark)}

        \end{groupplot}
        
        \node at ($(group c2r1)!0.5!(group c3r1) + (0cm, -2.5cm)$) {\ref{grouplegend}};

        \draw [
            very thick,
            decoration={
                brace,
                raise=-.15cm,
            },
            decorate
        ] ($(group c1r1.north west) + (-.3, .5)$) -- ($(group c1r1.north east) + (.2, .5)$) node [pos=0.5,anchor=center,yshift=.3cm] {\LARGE {\bf SHD} ($\downarrow$)};

        \draw [
            very thick,
            decoration={
                brace,
                raise=-.15cm,
            },
            decorate
        ] ($(group c2r1.north west) + (-.3, .5)$) -- ($(group c2r1.north east) + (.2, .5)$) node [pos=0.5,anchor=center,yshift=.3cm] {\LARGE {\bf FPR} ($\downarrow$)};
        
        \draw [
            very thick,
            decoration={
                brace,
                raise=-.15cm,
            },
            decorate
        ] ($(group c3r1.north west) + (-.3, .5)$) -- ($(group c3r1.north east) + (.2, .5)$) node [pos=0.5,anchor=center,yshift=.3cm] {\LARGE {\bf TPR} ($\uparrow$)};

        \draw [
            very thick,
            decoration={
                brace,
                raise=-.15cm,
            },
            decorate
        ] ($(group c4r1.north west) + (-.3, .5)$) -- ($(group c4r1.north east) + (.2, .5)$) node [pos=0.5,anchor=center,yshift=.3cm] {\LARGE {\bf FDR} ($\downarrow$)};

    \end{tikzpicture}}
    \end{subfigure}

    \caption{{\bf Subsampling with different DAG-thresholds.} The DAG-threshold transforms the real-valued adjacency matrix, to a binary one. As the threshold increases, the amount edges that remain part of the DAG decreases. The above confirms our findings from \cref{tab:results:subsample:full} in different settings.}
    \label{fig:sensitivity:thresh}
    \rule{\textwidth}{.5pt}
\end{figure}

We also report the same metrics as a function of the DAG-finding threshold in \cref{fig:sensitivity:thresh}, where the threshold is applied to the adjacency matrix to produce a binary matrix on which we compute the metrics. Of course, a threshold will be selected in practice; however, we show that for a range of plausible threshold values and all metrics that subsampling with our routine is indeed beneficial, compared to randomized subsampling. From this, it seems that the results we find in \cref{tab:results:subsample:full} are consistent even with changing thresholds.

\begin{table}[t]
    \centering
    \caption{{\bf Usefulness of our subsampling routine.} We sample ten different ER random graphs, and accompanying non-linear structural equations as in \citet{zheng2020learning}. From each system we then sample $n=2000$ samples, and evaluate \texttt{NOTEARS-MLP} {\it with} our subsampling routine from \cref{sec:method:single-origin} (indicated as {``\cmark''}) and {\it without} the subsampling routine, using random splits instead (indicated as {``\xmark''}). For each row, we repeat our experiment with different $K$. In both cases, we report the average performance in terms of SHD, FPR, TPR, and FDR, with std in scriptsize.}
    \vspace{-5pt}
    \label{tab:results:subsample:full}
    \begin{tabularx}{\textwidth}{r  *{4}{|CC}}
    \toprule
        {\it metric} & 
        
        \multicolumn{2}{c}{\bf SHD ($\downarrow$)}  &
        \multicolumn{2}{c}{\bf FPR ($\downarrow$)}  &
        \multicolumn{2}{c}{\bf TPR ($\uparrow$)}    &
        \multicolumn{2}{c}{\bf FDR ($\downarrow$)}  \\
        
        \midrule
        
        {\it Subsample}& 
        \cmark & \xmark & 
        \cmark & \xmark & 
        \cmark & \xmark & 
        \cmark & \xmark \\
        
        \toprule

     $K$& \multicolumn{8}{c}{\it varying amount of splits}\\
        \midrule
        $2$&
        {\footnotesize \bf 2.80}{\scriptsize $\pm$0.53}  &   {\footnotesize 3.40}{\scriptsize $\pm$0.58}&
        {\footnotesize  2.80}{\scriptsize $\pm$0.53}  &   {\footnotesize \bf 2.60}{\scriptsize $\pm$0.33}&
        {\footnotesize \bf 0.80}{\scriptsize $\pm$0.06}  &   {\footnotesize 0.71}{\scriptsize $\pm$0.07}&
        {\footnotesize \bf 0.28}{\scriptsize $\pm$0.05}  &   {\footnotesize 0.30}{\scriptsize $\pm$0.16}\\

        $3$&
        {\footnotesize \bf 3.00}{\scriptsize $\pm$0.37}  &   {\footnotesize 4.00}{\scriptsize $\pm$0.59}&
        {\footnotesize  2.00}{\scriptsize $\pm$0.51}  &   {\footnotesize \bf 1.60}{\scriptsize $\pm$0.45}&
        {\footnotesize \bf 0.73}{\scriptsize $\pm$0.04}  &   {\footnotesize 0.58}{\scriptsize $\pm$0.06}&
        {\footnotesize \bf 0.22}{\scriptsize $\pm$0.05}  &   {\footnotesize 0.24}{\scriptsize $\pm$0.17}\\

        $5$&
        {\footnotesize \bf 2.80}{\scriptsize $\pm$0.57}  &   {\footnotesize 4.40}{\scriptsize $\pm$1.29}&
        {\footnotesize  1.40}{\scriptsize $\pm$0.50}  &   {\footnotesize \bf 0.60}{\scriptsize $\pm$0.26}&
        {\footnotesize \bf 0.71}{\scriptsize $\pm$0.06}  &   {\footnotesize 0.53}{\scriptsize $\pm$0.15}&
        {\footnotesize 0.18}{\scriptsize $\pm$0.06}  &   {\footnotesize \bf  0.07}{\scriptsize $\pm$0.10}\\

    \bottomrule
    \end{tabularx}

\end{table}

\subsection{Multiple datasets}\label{app:additional_results:multiset}

Below we assess the scenario where indeed we have multiple datasets.  Essentially, we skip the step explained in \cref{sec:method:single-origin} and provide multiple datasets which we know to be differently distributed while respecting a shared underlying DAG: using the same underlying DAG, we sample different associated SEMs. We report these results in \cref{tab:app:multi_data_results:ER,tab:app:multi_data_results:ER}. As we have for other experiments, we arrive at  the same conclusion that D-Struct consistently performs better than the benchmark DSFs

\begin{table}[h]

    \centering
    \caption{{\bf Multiple datasets results on Erdos-Renyì (ER) graphs.} {\it First block:} We sample five different ER random graphs, and accompanying non-linear structural equations using an index-model. From each system, we then sample a varying number of samples and evaluate \texttt{NOTEARS-MLP} {\it with} D-Struct (indicated as {``\cmark''}) and {\it without} D-Struct (indicated as {``\xmark''}). {\it Second block:} We repeat the experiment in the first block but evaluate using \texttt{NOTEARS-Sob} with and without D-Struct. In both cases, we report the average performance in terms of SHD, FPR, TPR, and FDR, with std in scriptsize.}
    \label{tab:app:multi_data_results:ER}
    \begin{tabularx}{\textwidth}{r  *{4}{|CC}}
    \toprule
        {\it metric} & 
        
        \multicolumn{2}{c}{\bf SHD ($\downarrow$)}  &
        \multicolumn{2}{c}{\bf FPR ($\downarrow$)}  &
        \multicolumn{2}{c}{\bf TPR ($\uparrow$)}    &
        \multicolumn{2}{c}{\bf FDR ($\downarrow$)}  \\
        
        \midrule

                {\it D-Struct-MLP}& 
        \cmark & \xmark & 
        \cmark & \xmark & 
        \cmark & \xmark & 
        \cmark & \xmark \\
        
        \toprule
         
        $K$& \multicolumn{8}{c}{\it varying subset count}\\
        \midrule
        $2$&
        {\footnotesize \bf 3.00}{\scriptsize $\pm$1.05}  &   {\footnotesize 5.00}{\scriptsize $\pm$0.58}&
        {\footnotesize \bf 1.80}{\scriptsize $\pm$0.37}  &   {\footnotesize 5.00}{\scriptsize $\pm$0.58}&
        {\footnotesize \bf 0.76}{\scriptsize $\pm$0.12}  &   {\footnotesize  0.56}{\scriptsize $\pm$0.07}&
        {\footnotesize \bf 0.23}{\scriptsize $\pm$0.07}  &   {\footnotesize 0.50}{\scriptsize $\pm$0.06}\\

        $3$&
        {\footnotesize \bf 2.25}{\scriptsize $\pm$0.95}  &   {\footnotesize 2.67}{\scriptsize $\pm$0.45}&
        {\footnotesize \bf 1.75}{\scriptsize $\pm$0.47}  &   {\footnotesize 2.67}{\scriptsize $\pm$0.45}&
        {\footnotesize \bf  0.86}{\scriptsize $\pm$0.11}  &   {\footnotesize  0.81}{\scriptsize $\pm$0.05}&
        {\footnotesize \bf 0.19}{\scriptsize $\pm$0.06}  &   {\footnotesize 0.27}{\scriptsize $\pm$0.04}\\

        $5$&
        {\footnotesize \bf 1.667}{\scriptsize $\pm$1.20}  &   {\footnotesize 4.00}{\scriptsize $\pm$0.33}&
        {\footnotesize \bf 1.00}{\scriptsize $\pm$0.57}  &   {\footnotesize 4.00}{\scriptsize $\pm$0.33}&
        {\footnotesize  \bf 0.88}{\scriptsize $\pm$0.11}  &   {\footnotesize  0.66}{\scriptsize $\pm$0.04}&
        {\footnotesize \bf 0.12}{\scriptsize $\pm$0.07}  &   {\footnotesize 0.40}{\scriptsize $\pm$0.03}\\

        \midrule
        {\it D-Struct-SOB}& 
        \cmark & \xmark & 
        \cmark & \xmark & 
        \cmark & \xmark & 
        \cmark & \xmark \\
        
        \toprule
        
        $K$& \multicolumn{8}{c}{\it varying subset count}\\
        \midrule
        $2$&
        {\footnotesize \bf 2.75}{\scriptsize $\pm$0.48}  &   {\footnotesize 3.67}{\scriptsize $\pm$0.55}&
        {\footnotesize \bf 2.00}{\scriptsize $\pm$0.41}  &   {\footnotesize 3.67}{\scriptsize $\pm$0.55}&
        {\footnotesize \bf 0.78}{\scriptsize $\pm$0.06}  &   {\footnotesize  0.70}{\scriptsize $\pm$0.06}&
        {\footnotesize \bf 0.22}{\scriptsize $\pm$0.05}  &   {\footnotesize 0.37}{\scriptsize $\pm$0.05}\\

        $3$&
        {\footnotesize \bf 1.25}{\scriptsize $\pm$0.47}  &   {\footnotesize 3.33}{\scriptsize $\pm$0.33}&
        {\footnotesize \bf 1.00}{\scriptsize $\pm$0.41}  &   {\footnotesize 3.33}{\scriptsize $\pm$0.33}&
        {\footnotesize \bf  0.89}{\scriptsize $\pm$0.06}  &   {\footnotesize  0.74}{\scriptsize $\pm$0.04}&
        {\footnotesize \bf 0.11}{\scriptsize $\pm$0.05}  &   {\footnotesize 0.33}{\scriptsize $\pm$0.03}\\

        $5$&
        {\footnotesize \bf 2.00}{\scriptsize $\pm$0.41}  &   {\footnotesize 4.00}{\scriptsize $\pm$0.21}&
        {\footnotesize \bf 1.75}{\scriptsize $\pm$0.25}  &   {\footnotesize 4.00}{\scriptsize $\pm$0.21}&
        {\footnotesize  \bf 0.89}{\scriptsize $\pm$0.05}  &   {\footnotesize  0.66}{\scriptsize $\pm$0.02}&
        {\footnotesize \bf 0.18}{\scriptsize $\pm$0.03}  &   {\footnotesize 0.40}{\scriptsize $\pm$0.02}\\

    \bottomrule
    \end{tabularx}
    \vspace{-10pt}
\end{table}


\begin{table}[h]

    \centering
    \caption{{\bf Multiple datasets results on Scale-Free (SF) graphs.} {\it First block:} We sample five different SF random graphs, and accompanying non-linear structural equations using an index-model. From each system, we then sample a varying number of samples and evaluate \texttt{NOTEARS-MLP} {\it with} D-Struct (indicated as {``\cmark''}) and {\it without} D-Struct (indicated as {``\xmark''}). {\it Second block:} We repeat the experiment in the first block but evaluate using \texttt{NOTEARS-Sob} with and without D-Struct. In both cases, we report the average performance in terms of SHD, FPR, TPR, and FDR, with std in scriptsize.}
    \label{tab:app:multi_data_results:SF}
    \begin{tabularx}{\textwidth}{r  *{4}{|CC}}
    \toprule
        {\it metric} & 
        
        \multicolumn{2}{c}{\bf SHD ($\downarrow$)}  &
        \multicolumn{2}{c}{\bf FPR ($\downarrow$)}  &
        \multicolumn{2}{c}{\bf TPR ($\uparrow$)}    &
        \multicolumn{2}{c}{\bf FDR ($\downarrow$)}  \\
        
        \midrule

                {\it D-Struct-MLP}& 
        \cmark & \xmark & 
        \cmark & \xmark & 
        \cmark & \xmark & 
        \cmark & \xmark \\
        
        \toprule

        $K$& \multicolumn{8}{c}{\it varying subset count}\\
        \midrule
        $2$&
        {\footnotesize \bf 3.75}{\scriptsize $\pm$0.48}  &   {\footnotesize 4.67}{\scriptsize $\pm$0.55}&
        {\footnotesize \bf 0.83}{\scriptsize $\pm$0.35}  &   {\footnotesize 1.56}{\scriptsize $\pm$0.18}&
        {\footnotesize \bf 0.71}{\scriptsize $\pm$0.10}  &   {\footnotesize  0.62}{\scriptsize $\pm$0.08}&
        {\footnotesize \bf 0.28}{\scriptsize $\pm$0.10}  &   {\footnotesize 0.52}{\scriptsize $\pm$0.06}\\

        $3$&
        {\footnotesize \bf 2.00}{\scriptsize $\pm$0.58}  &   {\footnotesize 4.00}{\scriptsize $\pm$0.58}&
        {\footnotesize \bf 0.67}{\scriptsize $\pm$0.19}  &   {\footnotesize 1.22}{\scriptsize $\pm$0.15}&
        {\footnotesize \bf  0.86}{\scriptsize $\pm$0.08}  &   {\footnotesize  0.76}{\scriptsize $\pm$0.06}&
        {\footnotesize \bf 0.24}{\scriptsize $\pm$0.05}  &   {\footnotesize 0.41}{\scriptsize $\pm$0.05}\\

        $5$&
        {\footnotesize \bf 1.00}{\scriptsize $\pm$0.71}  &   {\footnotesize 5.67}{\scriptsize $\pm$0.50}&
        {\footnotesize \bf 0.17}{\scriptsize $\pm$0.17}  &   {\footnotesize 1.89}{\scriptsize $\pm$0.17}&
        {\footnotesize  \bf 0.86}{\scriptsize $\pm$0.10}  &   {\footnotesize  0.57}{\scriptsize $\pm$0.05}&
        {\footnotesize \bf 0.08}{\scriptsize $\pm$0.08}  &   {\footnotesize 0.58}{\scriptsize $\pm$0.04}\\

        \midrule
        {\it D-Struct-SOB}& 
        \cmark & \xmark & 
        \cmark & \xmark & 
        \cmark & \xmark & 
        \cmark & \xmark \\
        
        \toprule

        $K$& \multicolumn{8}{c}{\it varying subset count}\\
        \midrule
        $2$&
        {\footnotesize \bf 2.11}{\scriptsize $\pm$0.66}  &   {\footnotesize 6.56}{\scriptsize $\pm$0.45}&
        {\footnotesize \bf 0.59}{\scriptsize $\pm$0.17}  &   {\footnotesize 2.19}{\scriptsize $\pm$0.15}&
        {\footnotesize \bf 0.86}{\scriptsize $\pm$0.09}  &   {\footnotesize  0.49}{\scriptsize $\pm$0.06}&
        {\footnotesize \bf 0.24}{\scriptsize $\pm$0.08}  &   {\footnotesize 0.66}{\scriptsize $\pm$0.05}\\

        $3$&
        {\footnotesize \bf 2.67}{\scriptsize $\pm$0.44}  &   {\footnotesize 4.71}{\scriptsize $\pm$0.22}&
        {\footnotesize \bf 0.74}{\scriptsize $\pm$0.13}  &   {\footnotesize 1.52}{\scriptsize $\pm$0.07}&
        {\footnotesize \bf  0.76}{\scriptsize $\pm$0.05}  &   {\footnotesize  0.76}{\scriptsize $\pm$0.03}&
        {\footnotesize \bf 0.29}{\scriptsize $\pm$0.05}  &   {\footnotesize 0.46}{\scriptsize $\pm$0.02}\\

        $5$&
        {\footnotesize \bf 1.25}{\scriptsize $\pm$0.37}  &   {\footnotesize 5.67}{\scriptsize $\pm$0.37}&
        {\footnotesize \bf 0.21}{\scriptsize $\pm$0.09}  &   {\footnotesize 1.89}{\scriptsize $\pm$0.12}&
        {\footnotesize  \bf 0.89}{\scriptsize $\pm$0.06}  &   {\footnotesize  0.62}{\scriptsize $\pm$0.05}&
        {\footnotesize \bf 0.08}{\scriptsize $\pm$0.03}  &   {\footnotesize 0.57}{\scriptsize $\pm$0.04}\\

    \bottomrule
    \end{tabularx}
    \vspace{-10pt}
\end{table}

\subsection{DAGs: D-Struct vs NOTEARS}\label{app:notearsdags}

\begin{figure*}
    \centering
    \begin{subfigure}[t]{.3\textwidth}
        \includegraphics[width=\textwidth]{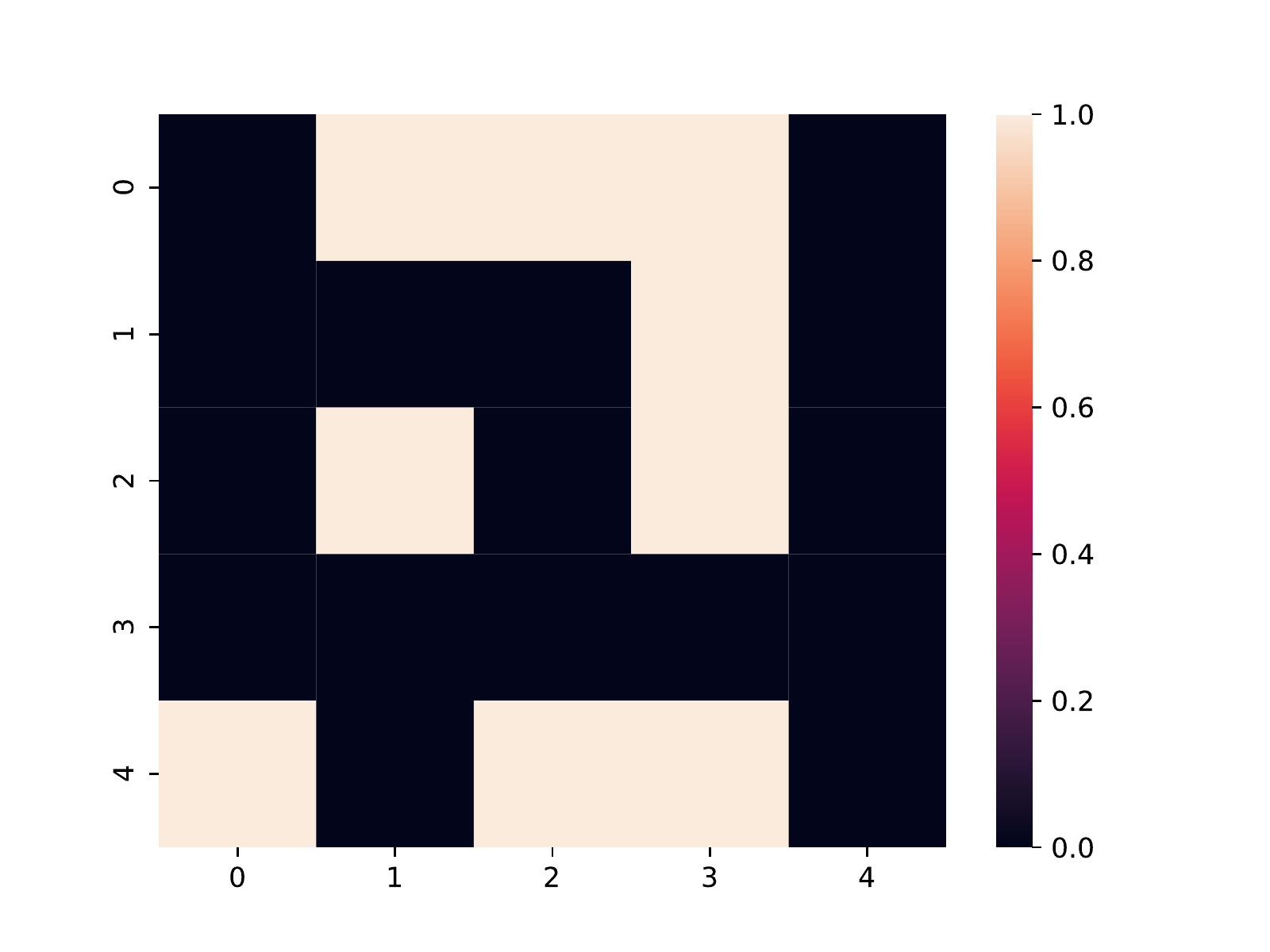}
        \caption{True DAG.}
    \end{subfigure}%
    ~
    \begin{subfigure}[t]{.3\textwidth}
        \includegraphics[width=\textwidth]{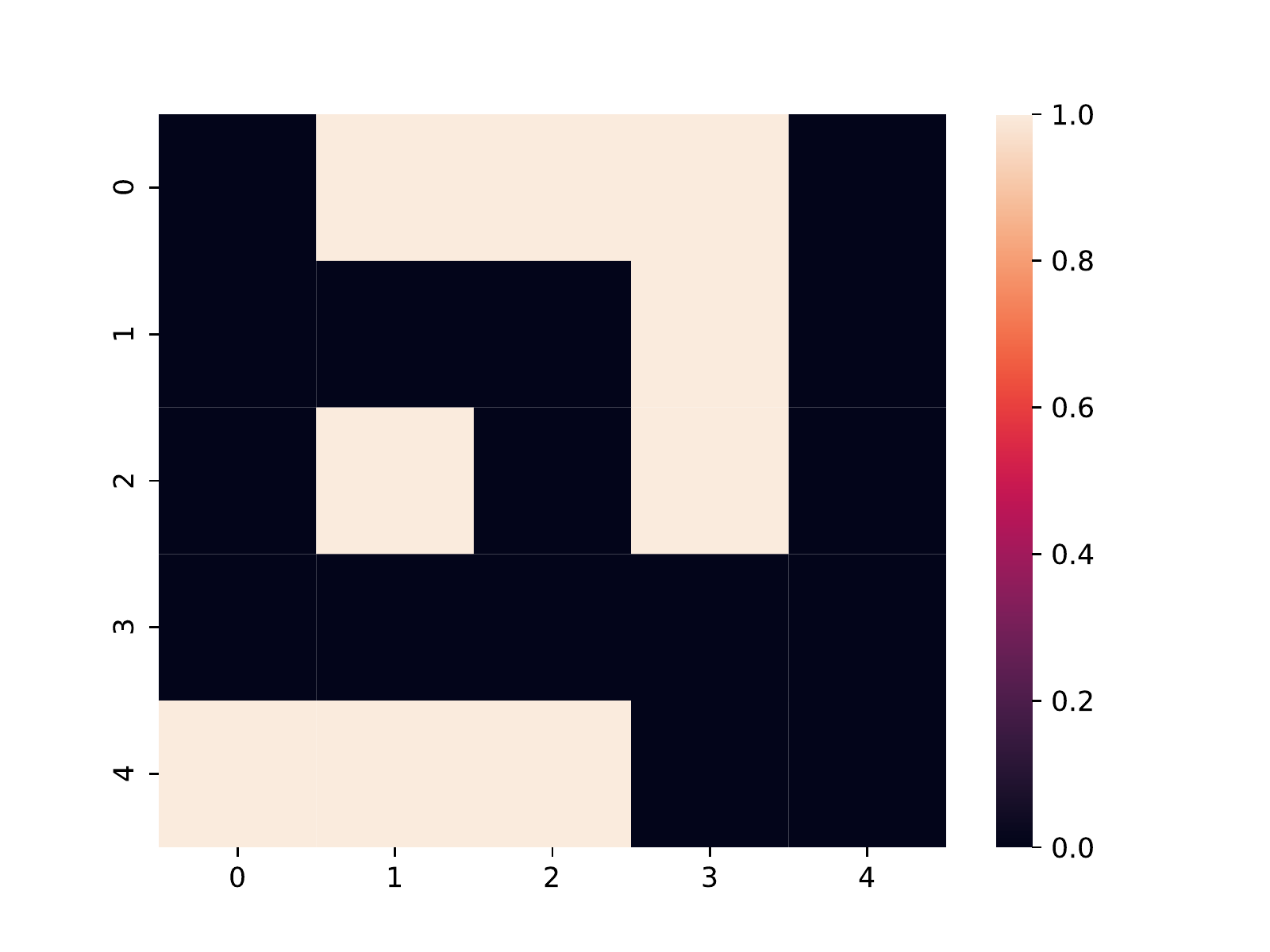}
        \caption{D-Struct estimated DAG.\\ SHD=2, FDR=0.11, FPR=1.}
    \end{subfigure}%
    ~
    \begin{subfigure}[t]{.3\textwidth}
        \includegraphics[width=\textwidth]{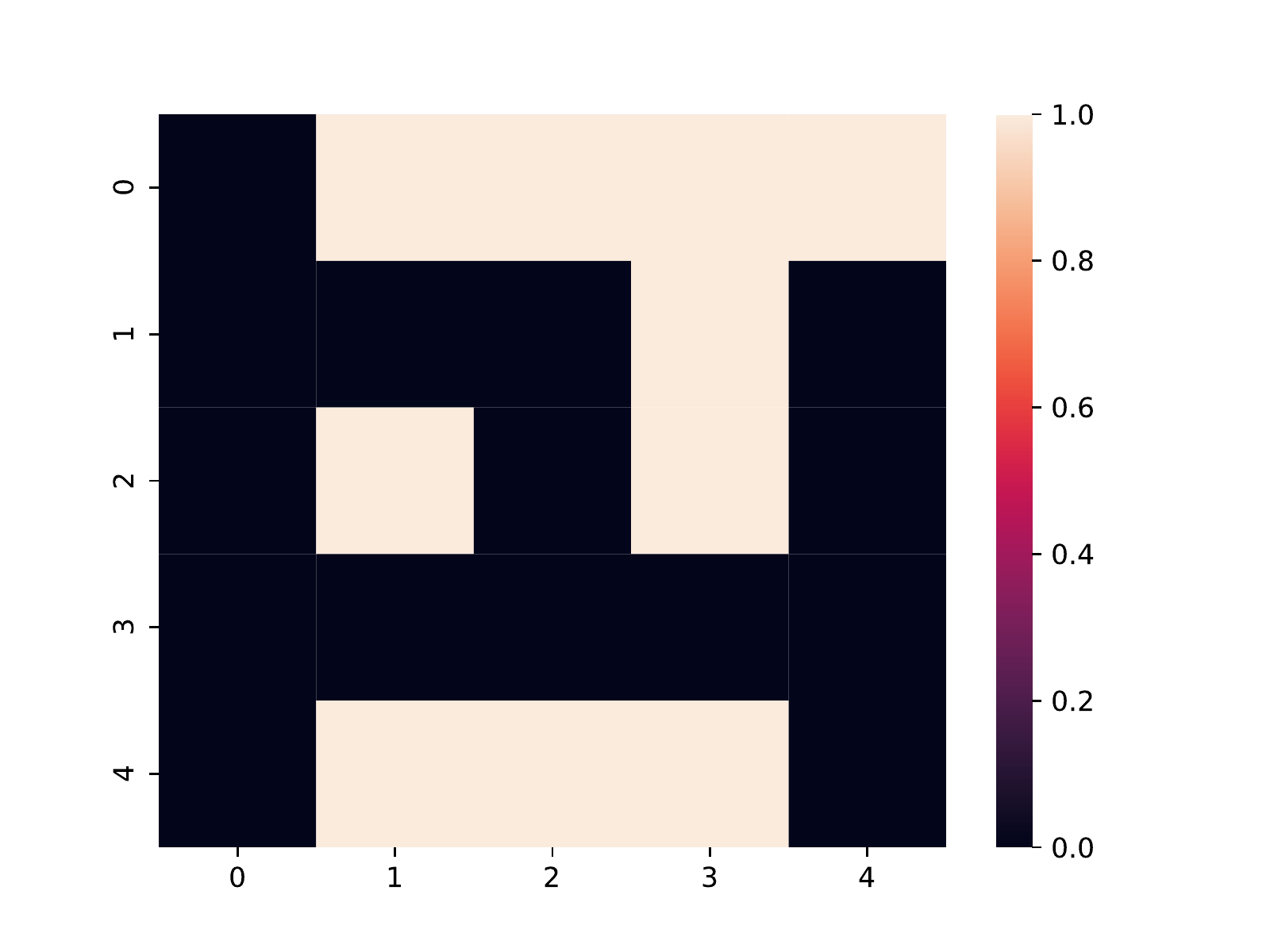}
        \caption{NOTEARS estimated DAG.\\ SHD=2, FDR=0.2, FPR=2.}
    \end{subfigure}%
    \caption{First independent run}\label{fig:compare:1}
    \rule{\textwidth}{.5pt}

\end{figure*}

\begin{figure*}[!h]
\centering
    \begin{subfigure}[t]{.3\textwidth}
        \includegraphics[width=\textwidth]{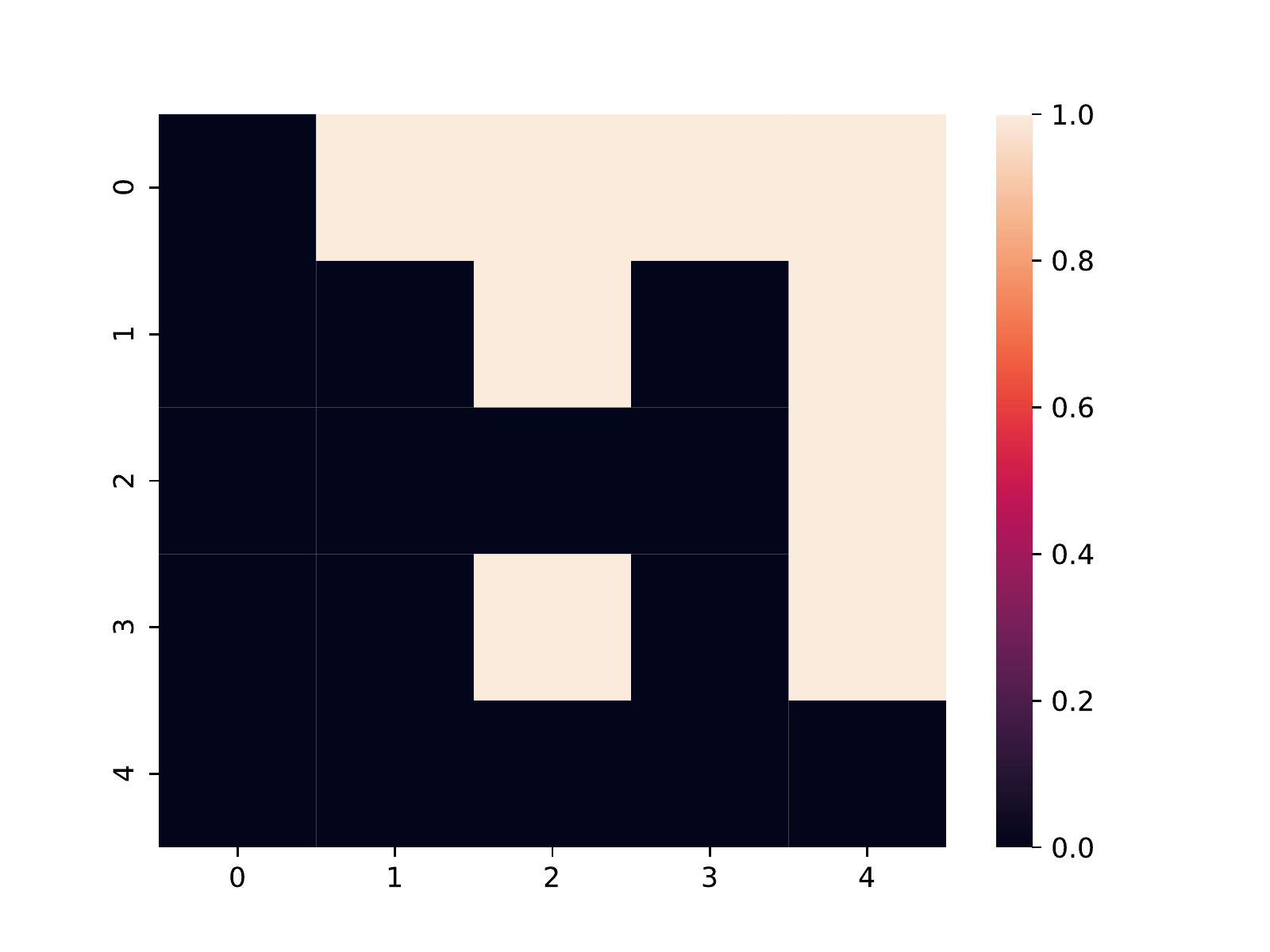}
        \caption{True DAG.}
    \end{subfigure}%
    ~
    \begin{subfigure}[t]{.3\textwidth}
        \includegraphics[width=\textwidth]{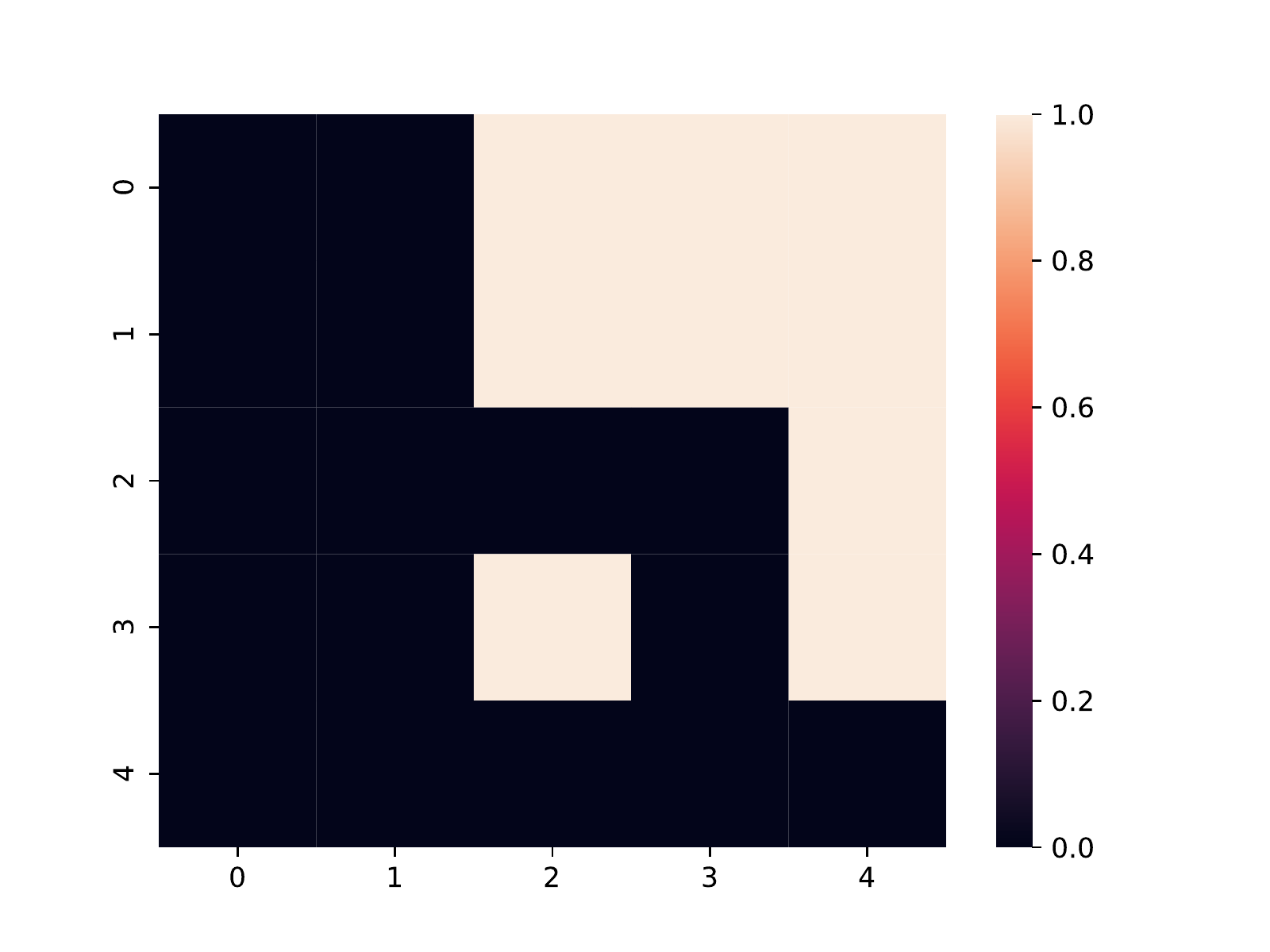}
        \caption{D-Struct estimated DAG.\\ SHD=2, FDR=0.11, FPR=1.}
    \end{subfigure}%
    ~
    \begin{subfigure}[t]{.3\textwidth}
        \includegraphics[width=\textwidth]{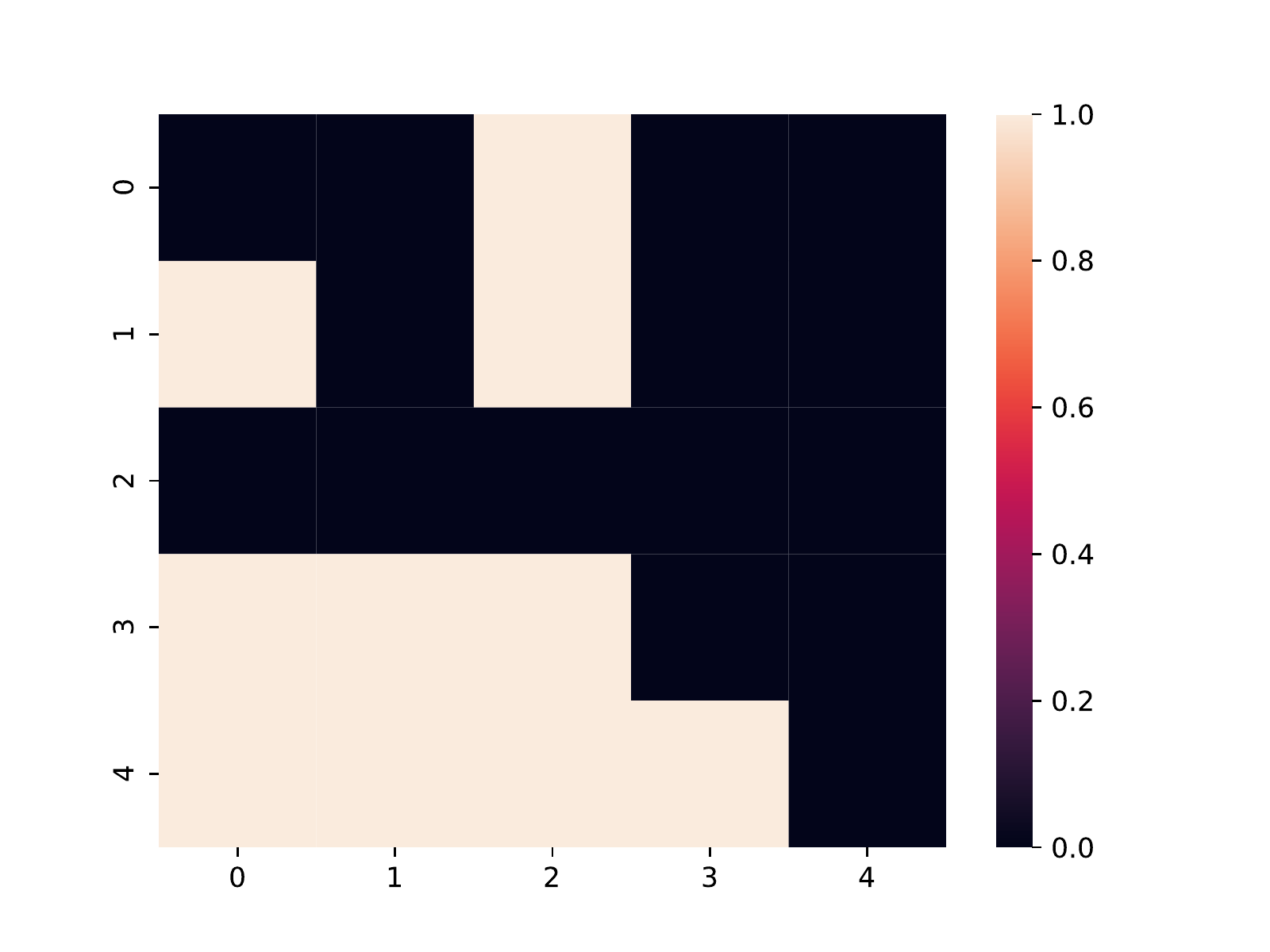}
        \caption{NOTEARS estimated DAG.\\ SHD=2, FDR=0.2, FPR=2.}
    \end{subfigure}%
    \caption{Second independent run}\label{fig:compare:2}
    \rule{\textwidth}{.5pt}

\end{figure*}

We wish to also highlight that indeed what is recovered by D-Struct is different from NOTEARS. For this, we refer to \cref{fig:compare:1,fig:compare:2}, each representing an independent run.

\begin{figure*}
\vspace{-40pt}
    \centering
    
    \begin{subfigure}{0.25\textwidth}
        \includegraphics[width=\textwidth]{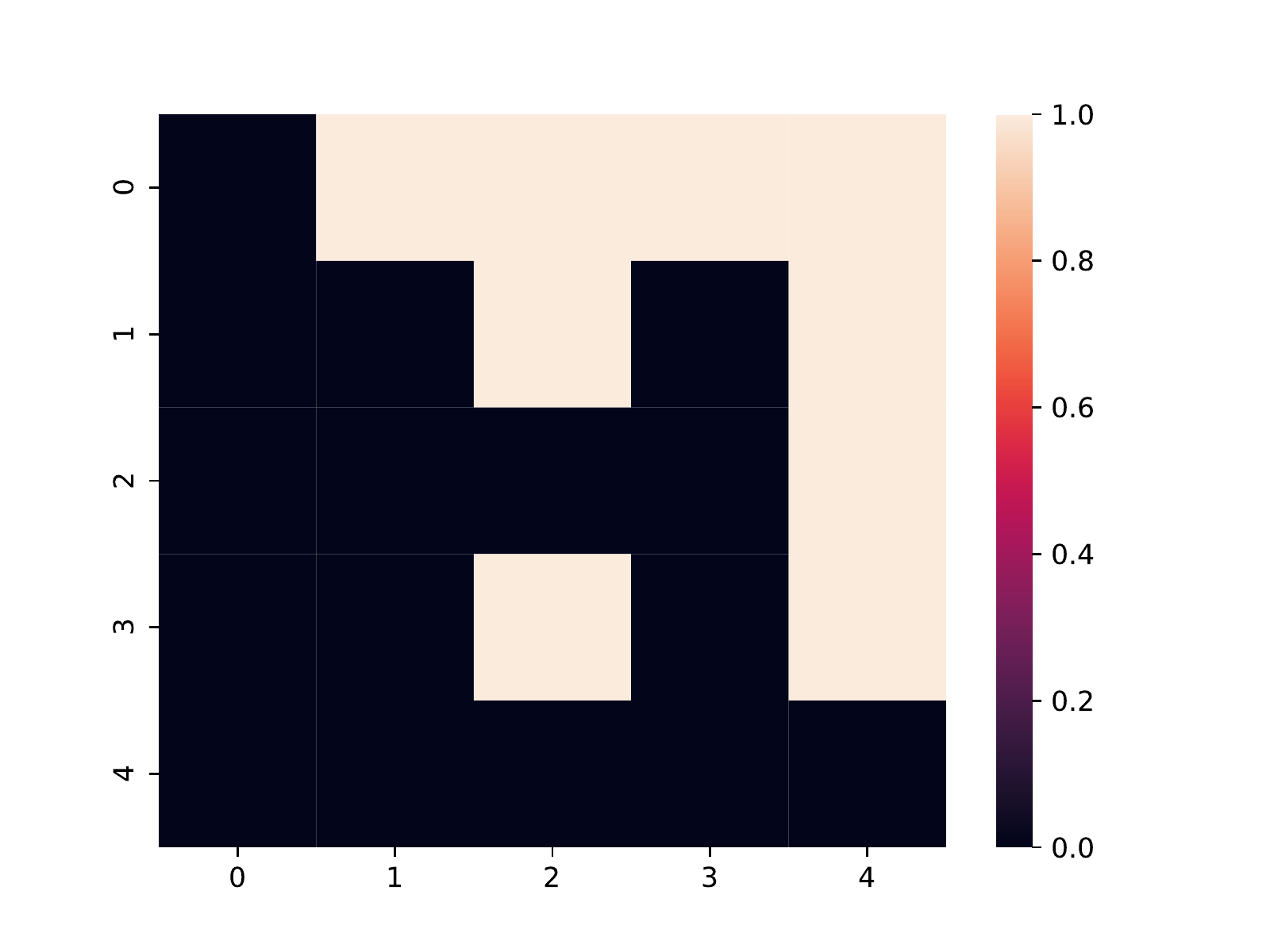}
        \caption{True DAG.}
    \end{subfigure}%
    
    \begin{subfigure}[t]{.25\textwidth}
        \includegraphics[width=\textwidth]{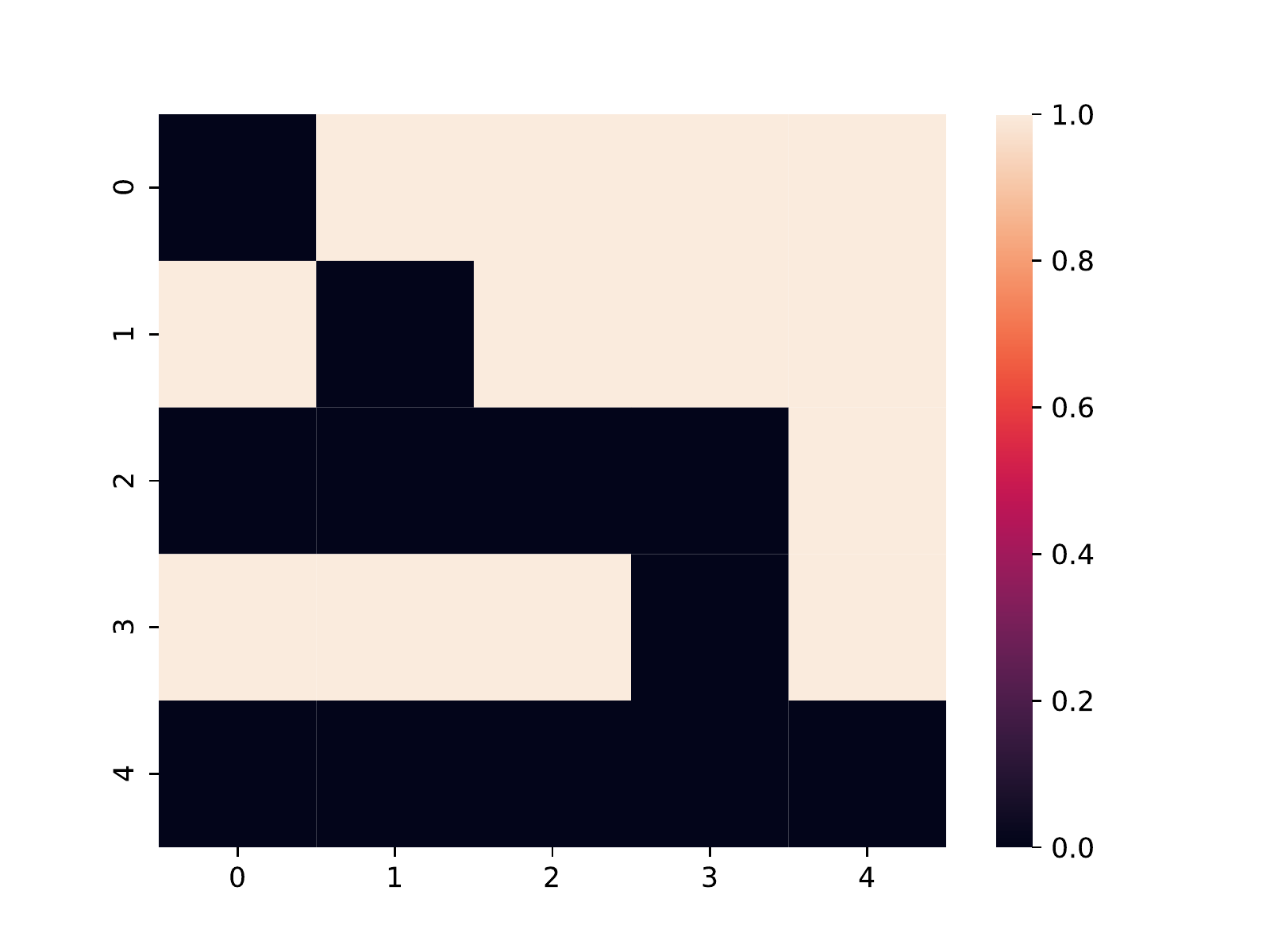}
        \caption{{\bf NOTEARS}: Mean.\\ This is not a DAG!}
    \end{subfigure}%
    ~
    \begin{subfigure}[t]{.25\textwidth}
        \includegraphics[width=\textwidth]{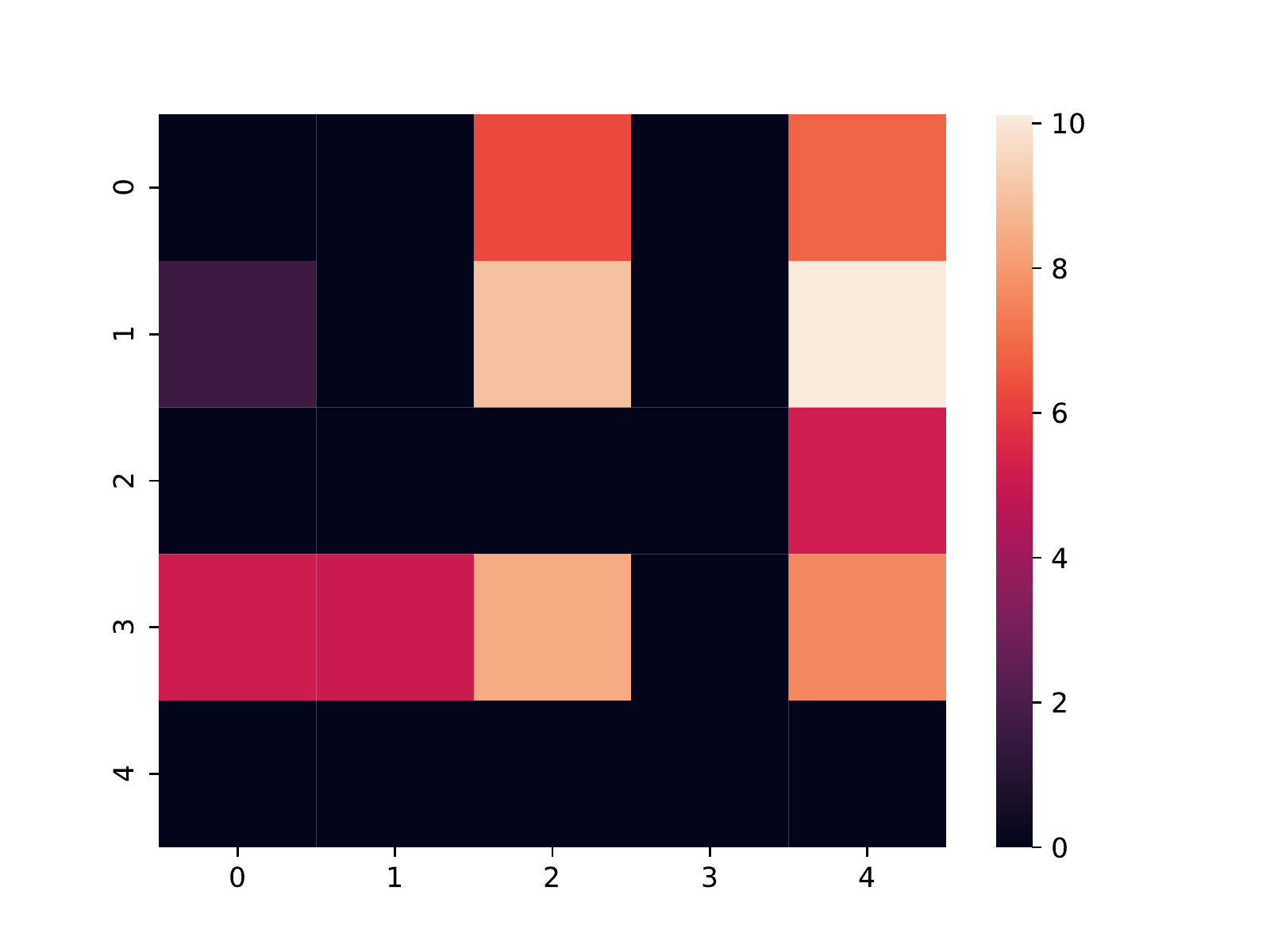}
        \caption{{\bf NOTEARS}: 1\textsuperscript{st} DAG.}
    \end{subfigure}%
    ~
    \begin{subfigure}[t]{.25\textwidth}
        \includegraphics[width=\textwidth]{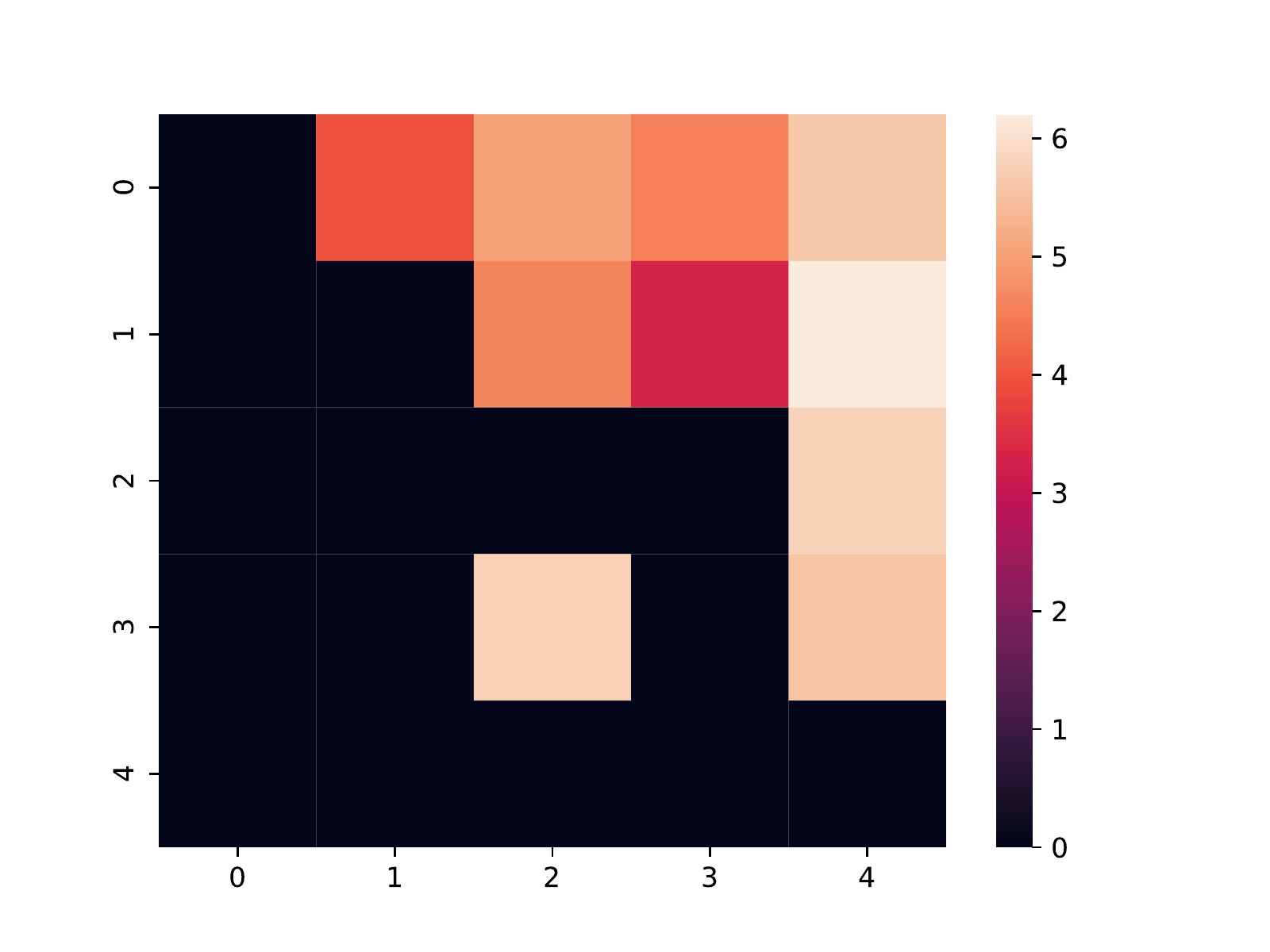}
        \caption{{\bf NOTEARS}: 2\textsuperscript{nd} DAG.}
    \end{subfigure}%
    ~
    \begin{subfigure}[t]{.25\textwidth}
        \includegraphics[width=\textwidth]{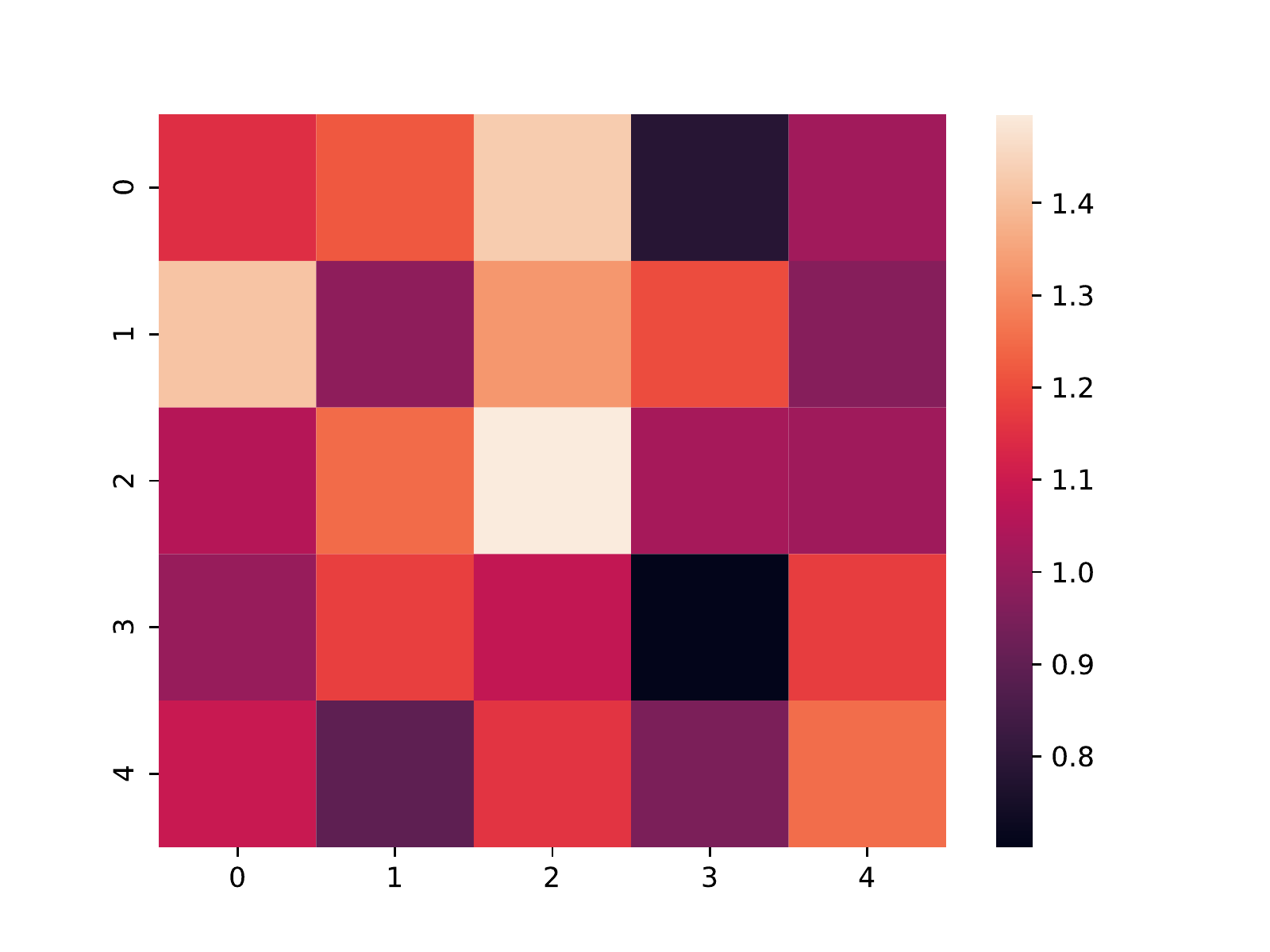}
        \caption{{\bf NOTEARS}: 3\textsuperscript{rd} DAG.}
    \end{subfigure}%
    
    \begin{subfigure}[t]{.25\textwidth}
        \includegraphics[width=\textwidth]{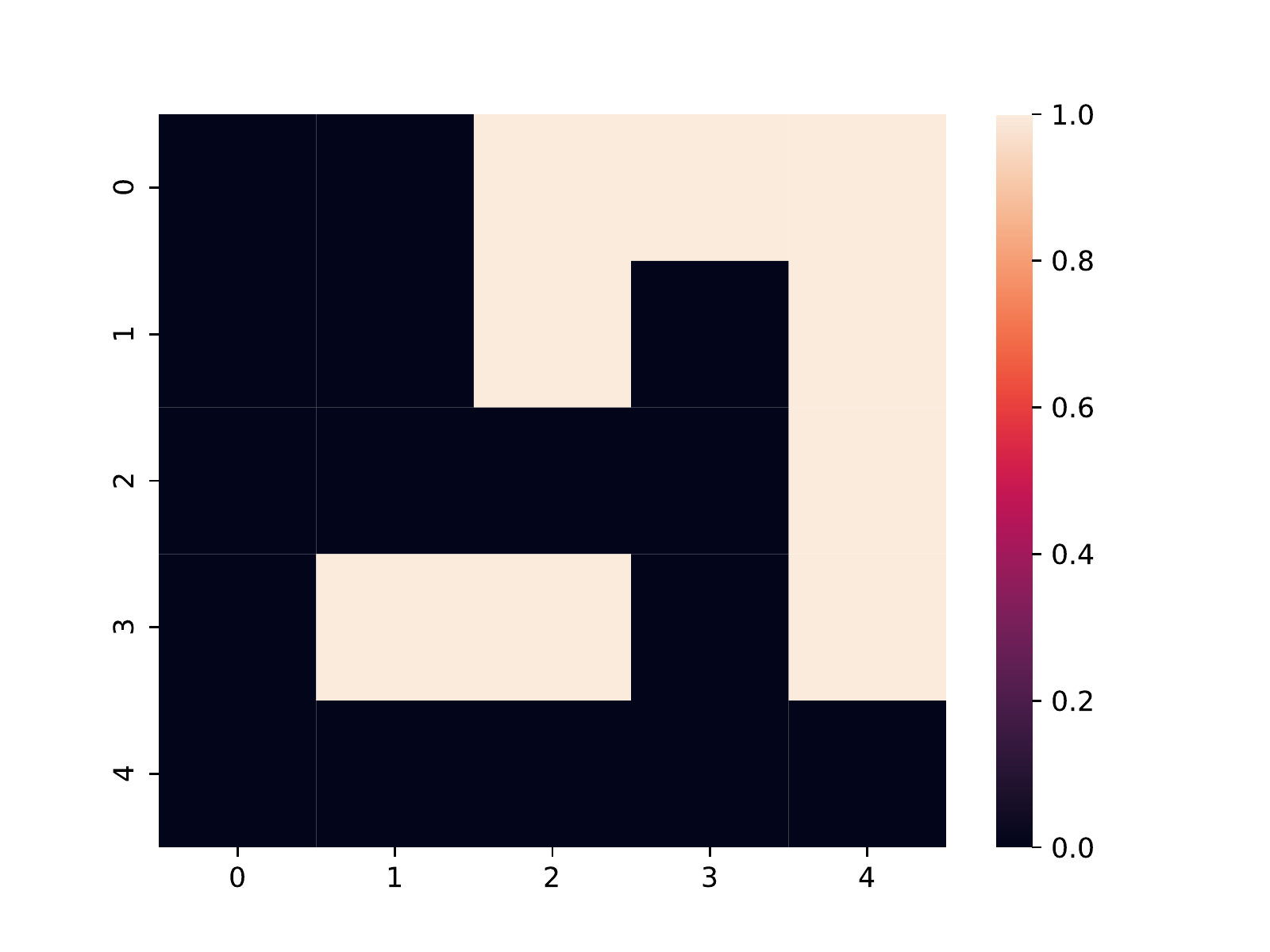}
        \caption{{\bf D-Struct}: Mean.\\ This {\bf is} a DAG!}
    \end{subfigure}%
    ~
    \begin{subfigure}[t]{.25\textwidth}
        \includegraphics[width=\textwidth]{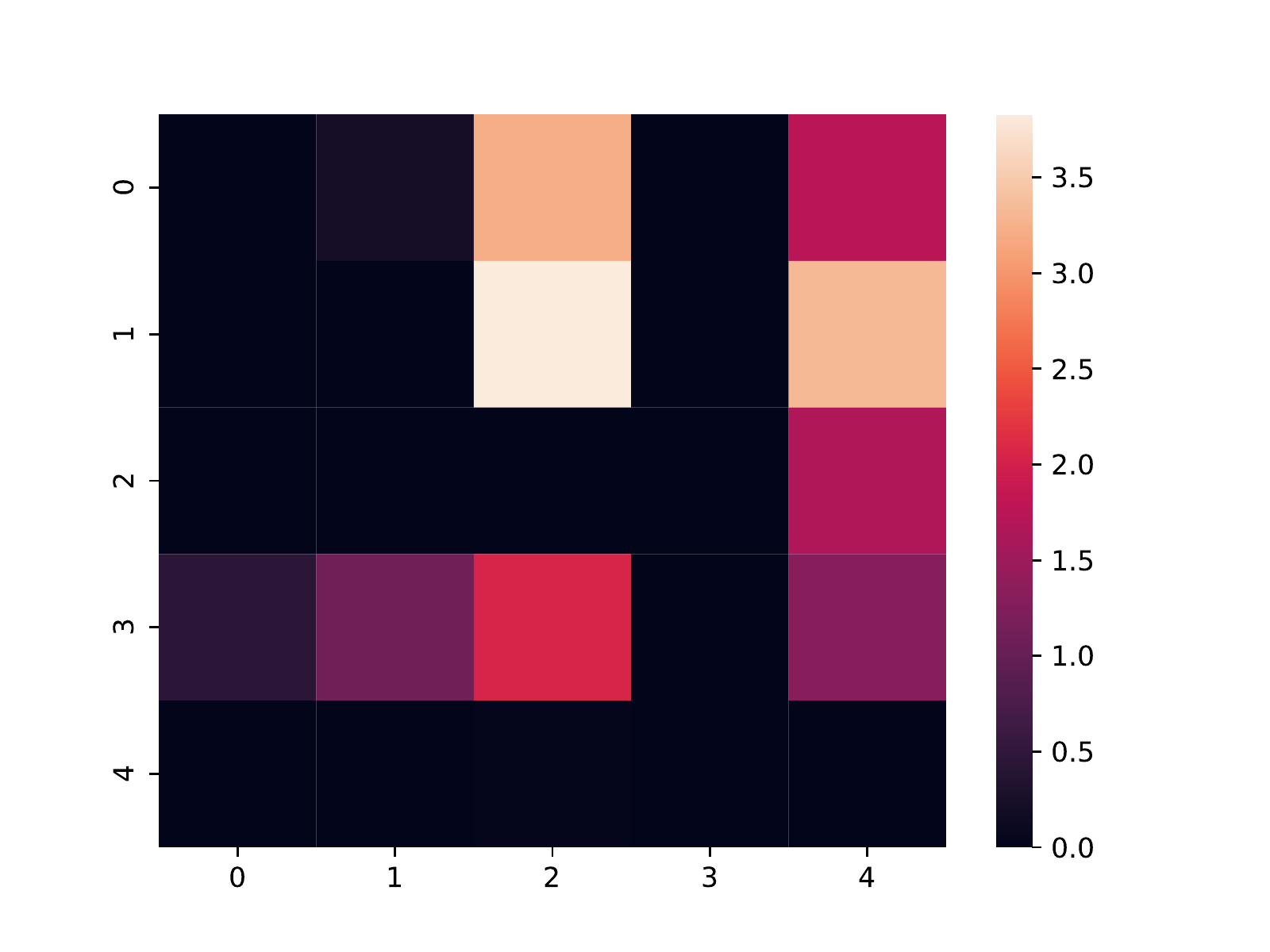}
        \caption{{\bf D-Struct}: 1\textsuperscript{st} DAG.}
    \end{subfigure}%
    ~
    \begin{subfigure}[t]{.25\textwidth}
        \includegraphics[width=\textwidth]{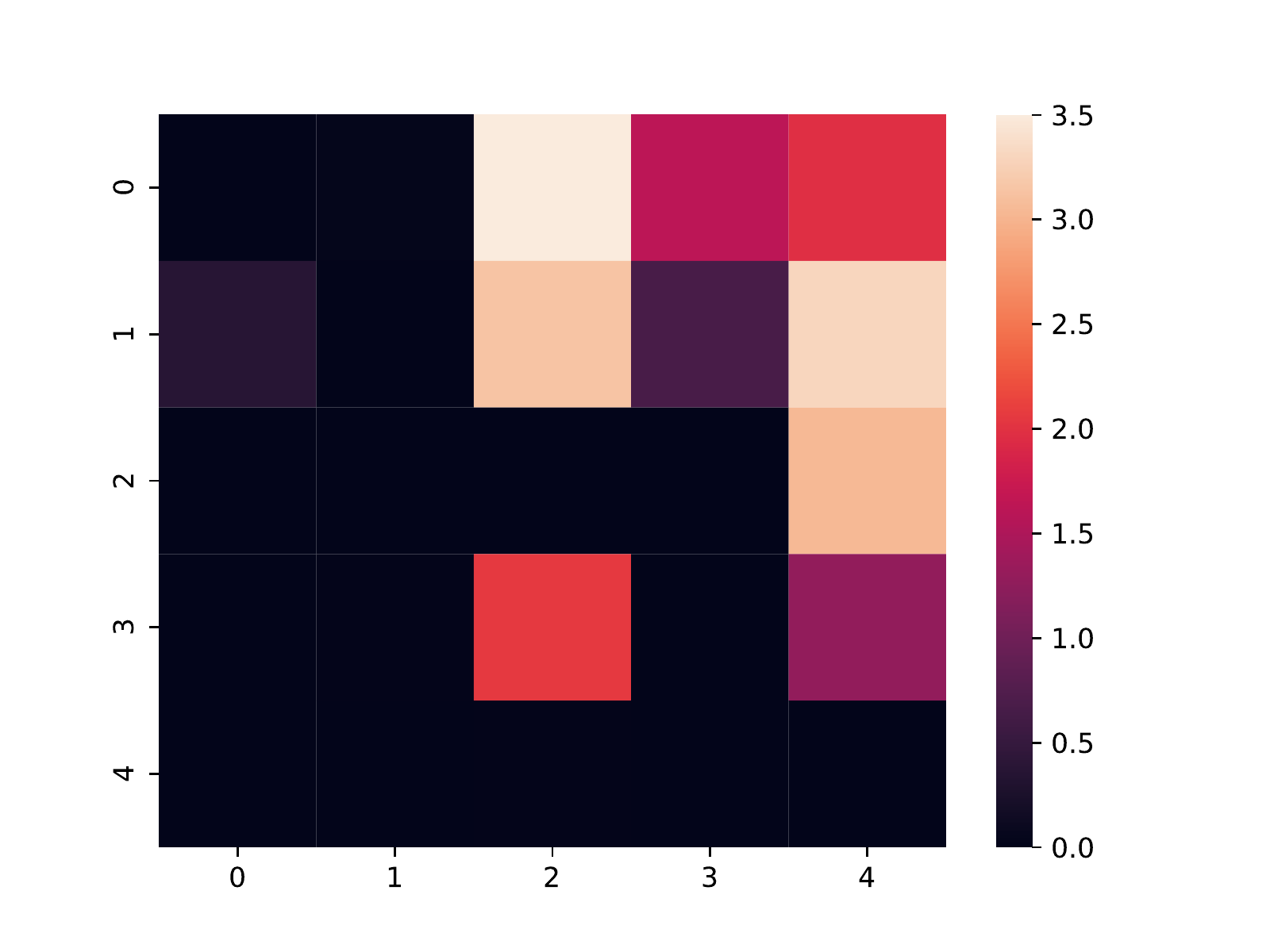}
        \caption{{\bf D-Struct}: 2\textsuperscript{nd} DAG.}
    \end{subfigure}%
    ~
    \begin{subfigure}[t]{.25\textwidth}
        \includegraphics[width=\textwidth]{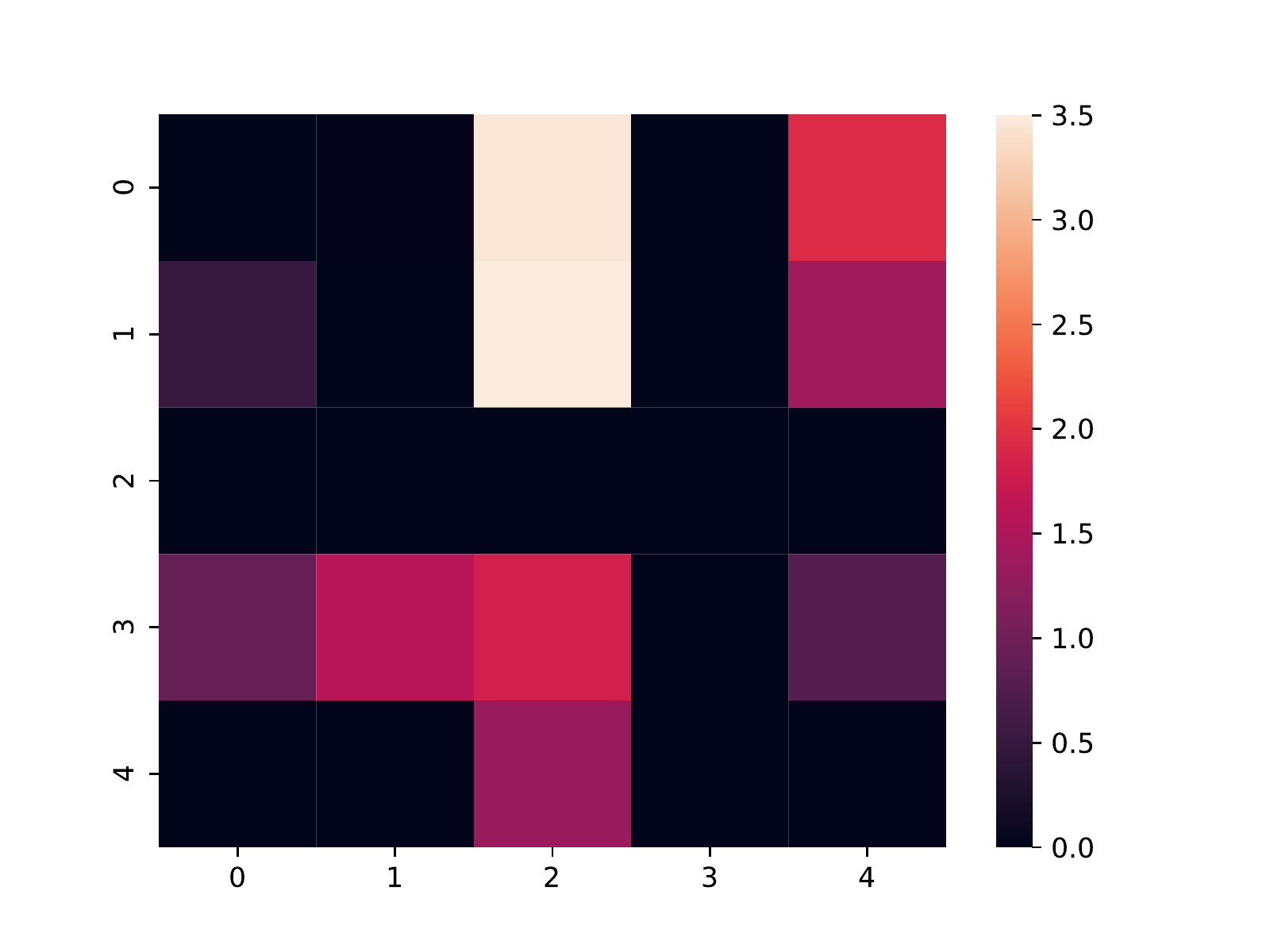}
        \caption{{\bf D-Struct}: 3\textsuperscript{rd} DAG.}
    \end{subfigure}%
  
    \caption{{\bf First independent run.} Note the differences between the three DAGS on each partition for NOTEARS (Row 1), the average is also not a DAG. Whereas, for D-Struct note the similarities by enforcing transportability, the average is also a DAG.}
    \rule{\textwidth}{.5pt}
    \label{fig:adj_matrix:1}

\end{figure*}

\begin{figure*}
    \vspace{-10pt}
    \centering
    \begin{subfigure}[t]{.25\textwidth}
        \includegraphics[width=\textwidth]{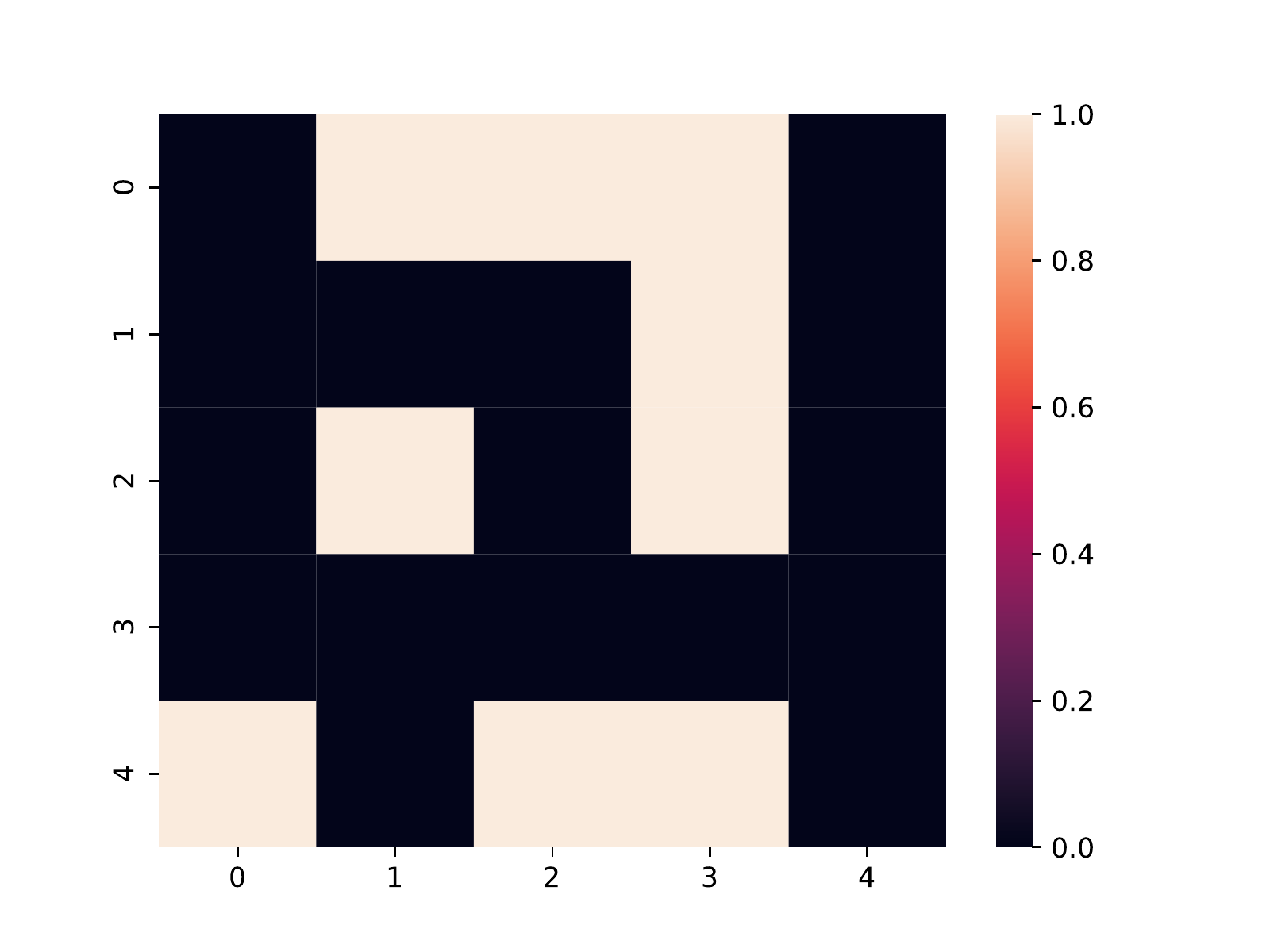}
        \caption{True DAG.}
    \end{subfigure}%

    \begin{subfigure}[t]{.25\textwidth}
        \includegraphics[width=\textwidth]{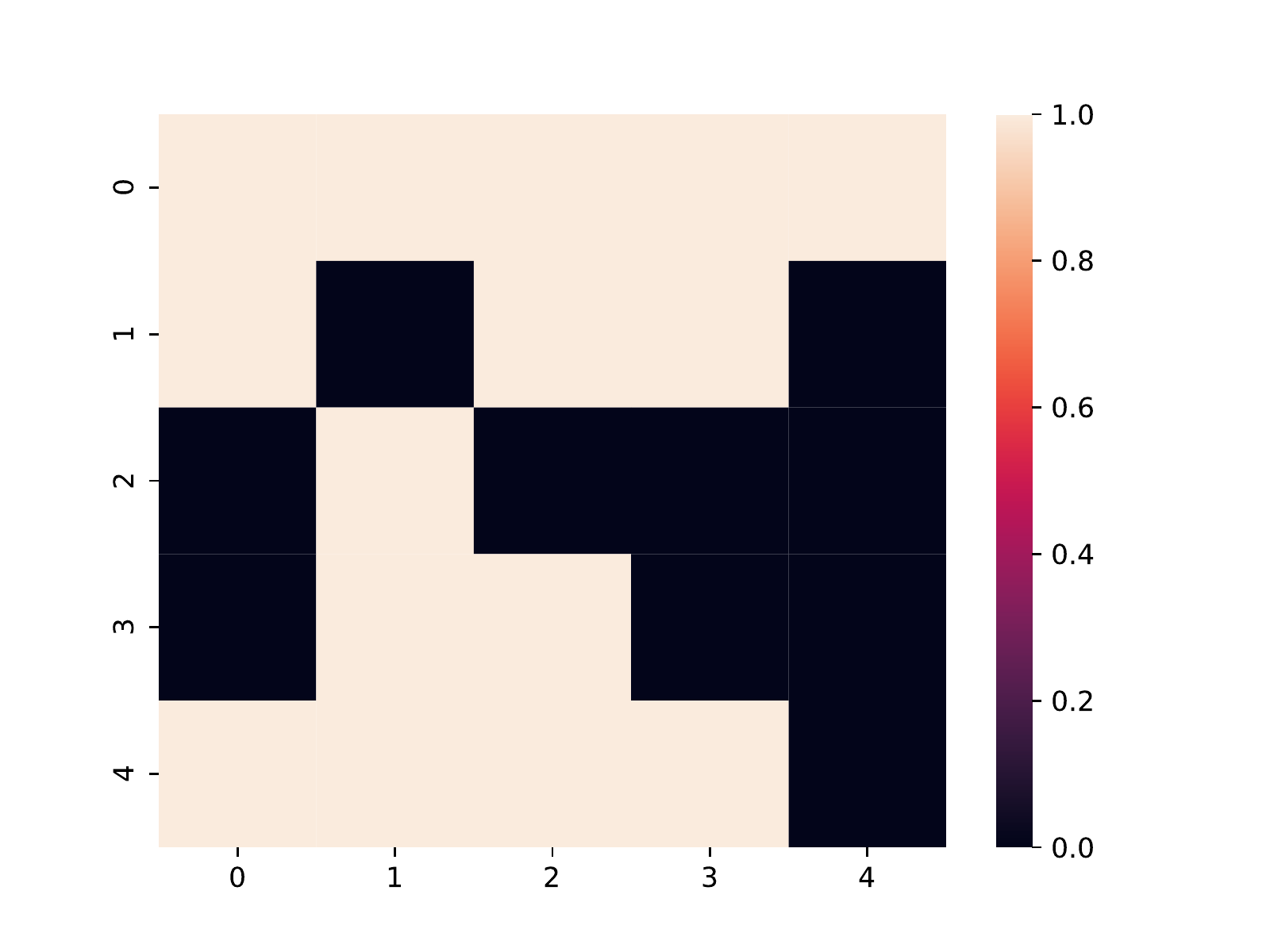}
        \caption{{\bf NOTEARS}: Mean.\\ This is not a DAG!}
    \end{subfigure}%
    ~
    \begin{subfigure}[t]{.25\textwidth}
        \includegraphics[width=\textwidth]{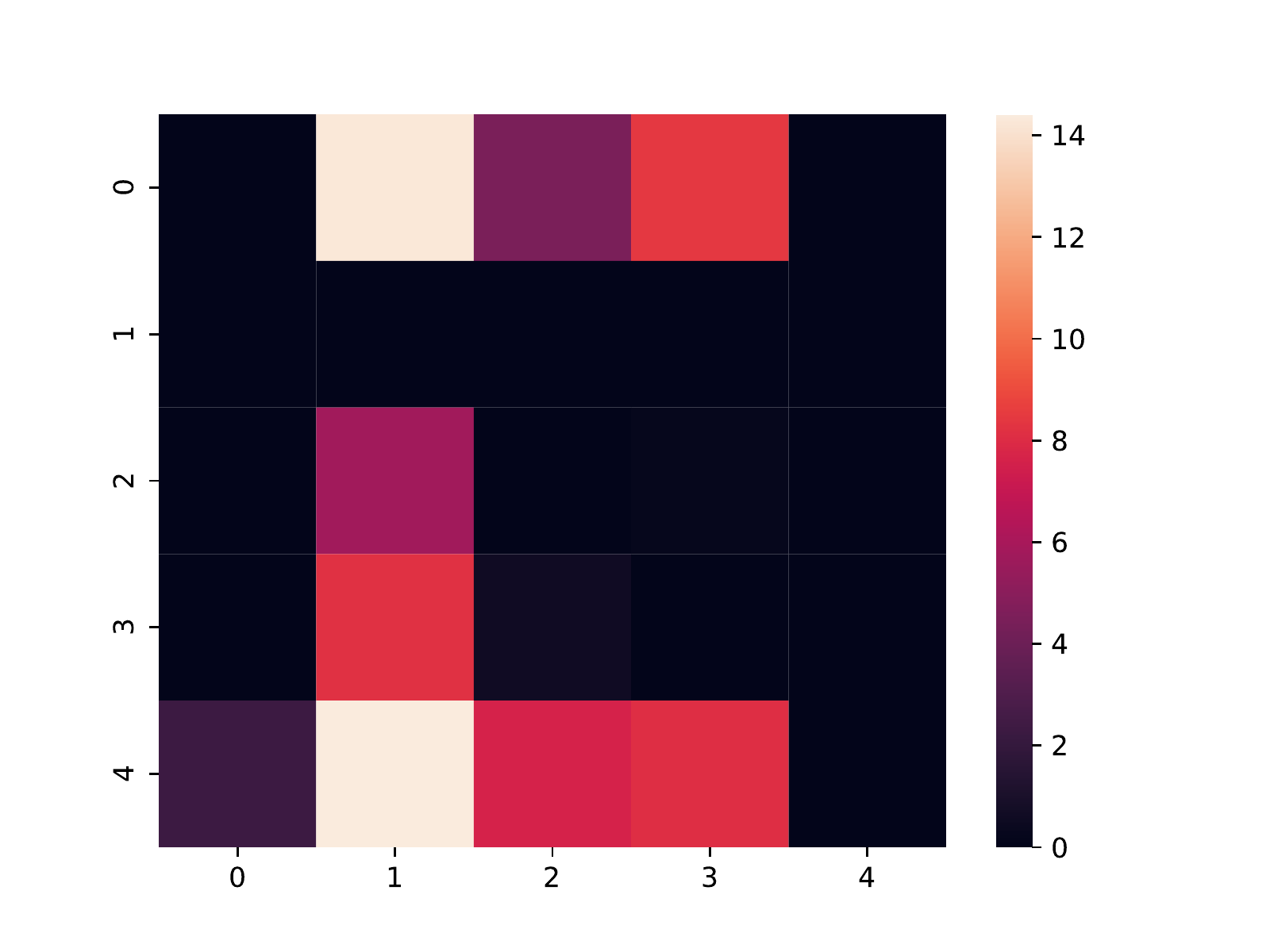}
        \caption{{\bf NOTEARS}: 1\textsuperscript{st} DAG.}
    \end{subfigure}%
    ~
    \begin{subfigure}[t]{.25\textwidth}
        \includegraphics[width=\textwidth]{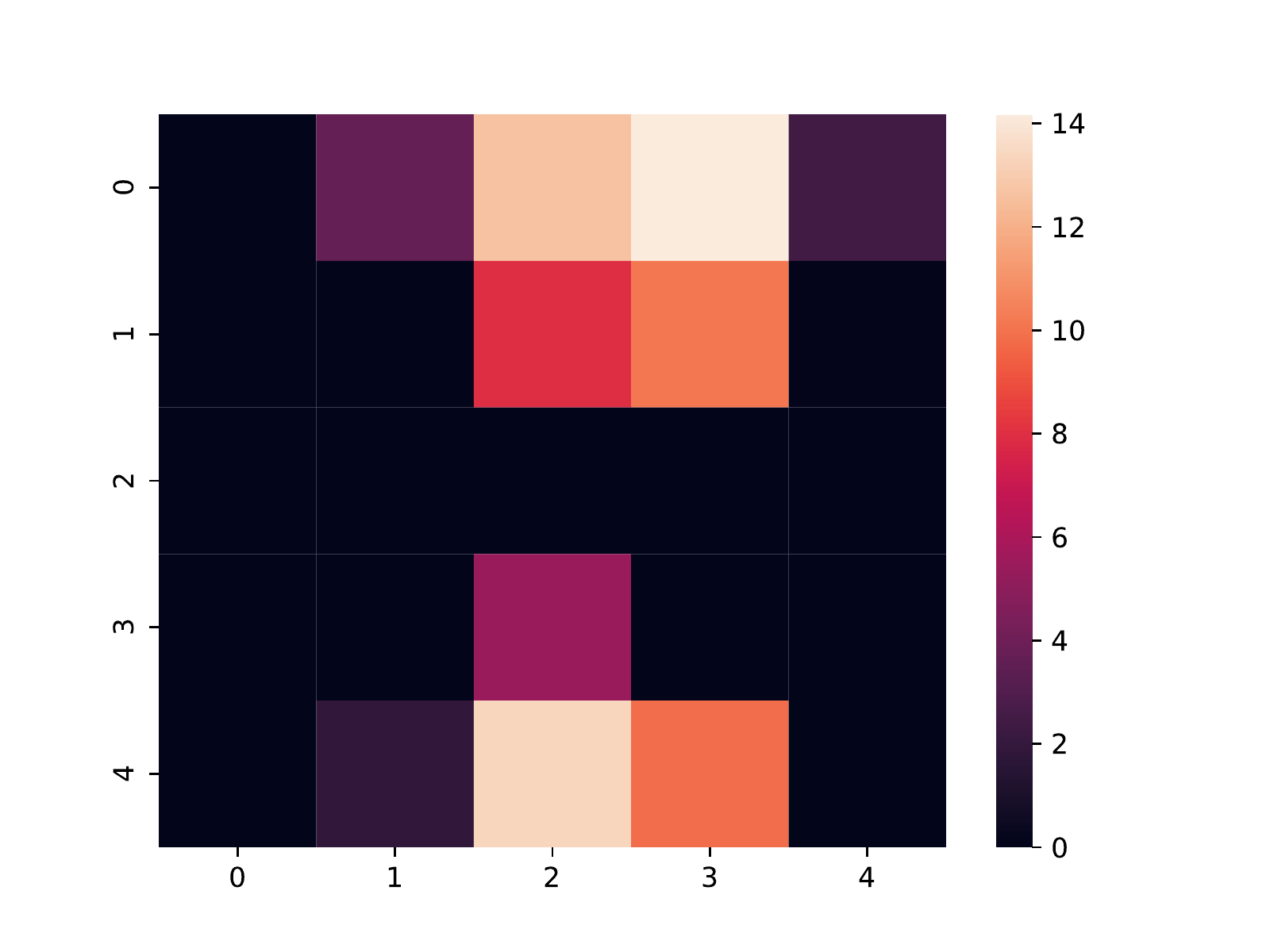}
        \caption{{\bf NOTEARS}: 2\textsuperscript{nd} DAG.}
    \end{subfigure}%
    ~
    \begin{subfigure}[t]{.25\textwidth}
        \includegraphics[width=\textwidth]{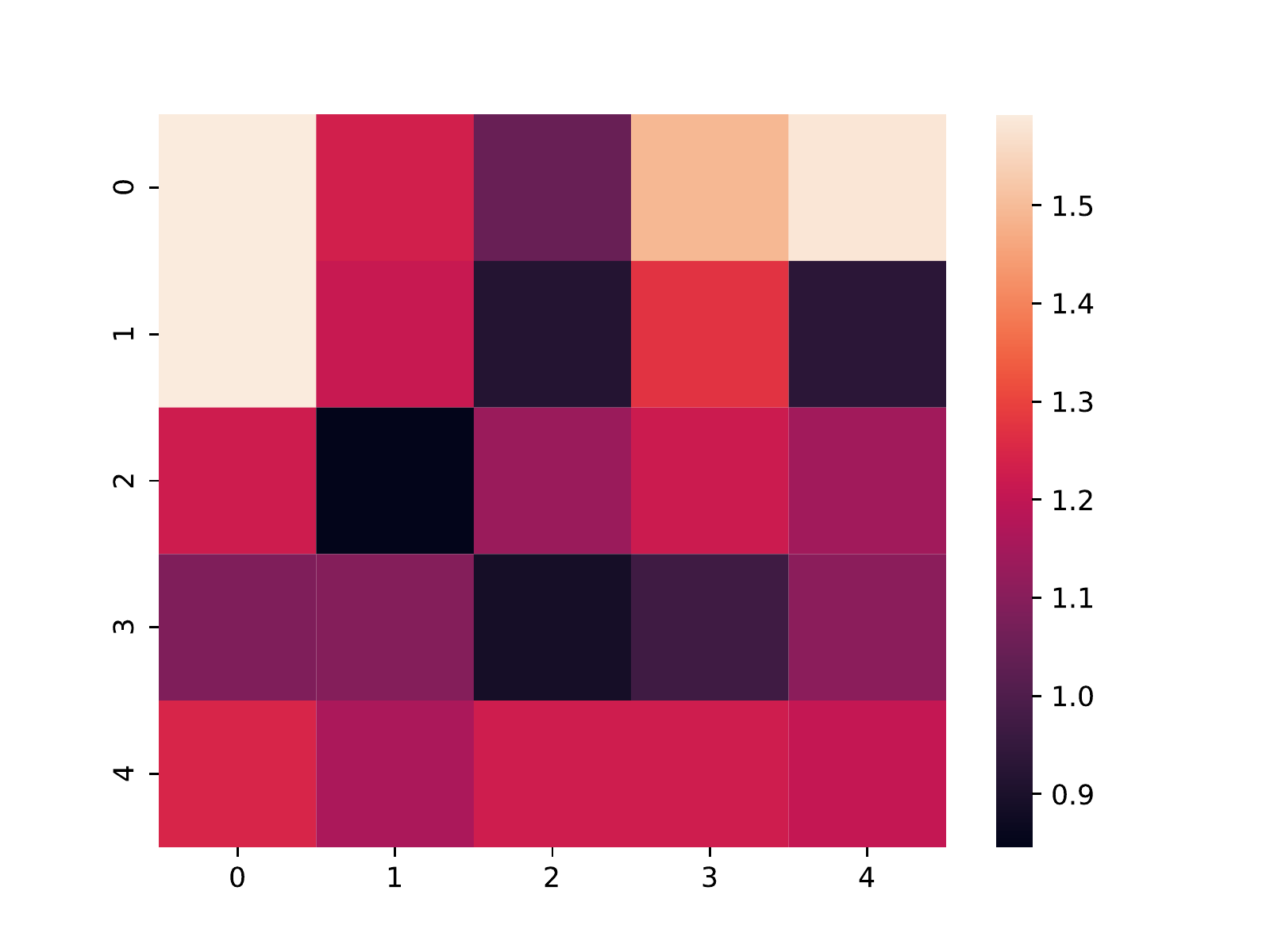}
        \caption{{\bf NOTEARS}: 3\textsuperscript{rd} DAG.}
    \end{subfigure}%
    
    \begin{subfigure}[t]{.25\textwidth}
        \includegraphics[width=\textwidth]{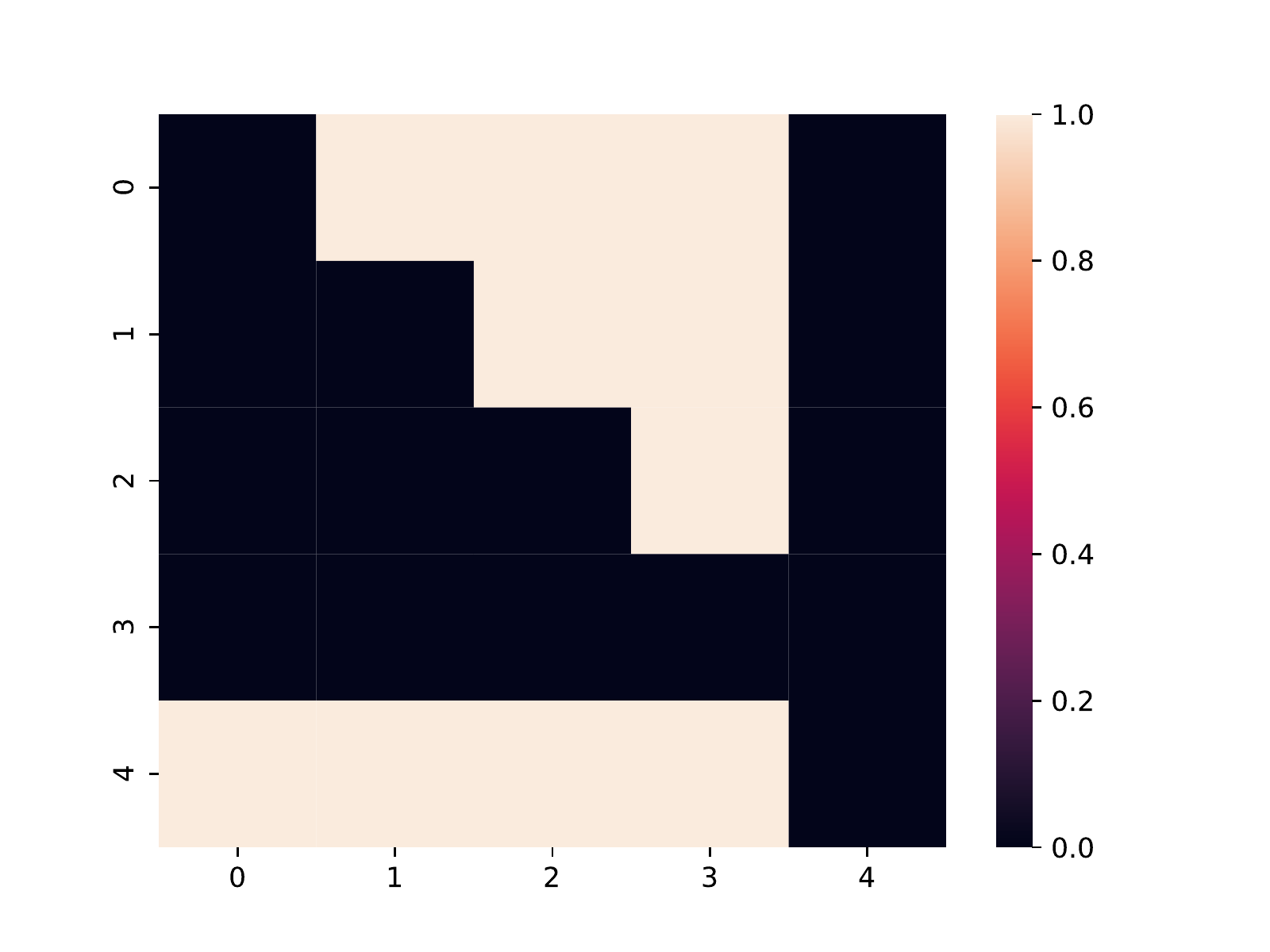}
        \caption{{\bf D-Struct}: Mean.\\ This {\bf is} a DAG!}
    \end{subfigure}%
    ~
    \begin{subfigure}[t]{.25\textwidth}
        \includegraphics[width=\textwidth]{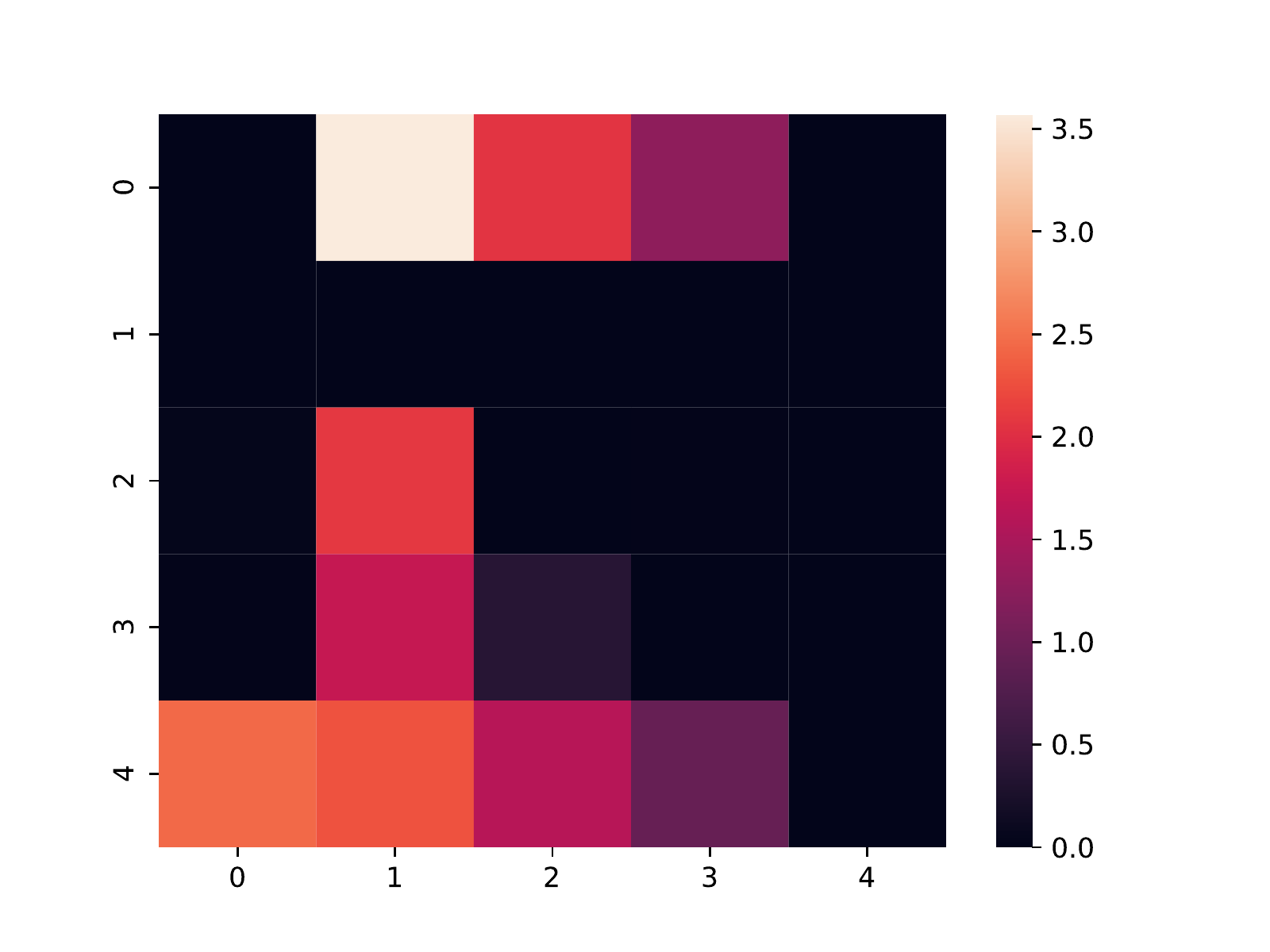}
        \caption{{\bf D-Struct}: 1\textsuperscript{st} DAG.}
    \end{subfigure}%
    ~
    \begin{subfigure}[t]{.25\textwidth}
        \includegraphics[width=\textwidth]{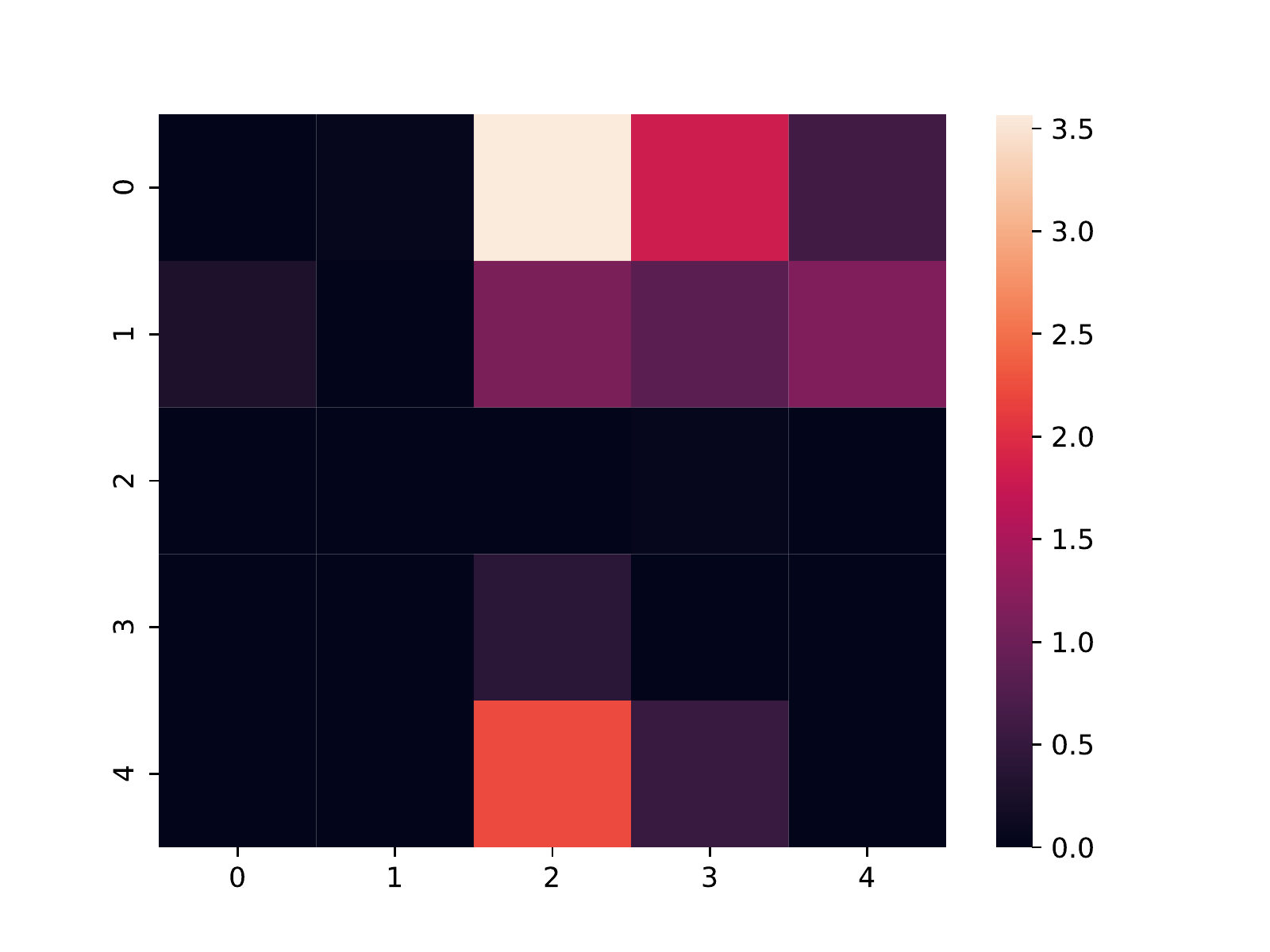}
        \caption{{\bf D-Struct}: 2\textsuperscript{nd} DAG.}
    \end{subfigure}%
    ~
    \begin{subfigure}[t]{.25\textwidth}
        \includegraphics[width=\textwidth]{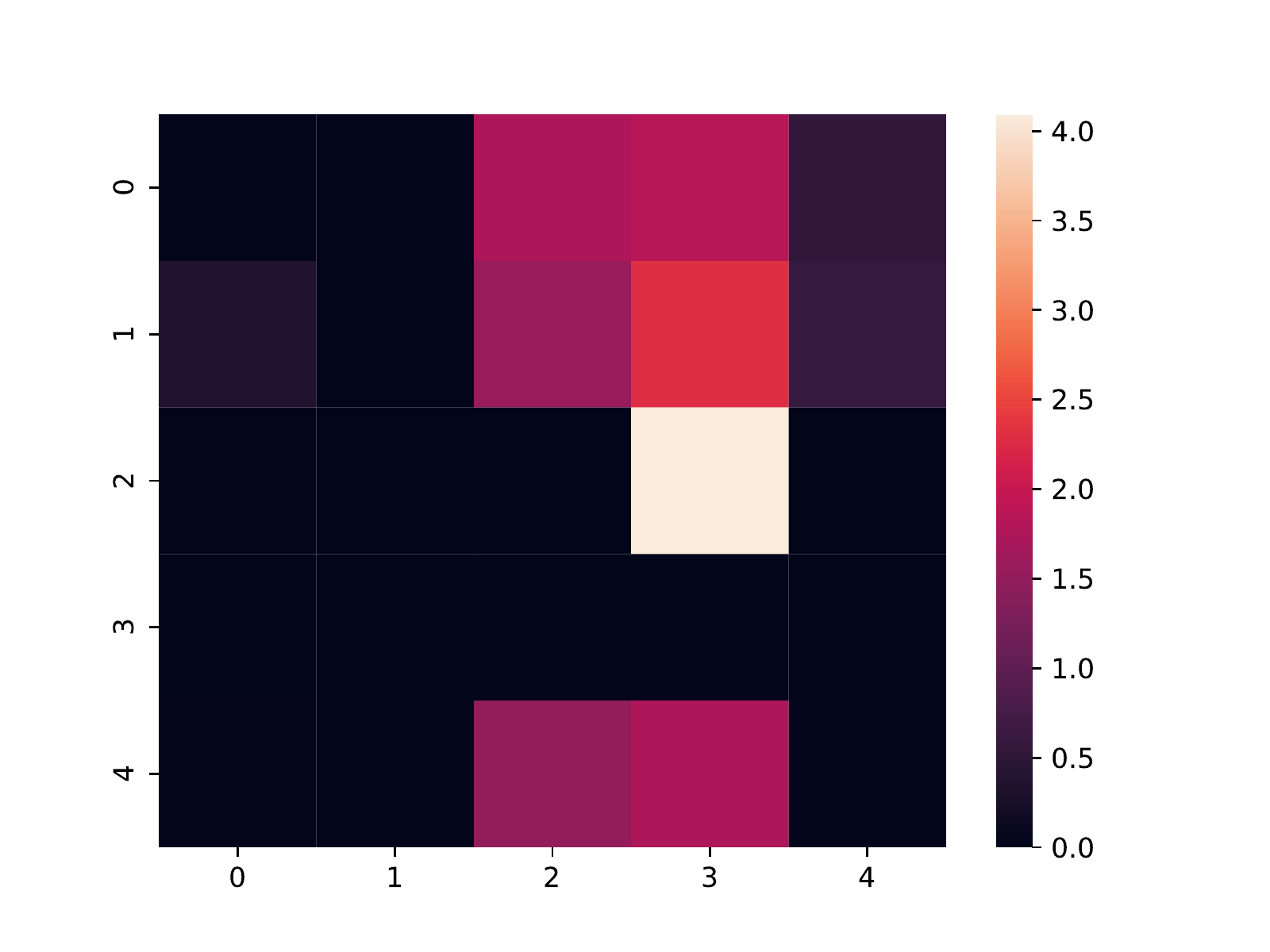}
        \caption{{\bf D-Struct}: 3\textsuperscript{rd} DAG.}
    \end{subfigure}%

    \caption{{\bf Second independent run.} Note the differences between the three DAGS on each partition for NOTEARS (Row 1), the average is also not a DAG. Whereas, for D-Struct note the similarities by enforcing transportability, the average is also a DAG.}
    \label{fig:adj_matrix:2}
    \rule{\textwidth}{.5pt}
\end{figure*}

\subsection{Gains from enforcing transportability}

A key concept of D-Struct is to enforce transportability, which is done using our novel loss function. 

\begin{equation*}
    \mathcal{L}(\mathcal{G}_k | \mathcal{D}_k) \coloneqq \mathcal{L}_\text{DSF}(\mathcal{G} | \mathcal{D}_k) + \alpha \mathcal{L}_\text{MSE}(A(\mathcal{G}_k)),
\end{equation*}

The question is what do we gain from the usage of the $\alpha$ term which is key to enforcing transportability. We conduct an experiment where we set $\alpha=0$. This not only assesses the importance of this term, but also without $\mathcal{L}_\text{MSE}$ this amounts to assessing $K$ \textit{independent} versions of vanilla NOTEARS. 

\textit{Results:} When we combine the $K$ DAGs by averaging them, the result is NOT a DAG. 

This highlights that indeed that (1) transportability is key as part of this formulation and (2) that simply running parallel versions of NOTEARS is not a sufficient solution.

We highlight this by showing the independent DAGs discovered without transportability enforced, the average of the DAGs and the true DAG. These results are reported in \cref{fig:adj_matrix:1,fig:adj_matrix:2}

\section{Causal interpretation and uniqueness} \label{app:causality}

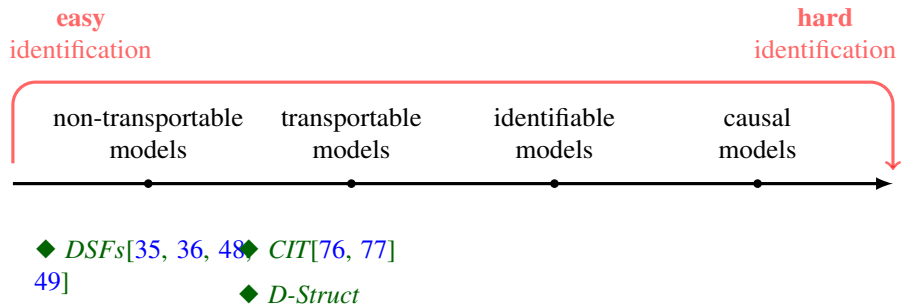
\begin{figure}
    \centering
    \begin{tikzpicture}[
            dot/.style={circle, fill=black, draw=black,  minimum size=1mm, inner sep=0},
            scale=.9,
            annotation/.style={very thick, red, opacity=.6, align=center},
        ]
        
        \draw[-latex, very thick] (1,0) -- (14,0);
        
        \node[dot,] at (3, 0) {};
        \node[text width=3cm, align=center] at (3, .75) {non-transportable models};
        \node[text width=3cm, align=left, anchor=north, DarkGreen] at (3, -.3) {\point{DSFs}\hspace{-8pt} \citep{zheng2018dags,zheng2020learning,yu19a,yu21a}};
        
        \node[dot,] at (6, 0) {};
        \node[text width=3cm, align=left, anchor=north, DarkGreen] at (6, -.3) {\point{CIT}\hspace{-8pt} \citep{spirtes2000causation,verma1990} \point{D-Struct}};
        \node[text width=2cm, align=center] at (6, .75) {transportable models};
        
        \node[dot,] at (9, 0) {};
        \node[text width=2cm, align=center] at (9, .75) {identifiable models};
        
        \node[dot,] at (12, 0) {};
        \node[text width=2cm, align=center] at (12, .75) {causal models};

        \node[annotation, anchor=south, text width=2cm] (ann_1) at (2, 1.7) {{\bf easy} identification};
        \node[annotation, anchor=south, text width=2cm] (ann_2) at (13, 1.7) {{\bf hard} identification};

        \draw[->, rounded corners=3mm, annotation] (1, .3) -- (1, 1.5) -- (14, 1.5) -- (14, .2);

    \end{tikzpicture}
    \caption{{\bf Comparison of methods w.r.t. identification and uniqueness.} The ultimate goal of structure learning is to come up with unique and correct structures. Once we recover the one true DAG, we may interpret the structure as a causal model. However, discovering a causal structure using only observational data is not possible. Yet, we can {\it approach} it with methods that restrict the set of possible DAGs. From this illustration, we gather that D-Struct is an attempt to restrict the solution space of DSFs, going one step further towards unique solutions.}
    \label{fig:axis}
    \rule{\textwidth}{.5pt}
\end{figure}

{\bf Causality.} Causal relationships between variables are often expressed as DAGs \citep{pearl2009causality}. While D-Struct is able to recover DAGs more reliably, there is actually no guarantee that the found DAG can be interpreted as a causal DAG. There is a simple reason for this: we do not make any additional identification assumptions on the structural equations when learning DAGs, at least not beyond what is already assumed in the used DSFs. Furthermore, should D-Struct be combined with a DSF that {\it is} able to recover a causal DAG\footnote{We know of none that is able to.}, the way in which the $K$ internal DAGs are combined may violate these assumptions (recall DAG combination from \cref{app:experiment-details}).

With D-Struct, we recover a Bayesian network (BN), which is directed, yet the included directions are not necessarily meaningful.  The only guarantee we have with BNs is that they resemble a distribution, which express some conditional distributions (as per the independence sets in \cref{sec:prelims}). Order is not accounted for in these independence sets. For more information regarding this, we refer to \cref{app:defs} and \citet{koller2009probabilistic}. 

However, as is indicated in \citet[Chapter 21]{koller2009probabilistic}, a ``good'' BN structure should correspond to causality, where edges $X \to Y$ indicate that $X$ causes $Y$. \citet{koller2009probabilistic} state that BNs with a causal structure tend to be sparser. Though, if queries remain probabilistic, it doesn't matter whether or not the structure is causal, the answers will remain the same. Only when we are interested in interventional queries (by using do-calculus) do we have to make sure the DAG is a causal one. 

{\bf Uniqueness.} The above is a pragmatic view. To our knowledge, there is no real proof stating that sparser DAGs are (even more likely to be) causal. However, it could offer guidance to try and recover a causal DAG, assuming it to be sparse \citep{notears_sparsity2020}. The latter, of course, is assuming that there exists a {\it unique} or {\it correct} DAG, which is something we implicitly assume to be true. Naturally, when aiming to make a discovery, we aim to recover a {\it true} DAG, where a truthful DAG corresponds with a DAG that can be uniquely recovered.

However, there is a difference between {\it a} unique DAG, and {\it the} unique DAG. Where the former is a matter of identifiability (discussed more below), the latter is one of causality. With the latter we mean: ``can a method actually recover the unique causal DAG?'' From \citet{meek1995strong} and \citet{meek1995strong} we learn that, from observational data alone, this is impossible and should thus not be a goal if one is not willing to make additional assumptions.

We stress that transportability is a weaker goal than identifiability. Enforcing transportability does not guarantee unique or repeatable results. Take CIT-based methods--- which we know to be fully transportable. While it is true that the same set of independence statements will always result in the same DAG (i.e. transportability), it is not necessarily true that we will always recover the same independence statements. Depending on which independence test one uses to build the set of independence statements, the resulting DAG may look entirely different. Similarly for D-Struct, while D-Struct does encourage similar DAGs (see for example \cref{app:additional-experiments}), we have no guarantee to recover the {\it same} DAG over different runs. The latter is a requirement for identifiability \citep{walter2014identifiability} as identifiability requires the model to always converge to the same set of parameters.

However, we do believe transportability is a vehicle to bring us closer to unique identification with DSFs. It is clear from our experiments that transportable learners greatly improve edge accuracy. As our synthetic setup is governed by one (and thus unique) graph, having a more accurate learner means a learner that discovers a DAG that is more like the unique, underlying graphical model.  Consider \cref{fig:axis} for an illustration comparing the relevant methods in terms of model identification.

\section{Transportability in non-overlapping domains} \label{app:non-overlapping-domains}

Consider the multi-origin setting, where we have at least two datasets, each stemming from a different source. It is entirely possible that, given the different sources, these datasets are not comparable in terms of recorded features. We can recognise two major manifestations of this phenomenon: either (i) the supports of the datasets do not match, or (ii) the dimensions do not match.

{\bf (i) Different support.} Recall from \cref{sec:prelims} that DAGs encode a set of independence statements. As such, it is mainly independence that governs structure. Transportability in the setting of conflicting support, thus requires some (mild) assumptions. Specifically, we require that independence holds, regardless of support. This is mostly a pragmatic assumption. If for example, we find that $\mathcal{X}_i \independent \mathcal{X}_j$, where each component denotes a dimension in $\mathcal{X}$, we usually don't specify over what support this independence holds. Implicitly, we assume that independence holds, regardless of what area in $\{\mathcal{X}_i, \mathcal{X}_j\}$ we find ourselves in.

Note that the chosen distributions in $\mathcal{P}$ in \cref{sec:method:single-origin} govern the entire domain $[N]$, and as a consequence $\mathcal{X}$. As such, the problem of conflicting support does not manifest in our solution  of single-origin D-Struct. In case one chooses distributions that do not cover $[N]$ equally, we have to assume independence is constant across different supports (i.e. the assumption explained above).

{\bf (ii) Different dimensions.} A more difficult setting of conflicting domains, is when we record different variables in each of the multi-origin datasets. In order for a DAG to be transportable, we {\it require} the variable sets to correspond. As such, we are only able to work with overlapping intersections of the non-overlapping domains. Doing so requires some additional assumptions on the noise: assuming we record some noise on each variable, we have to make the additional assumption that the noise is independent of the other variables, or at least the variables outside the intersection between domains. The latter is made quite often, and should not limit the applicability of D-Struct in this setting too much (recall that the applicability of D-Struct is mostly determined by the used DSF). The reason relates to the second assumption, below.

The second assumption is a bit stricter: any variables outside the intersection cannot be confounding variables inside the intersection. If two variables have no direct edges, and the nodes part of an indirect edge fall outside the domain-intersection, we have to expect the DSF to find an edge between these two nodes. While this direct edge is wrong, this is actually the expected behaviour of most DSFs as the algorithms will find these variables to be correlated (due to the third, now unobserved, variable). The only way to overcome these situations is to use DSFs that naturally handle unobserved confounding.

{\bf Related work.} Some work on structure discovery from multiple (non-overlapping) domains has been proposed. For example, \citet{ghassami2018multi} in the linear setting, \citet{peters2015} for the interventional setting, or \citet{huang2020causal} in the temporal setting. While the difference between the first (\citep{ghassami2018multi}) is clear (only focusing on linear systems, whereas we focus on a non-parametric setting), the others are not immediately clear. Some intuition into the difference can be achieved by considering that both the interventional and temporal {\it know} where the difference in distribution is coming from. So much so, that the known difference is exploited when garnering (causal) structural information. We believe applying our findings on transportability to the settings described earlier can be a promising new avenue of research.

\section{Definitions} \label{app:defs}

\begin{definition}[Markov blanket.] \label{def:markov_blanket}
A Markov blanket of a random variable $X_i$ in a random set $\mathcal{X} \coloneqq \{X_1, \dots, X_d\}$ is any subset $\mathcal{X}' \subset \mathcal{X}$ where, when conditioned upon, results in independence between $\mathcal{X}\setminus\mathcal{X}'$ (the other variables) and $X_i$,
\begin{equation} \label{eq:independence}
    X_i \independent \mathcal{X}\setminus\mathcal{X}' | \mathcal{X}'.
\end{equation}
We will denote the Markov blanket of $X_i$ as $\mathcal{X}'(X_i)$.
\end{definition}

In principle, \cref{def:markov_blanket} means that $\mathcal{X}'$ contains all the information present in $\mathcal{X}$ to infer $X_1$. Note that this does not mean that $\mathcal{X}\setminus\mathcal{X}'$ contains {\it no} information to infer $X_1$, but variables in $\mathcal{X}'$ are sufficient to predict $X_1$.

One step further, is a {\it Markov boundary} \citep{pearl1988probabilistic}:

\begin{definition}[Markov boundary.] \label{def:markov_boundary}
A Markov boundary of a random variable $X_i$ of a random set $\mathcal{X} \coloneqq \{X_1, \dots, X_d\}$ is any subset $\mathcal{X}^- \subset \mathcal{X}$ which is a Markov blanket (\cref{def:markov_blanket}) itself, but does not contain any proper subset which itself is a Markov blanket. We will denote the Markov boundary of $X_i$ as $\mathcal{X}^-(X_i)$.
\end{definition}

We can relate the Markov boundary (\cref{def:markov_boundary}) to probabilistic graphical modelling, as from a simplified factorisation (in \cref{eq:decom}), we can compose a Bayesian network. Specifically, each variable $X_j \in \mathcal{X}^-(X_i)$ depict one of three types of relationships: $X_j$ is a parent of $X_i$, denoted as $\pa{X_i} = X_j$; $X_j$ is a child of $X_i$, denoted as $\ch{X_i} = X_j$; or $X_j$ is a parent of a child of $X_i$, denoted as $\pa{\ch{X_i}} = X_j$. Assuming that $\mathbb{P}_\mathcal{X}$ is governed by a Markov random field (rather than a Bayesian network) simplifies things, as the Markov boundary depicts only directly connected variables.

While the above may suggest that the Markov boundary only implies a vague graphical structure, doing this for every variable in $\mathcal{X}$ will strongly constrain the possible graphical structures respecting any found independence statements. D-separation (\cref{def:d-separation}) is then used to further limit the set of potential DAGs \citep{pearl2009causality, peters2017elements}. Relating the above definitions to those discussed in \cref{sec:prelims}. For more information regarding the above, we refer to \citet{koller2009probabilistic}.

\begin{definition}[d-separation \citep{peters2017elements}.] \label{def:d-separation}
In a DAG $\mathcal{G}$, a path between nodes $\mathcal{X}_i$ and $\mathcal{X}_j$ is blocked by a set $\mathcal{X}_d \subset \mathcal{X}$ (which excludes $\mathcal{X}_i$ and $\mathcal{X}_j$) whenever there is a node $\mathcal{X}_k$, such that one of two holds:

\quad (1) $\mathcal{X}_k \in \mathcal{X}_d$ and
\vspace{-2pt}\begin{align*}
    &\mathcal{X}_{k-1} \leftarrow \mathcal{X}_k \leftarrow \mathcal{X}_{k+1},\\
    \text{or}\quad &\mathcal{X}_{k-1} \rightarrow \mathcal{X}_k \rightarrow \mathcal{X}_{k+1},\\
    \text{or}\quad &\mathcal{X}_{k-1} \leftarrow \mathcal{X}_k \rightarrow \mathcal{X}_{k+1}.
\end{align*} 
\quad (2) neither $\mathcal{X}_k$ nor any of its descendants is in $\mathcal{X}_d$ and
\begin{equation*}
    \mathcal{X}_{k-1} \rightarrow \mathcal{X}_k \leftarrow \mathcal{X}_{k+1}.
\end{equation*}
Furthermore, in a DAG $\mathcal{G}$, we say that two disjoint subsets $\mathcal{A}$ and $\mathcal{B}$ are d-seperated by a third (also disjoint) subset $\mathcal{X}_d$ if every path between nodes in $\mathcal{A}$ and $\mathcal{B}$ is blocked by $\mathcal{X}_d$. We then write
\begin{equation*}
    \mathcal{A} \independent_{\mkern-7mu\mathcal{G}\mkern5mu} \mathcal{B} | \mathcal{X}_d.
\end{equation*}
When $\mathcal{X}_d$ d-seperates $\mathcal{A}$ and $\mathcal{B}$ in $\mathcal{G}$, we will denote this as $\text{d-sep}_\mathcal{G}(\mathcal{A};\mathcal{B}|\mathcal{X}_d)$.
\end{definition}

\begin{definition}[Faithfulness from \citet{peters2017elements}.] \label{def:faithfulness}

Consider a distribution $\mathbb{P}_\mathcal{X}$ and a DAG $\mathcal{G}$

\quad\quad\begin{minipage}[t]{.8\textwidth}
(i) $\mathbb{P}_\mathcal{X}$ is faithful to $\mathcal{G}$ if 
\begin{equation*}
    \mathcal{A} \independent \mathcal{B} | \mathcal{C} \Rightarrow \mathcal{A} \independent_{\mkern-7mu\mathcal{G}\mkern5mu} \mathcal{B} | \mathcal{C},
\end{equation*}

for all disjoint sets $\mathcal{A}, \mathcal{B}$ and $\mathcal{C}$.
\end{minipage}

\quad\quad\begin{minipage}[t]{.8\textwidth}
(ii) a distribution satisfies causal minimality with respect to $\mathcal{G}$ if it is Markovian with respect to $\mathcal{G}$, but not to any proper subgraph of $\mathcal{G}$.
\end{minipage}

\end{definition}

Part (i) posits an implication that is the opposite of the global Markov condition
\begin{equation*}
    \mathcal{A} \independent_{\mkern-7mu\mathcal{G}\mkern5mu} \mathcal{B} | \mathcal{C} \Rightarrow \mathcal{A} \independent \mathcal{B} | \mathcal{C},
\end{equation*}
for which we refer to \citet[Def. 6.21]{peters2017elements}.

Part (ii) is actually implied when part (i) is satisfied, when $\mathbb{P}_\mathcal{X}$ is Markovian w.r.t. $\mathcal{G}$, as per \citet[prop. 6.35]{peters2017elements}. To have an idea of when faithfulness is not satisfied, we refer to \citet{zhang2008detection} and \citet[Theorem 3.2]{spirtes2000causation}.

\section{Incorporating prior knowledge on $\mathcal{I}(\mathbb{P})$ using L-BFGS-B}\label{app:priors}

Consider the following, where we wish to discover a structure between 3 variables: $X$, $Y$, $Z$, where the ground truth satisfies $X \independent Y | Z$. According to the rules of $d$-speration (cfr. \cref{def:d-separation}), we are always in a structure where $X$ and $Y$ are {\it only} directly connected to $Z$, i.e. no direct connection between $X$ and $Y$ exists. Let us further assume that the system is linear (as this is what vanilla NOTEARS assumes, but without loss of generality towards recent NOTEARS extensions), then we have the following,

\begin{mdframed}
\begin{minipage}[t]{.3\textwidth}
\quad {\bf structural equations}
    \begin{align*}
        X &\coloneqq \epsilon_X,\\
        Z &\coloneqq \beta_{Z, X} X + \epsilon_Z,\\
        Y &\coloneqq \beta_{Y, Z} Z + \epsilon_Y,
    \end{align*}
\end{minipage}%
~
\begin{minipage}[t]{.3\textwidth}
    \quad {\bf structure}
    \begin{equation*}
        \begin{tikzpicture}[
            roundnode/.style={circle, draw=black, fill=white,  minimum size=5mm, inner sep=0},
        ]
              \node[roundnode] (x) at (0,0) {$X$};
              \node[roundnode] (y) at (2,0) {$Y$};
              \node[roundnode] (z) at (1,0) {$Z$};
              
              \draw[->] (x) -- (z);
              \draw[->] (z) -- (y);
        \end{tikzpicture}
    \end{equation*}
\end{minipage}%
~
\begin{minipage}[t]{.3\textwidth}
\quad {\bf adjacency matrix}
    \begin{equation*}
        A = \begin{pmatrix}
            0 & 0 & 1\\
            0 & 0 & 0\\
            0 & 1 & 0
        \end{pmatrix}.
    \end{equation*}
\end{minipage}%
\end{mdframed}

Naturally, using only conditional independence, the directions of the arrows are not identifiable as explained above. However, NOTEARS is unable to narrow it down to the equivalence classes expressed in \cref{def:d-separation}. The reason is simple, NOTEARS' three optimisation components (the $h$-measure, an $L_2$ loss, and an $L_1$ regularizer on $A$, \citep{notears_sparsity2020}) are satisfied exactly the same with the following system:

\begin{mdframed}
\begin{minipage}[t]{.3\textwidth}
\quad {\bf structural equations}
    \begin{align*}
        X &\coloneqq \epsilon_X,\\
        Z &\coloneqq \beta_{Z, X} X + \epsilon_Z,\\
        Y &\coloneqq \beta_{Y, X} X + \epsilon_Y',
    \end{align*}
\end{minipage}%
~
\begin{minipage}[t]{.3\textwidth}
    \quad {\bf structure}
    \begin{equation*}
        \begin{tikzpicture}[
            roundnode/.style={circle, draw=black, fill=white,  minimum size=5mm, inner sep=0},
        ]
              \node[roundnode] (x) at (0,0) {$X$};
              \node[roundnode] (y) at (2,0) {$Y$};
              \node[roundnode] (z) at (1,0) {$Z$};
              
              \draw[->] (x) -- (z);
              \draw[->] (x) to[out=315, in=225] (y);
        \end{tikzpicture}
    \end{equation*}
\end{minipage}%
~
\begin{minipage}[t]{.3\textwidth}
\quad {\bf adjacency matrix}
    \begin{equation*}
        A' = \begin{pmatrix}
            0 & 1 & 1\\
            0 & 0 & 0\\
            0 & 0 & 0
        \end{pmatrix},
    \end{equation*}
\end{minipage}%
\end{mdframed}

where $\beta_{Y, X} = \beta_{Y,Z}\beta_{Z, X}$, and $\epsilon_Y'=\beta_{Y, Z}\epsilon_Z + \epsilon_Y$ resulting in $Y$ being determined again by a simple linear equation. Both systems allow the same data to be generated, however under the constraint that $X \independent Y | Z$ only the former is possible.

We argue that NOTEARS (and extensions) are unable to differentiate between them. Consider the components optimised by NOTEARS: both solutions propose a DAG (i.e. $h(A) = h(A') = 0$); each DAG has an equal amount of arrows, leading to the same $L_1$-loss across $A$ and $A'$; and each equation is linear so NOTEARS is able to perfectly converge to each solution using its $L_2$ loss. Given that each component scores exactly the same, NOTEARS is unable to differentiate between these two results. Crucially however, in the latter system $X$ is {\it always} dependent of $Y$, resulting in $X \not\independent Y |Z$ (and even $X \not\independent Y$ eliminating v-structures) which is completely opposite to the former system.

{\bf Prior Markov independencies.} We can however force known independence statements into DSFs a priori, using the L-BFGS-B optimizer. For example, consider the following $I = X_i \independent X_j | Z$. If $I$ is known a priori, then we also know there cannot (under any circumstance) exist a direct link between $X_i$ and $X_j$ as this would immediately contradict $I$ which in turn would invalidate a structure proposing such a link.

As such, we propose to fix these directed edges to $0 \to \mathcal{A}_{ij}(\mathcal{G}), \mathcal{A}_{ji}(\mathcal{G})$, and exclude them from gradient calculation. This will not only constrain each DSL in step 2 above resulting in easier convergence, but it will also enforce any known $\mathcal{I}(\mathbb{P}_\mathcal{X})$ to be taken into account. Setting $A_{X,Y} = A_{Y,X} = 0$ would immediately restrict NOTEARS from converging to this false solution as the solution would require $A_{X,Y}$ to be $1$. The same approach is currently used in NOTEARS (and consequentially D-Structs parallel DSFs), by setting bounds of each diagonal element in $A$ to $(0, 0)$.

Setting some elements to $0$ using the L-BFGS-B bounds, we effectively limit the set of possible solutions. In fact, when applied to the above problems, the second solution would sit {\it outside} the set of possible solutions, ensuring that NOTEARS cannot converge to it. 

\section{Additional details on subsampling from different distributions} \label{app:subsample}

In \cref{sec:method:single-origin} we introduced a method to sample subsets from a single-origin dataset such that the subsets correspond to distinct user-defined distributions. To provide some additional detail, we shall first discuss the general case, and then move on to discuss how we implemented this in D-Struct. 

\subsection{The general way} \label{app:subsample:general}
A high-level view of our subsampling routine is provided in \cref{fig:subsets}. From \cref{fig:subsets} we learn that we need two ingredients for our subroutine to work:
\begin{enumerate}
    \item We need a dataset that spans some domain $\mathcal{X}$. We can retrieve this domain simply by calculating the maximum and minimum value of each dimension in $\mathcal{X}$. {\it We have illustrated a simple dataset in \cref{fig:subsets:dataset}}.
    \item We need a set of $K$ distinct distributions that span $\mathcal{X}$. In principle, there is no constraint on these, besides them being different from one another, and each region in $\mathcal{X}$ having a non-zero probability of being sampled. {\it This is illustrated in \cref{fig:subsets:dists}}.
\end{enumerate}

Using the above two ingredients, we create $K$ empty subsets. For each subset, we then define one distribution, illustrated in \cref{fig:subsets:dists}. In \cref{fig:subsets} we used a Gaussian for each subset as they span the domain, and are simple to evaluate. Using these distributions, we will fill each subset using data from \cref{fig:subsets:dataset}. Each data point in our dataset is evaluated $K$ times: using the user-defined distributions in \cref{fig:subsets:dists}, we either include the sample in the corresponding subset, or not. When the probability of being sampled is {\it high enough}, it is included, when it is not high enough, it is excluded. High enough could be determined by something simple as a threshold, or something less parametric as a Bernoulli experiment. When finished, the subsamples look like \cref{fig:subsets:sampled}. 

Alas, Gaussian distributions become more difficult to handle with increasing dimensionality as data is spread sparser in high dimensions. The provided high-level example may serve well as a (visual) explanation of our subroutine, it does not work well in practice. As such, we used a different implementation for D-Struct, which we explain in \cref{sec:method:single-origin}, and in more detail below.

\subsection{How it's implemented in D-Struct}

Recall that the main issue with the simple Gaussian implementation above is that it does not scale well to high dimensions. As such, we need a different implementation that scales to high dimensions.

{\bf Defining the distributions.} We do this using a very simple idea: rather than sampling in covariate space, we sample the dataset's {\it indices}, which correspond to a sample's covariates. However, before we do this, we need to make sure that the indices are in some way correlated with the covariates, which is not the case for a standard dataset as they are sampled i.i.d.

To provide some correlation between index and covariates, we first sort the covariates and reindex the dataset. This way, a smaller set of covariates now corresponds with a smaller index-value. Note that it is unimportant whether we sort descending or ascending, the only thing that matters is that there is some {\it logical} ordering.

Having an index that is correlated with the covariates allows us to define a distribution over the indices (which are one-dimensional) rather than over the covariates (which are $d$-dimensional). We chose the beta distribution as our user-specified distribution, where each of the $K$ distributions is given different parameters. The advantage a beta distribution has is its flexibility to move its density over the entire domain (contrasting Gaussian distributions which are symmetrical). This point is illustrated in \cref{fig:k-distributions}.

{\bf Sampling data.} Once we have defined our distributions, we can use them to sample data. As with our high-level idea in \cref{app:subsample:general}, we will evaluate each data point $K$ times to determine whether or not it should be included in each subset. However, rather than evaluating the chosen distributions using the covariates directly, we now use the index instead. Regardless of the number of dimensions we have, the index remains one-dimensional.

Evaluating a sample in D-Struct is done using a Bernoulli experiment: with the beta distributions we query the probability of being sampled and provide it to a Bernoulli experiment, the outcome determines inclusion or exclusion.

\section{CIT-based methods, score-based methods and faithfulness} \label{app:related_work}

\subsection{CIT-based methods}
CIT-based methods such as the well-known PC-algorithm, the SGS algorithm, or the inductive causation (IC) algorithm all require faithfulness as per \cref{def:faithfulness}. The reason is such that they render the Markov equivalence class identifiable. As we have explained in \cref{sec:method:general}, using d-separation we have a one-to-one correspondence to this class of DAGs. Any query of a d-separation statement can therefore be answered by checking the corresponding conditional independence test \cite{peters2007machine}.

Most CIT-based methods have 2 main phases, based on a set of conditional independence statements. Assuming the latter is a correct set (that is, we have correctly inferred all the independence statements present in $\mathbb{P}_\mathcal{X}$ ), we first infer a skeleton graph, and then orient the edges. After these two phases, we have either a fully identified DAG, or Markov equivalence graphs in case there are edges we were not able to orient.

{\bf Phase 1: inferring a skeleton.} Based on \cref{lem:ic-sgs} (below) introduced in \citet{verma1990}, the SGS and IC algorithm build a skeleton from a completely unconnected graph.

\begin{lemma}\label{lem:ic-sgs}
The following two statements hold:

\quad\quad\begin{minipage}[t]{.8\textwidth}
(i) Two nodes $X$ and $Y$ in a DAG ($\mathcal{X}, \mathcal{E}$) are adjacent iff they cannot be d-separated by any subset $\mathcal{S} \subset \mathcal{X}\setminus\{X,Y\}$.
\end{minipage}

\quad\quad\begin{minipage}[t]{.8\textwidth}
(ii) If two nodes $X$ and $Y$ in a DAG ($\mathcal{X}, \mathcal{E}$) are not adjacent, then they are d-separated by either $\text{Pa}_X$ or $\text{Pa}_Y$.
\end{minipage}
\end{lemma}

Clearly, by using the above lemma, SGS \citep{spirtes2000causation} and IC \citep{pearl2009causality} {\it require} faithfulness. Contrasting methods that build from an unconnected graph is the PC-algorithm, which does the reverse: PC starts with a fully connected graph and step-by-step removes edges when they violate (ii) in \cref{lem:ic-sgs}. While a different approach, both require d-separation, i.e. this too requires faithfulness to hold!

{\bf Phase 2: orienting the edges.} As per \citet{meek1995strong}, there exists a set of graphical rules that is shown to correctly orient the edges based only on d-separation. Of course, this requires a {\it complete} set of correct independence statements which is arguably a much stricter assumption than faithfulness. 

Essentially, we can relax the assumption of a complete set of independencies, but we'll have to replace it with other assumptions. One such example is assuming a $\mathbb{P}_\mathcal{X}$ to be Gaussian (which is also quite strict, but it serves our example). With the latter assumption, we can test for {\it partial correlation} \citep[Appendices A.1 and A.2]{peters2017elements}, which allows us to identify the underlying Markov equivalence class \citep{kalisch2007estimating}. Furthermore, by additionally assuming a condition called {\it strong faithfulness} \citep{zhang2012strong,uhler2013geometry}, we have uniform consistency \citep{kalisch2007estimating}. We refer to \citet[Ex. 7.9]{peters2017elements} for an example.

\subsection{(Differentiable) Score-based methods}

Contrasting CIT-based methods are score-based methods. Score-based methods generalise our differentiable score-based methods and non-differentiable methods. Contrasting CIT-based methods, which directly encode the independence statements governing $\mathbb{P}_\mathcal{X}$ into $\mathcal{G}$, a score-based method will evaluate $\mathcal{G}$ on how well it fits the observed data. The rationale behind these score-based methods is that wrongly encoded independence statements will yield poor model fits \citep{geiger1994learning,heckerman2006bayesian}.

We can formalise a score-based method as a function $S$ which is to be optimised over candidate DAGs:
\begin{equation*}
    \hat{\mathcal{G}} \coloneqq \argmax_{\mathcal{G} \text{ DAG over } \mathcal{D}\in\mathcal{X}} S(\mathcal{D}, \mathcal{G}).
\end{equation*}
As such, there are two elements that comprise a score-based method: (i) the function $S$, and (ii) the way we optimise $S$. In our case, that is:
\begin{enumerate}[label={(\roman*)}]
    \item $S$ corresponds to \cref{eq:loss}, which is in large part determined by the underlying DSF through $\mathcal{L}_\text{DSF}$.
    \item $S$ is optimised using gradient-optimisation, which has proven very efficient in this problem setting
\end{enumerate}

Importantly, the rationale behind these methods does {\it not} require the faithfulness assumption for them to work. The latter may lead to violations against d-separation in case faithfulness does hold. However, in \cref{app:priors} we show how we can combat this by also incorporating any known independencies into our graph (which {\it does} require the faithfulness assumption to hold for those independence statements) using the L-BFGS-B optimisation algorithm.

\end{document}